%% file: main.tex
\documentclass{article} % For LaTeX2e
\usepackage{iclr2026_conference,times}

\input{math_commands.tex}

\usepackage{hyperref}
\usepackage{url}
\usepackage{booktabs}
\usepackage{tabularx}
\usepackage{makecell}
\usepackage{algorithm}
\usepackage{algpseudocode}
\usepackage{multirow} 
\usepackage{graphicx}
\usepackage{enumitem}
\usepackage{adjustbox}
\usepackage{rotating}
\usepackage{wrapfig}
\usepackage[skip=4pt]{caption}
\usepackage{pgfplots}
\pgfplotsset{compat=1.18}
\usepackage{pgfplotstable}
\usepgfplotslibrary{groupplots}
\usepackage{caption}
\usepackage{subcaption}
\usepackage{xcolor}
\usepackage{xspace}
\usepackage{pifont} % for checkmarks and crosses
\newcommand{\cmark}{\ding{51}} % check mark
\newcommand{\xmark}{\ding{55}} % cross mark

\newcommand{\ourmethod}{{LIGHT}\xspace}
\newcommand{\ourdataset}{{BEAM}\xspace}

\usepackage{longtable}

\usepackage[utf8]{inputenc}
\usepackage[most]{tcolorbox}
\usepackage{lipsum} 
\usepackage{enumitem}
\usepackage{listings}
\usepackage{xcolor}
\lstdefinelanguage{json}{
    basicstyle=\ttfamily\small,
    showstringspaces=false,
    breaklines=true,
    backgroundcolor=\color{gray!5},
    frame=none,
    literate=
      {"}{{{\color{black}{"}}}}{1},
}

\algrenewcommand\algorithmicrequire{\textbf{Input:}}
\algrenewcommand\algorithmicensure{\textbf{Output:}}

\title{Beyond a Million Tokens: Benchmarking and Enhancing Long-Term Memory in LLMs}

\author{Mohammad Tavakoli$^1$, Alireza Salemi$^2$, Carrie Ye$^1$, Mohamed Abdalla$^1$, \\
\textbf{Hamed Zamani$^2$, J.~Ross Mitchell$^1$} \\
$^1$University of Alberta \quad
$^2$University of Massachusetts Amherst \quad \\
\texttt{\{tavakol5, cye, mabdall2, jmitche2\}@ualberta.ca} \\
\texttt{\{asalemi, zamani\}@cs.umass.edu}
}

\iclrfinalcopy % Uncomment for camera-ready version, but NOT for submission.
\begin{document}

% \vspace{-2cm}
\maketitle
\vspace{0cm}
\begin{abstract}
\vspace{0cm}
Evaluating the abilities of large language models (LLMs) for tasks that require long-term memory and thus long-context reasoning, for example in conversational settings, is hampered by the existing benchmarks, which often lack narrative coherence, cover narrow domains, and only test simple recall-oriented tasks. This paper introduces a comprehensive solution to these challenges. First, we present a novel framework for automatically generating long (up to 10M tokens), coherent, and topically diverse conversations, accompanied by probing questions targeting a wide range of memory abilities. From this, we construct \ourdataset, a new benchmark comprising 100 conversations and 2,000 validated questions. Second, to enhance model performance, we propose \ourmethod--a framework inspired by human cognition that equips LLMs with three complementary memory systems: a long-term episodic memory, a short-term working memory, and a scratchpad for accumulating salient facts. Our experiments on \ourdataset reveal that even LLMs with 1M token context windows (with and without retrieval-augmentation) struggle as dialogues lengthen. In contrast, \ourmethod consistently improves performance across various models, achieving an average improvement of 3.5\%–12.69\% over the strongest baselines, depending on the backbone LLM. An ablation study further confirms the contribution of each memory component.
\end{abstract}

\section{Introduction}

Large language models (LLMs) have been deployed across diverse applications, including open-domain conversational agents \citep{laban2025llmslostmultiturnconversation,chen2025broaden}, retrieval-augmented generation (RAG) for open-domain question answering and fact checking \citep{lewis2020retrieval,salemi2025planandrefinediversecomprehensiveretrievalaugmented,10.1145/3731120.3744584,kim2024retrievalenhancedmachinelearningsynthesis}, long-document and code analysis \citep{li2025surveylongdocumentretrievalplm,jelodar2025largelanguagemodelsllms,10.5555/3698900.3698947}, and scientific or legal research \citep{rueda2025understandingllmscientificreasoning,NGUYEN2025106165}.
Many of these tasks demand models capable of processing long inputs, motivating LLMs such as Gemini \citep{google2025gemini2flash} with input windows of up to 1M tokens.
Among these domains, conversational systems present an intuitive and critical need for extended context, as users often engage in protracted, multi-session dialogues that require consistent memory across lengthy interactions \citep{zhong2024memorybank,xu2022long,du2024perltqa,tan2025membench}. This highlights the importance of evaluating how well LLMs can reason over and utilize long conversational histories.

While there are many prior efforts on studying and evaluating long-term memory of LLMs \citep{kim2024dialsim,xu2021beyond,maharana2024evaluating,zhong2024memorybank,xu2022long,du2024perltqa,tan2025membench}, existing benchmarks have fundamental limitations. Most extend conversation length by artificially concatenating short sessions of different users, producing dialogues with abrupt topic shifts and weak narrative coherence. Such a construction artificially simplifies evaluation because distinct segments are easily separable, reducing the need for true long-range reasoning. Furthermore, these datasets typically target narrow domains---often limited to personal-life scenarios---leaving many real-world application areas underrepresented. Finally, they emphasize simple context recall, overlooking other critical memory abilities such as contradiction resolution, recognizing evolving information, and instruction following.

\begin{figure}[t]
    \centering
    \includegraphics[width=\textwidth]{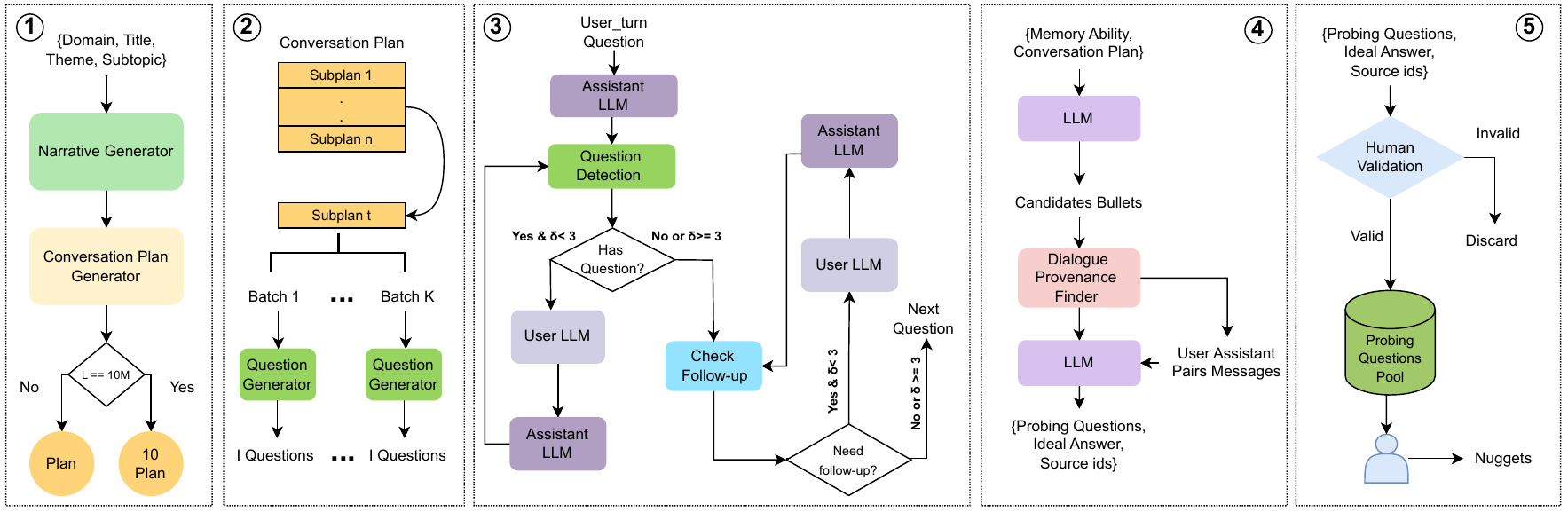}
    \caption{Overview of \ourdataset generation process. In the first stage, conversation plans are created for each chat seed. In the second stage, user utterances are generated from the conversation plans. In the third stage, assistant responses are produced. In the fourth stage, probing questions are generated based on the targeted memory abilities and corresponding conversation plans. In the final stage, invalid probing questions are filtered out, and nuggets are created for the validated set.}
    % \caption{Overview of \ourdataset generation process.}
    \label{fig:data_generation_diagram}
\end{figure}

To address these limitations, this paper presents a framework for automatically generating long coherent conversations between a user and an AI assistant---scaling up to 10M tokens on diverse domains---with a set of probing questions designed to evaluate diverse memory abilities of any LLM on the generated dialogues. An overview of the data generation framework is shown in Figure \ref{fig:data_generation_diagram}. This framework begins by defining a high-level conversation plan---a narrative for a particular domain and a simulated user with generated attributes---that outlines the overall flow of the dialogue. This plan is recursively decomposed into finer sub-plans that specify the storyline and its progression. From these sub-plans we generate chronologically ordered user turns, which are then expanded with corresponding assistant responses. To increase realism, the system injects follow-up questions and clarifications from both sides. Finally, we automatically create a set of probing questions that target ten distinct memory dimensions, with a focus on complicated and multi-hop reasoning, which are then validated by human annotators to ensure high quality. Using this pipeline, we construct the \ourdataset dataset: 100 diverse conversations ranging from 100 K to 10 M tokens each, accompanied by 2000 probing questions to evaluate the memory capabilities of LLMs.

To improve LLM performance on probing questions, we introduce the \ourmethod framework (Figure~\ref{fig:diagram}), which is applicable to both open-source and proprietary LLMs, inspired by research in human cognitive science and human's memorization and recall process \citep{sridhar2023cognitive, binder2011neurobiology}. This framework integrates three complementary memories: (1) episodic memory, a long-term index of the full conversation used for retrieval; (2) working memory, capturing the most recent user–assistant turns; and (3) a scratchpad, where after each turn the model reasons over the dialogue and records salient facts for future use. At inference, the LLM draws jointly on retrieved episodic content, the working memory, and the accumulated scratchpad to generate accurate answers.

To evaluate LLM memory capabilities and the effectiveness of our method, we conduct experiments on the constructed dataset, \ourdataset, using both open-source and proprietary models. Results show that even LLMs with long context windows perform substantially worse as conversation length increases. Our method improves the LLM's performance in answering the probing questions by 3.5\%–12.69\% on average over the best-performing baseline, depending on the backbone model and conversation length. An ablation study further reveals the contribution of each \ourmethod component on the performance. To support future work, we release all code, data, and evaluation scripts.\footnote{Available at: \href{https://github.com/mohammadtavakoli78/BEAM}{https://github.com/mohammadtavakoli78/BEAM}}

\section{\ourdataset: Benchmarking memory Capabilities of LLMs}

\subsection{Problem Formulation}
\label{sec:problem-formulation}

Let $\mathcal{D} = \{T_i\}_{i=1}^{|\mathcal{D}|}$ denote a collection of $|\mathcal{D}|$ conversations between users and a conversational agent $\pi$. Each conversation is represented as $\mathcal{T} = \{t_i\}_{i=1}^{|\mathcal{T}|}$, where $t_i \in \mathcal{T}$ corresponds to the $i$\textsuperscript{th} utterance (turn) in the dialogue. The objective of this work is to systematically evaluate a predefined set of memory abilities $\mathcal{M}$ exhibited by $\pi$ across conversations. For each memory ability $m \in \mathcal{M}$, we construct a probing dataset of size $N$, denoted as $\mathcal{Q}_m = \{(x_i, y_i)\}_{i=1}^{N}$, where $x_i$ is a probing question and $y_i$ is the corresponding ground-truth answer set. Each probing question $(x, y) \in \mathcal{Q}_m$ is appended as the $(|\mathcal{T}|+1)$\textsuperscript{th} turn in the dialogue, and the system generates a response $\hat{y} = \pi(x; \mathcal{T})$ based on the conversation. The generated response is then evaluated using an ability-specific scoring function $\mu_m$, producing a performance score $s = \mu_m(x, y, \hat{y})$. The goal of this work is to quantify the performance of conversational systems on each memory ability in $\mathcal{M}$.

\subsection{Benchmark Creation}
Our goal is to evaluate how well LLMs can answer questions that depend on long-term conversational memory. 
We measure performance across ten complementary abilities, seven drawn from prior benchmarks and three newly introduced here—\textit{Instruction Following}, \textit{Event Ordering}, and \textit{Contradiction Resolution} (see Table~\ref{tab:benchmark_comparison} in Appendix~\ref{sec:dataset_statistics}).
\textit{Abstention} evaluates whether a model withholds answers when evidence is missing.
\textit{Contradiction Resolution} tests the capacity to detect and reconcile inconsistent statements across widely separated turns, maintaining global coherence.
\textit{Event Ordering} assesses whether a model can recognize and reconstruct the sequence of evolving information in the dialogue.
\textit{Information Extraction} measures recall of entities and factual details in long histories.
\textit{Instruction Following} examines sustained adherence to user-specified constraints over long contexts.
\textit{Information Update} evaluates revising stored facts as new ones appear.
\textit{Multi-hop Reasoning} probes inference that integrates evidence across multiple, non-adjacent dialogue segments.
\textit{Preference Following} captures personalized responses that adapt to evolving preferences.
\textit{Summarization} assesses the ability to abstract and compress dialogue content, while \textit{Temporal Reasoning} tests reasoning about explicit and implicit time relations.
Together, these abilities evaluate a system's capacity to maintain, update, and manipulate information throughout extended conversations (see Appendix~\ref{appendix:ability_examples} for examples of each ability). Given these abilities and the formulation in Section \ref{sec:problem-formulation}, the benchmark requires three components: 1) a user–assistant conversation, 2) probing questions targeting key memory abilities, and 3) an evaluation methodology to assess the model's responses. The overall statistics of the constructed benchmark are summarized in Table~\ref{tab:dataset_statistics} in Appendix~\ref{sec:dataset_statistics}. The rest of this section details the process used to construct these components.

\paragraph{Overview:} The overview of our framework for creating conversations, probing questions, and the evaluation strategy is illustrated in Figure~\ref{fig:data_generation_diagram}.
The process begins by generating a simulated conversation between a user and an assistant. Structured conversation plans are first produced to guide the flow of the synthetic interactions. Each plan specifies sufficient information to generate both user and assistant turns, ensuring a coherent and natural conversational trajectory. While a typical exchange consists of a user question followed by an assistant response, realistic dialogues often involve follow-ups for clarification, elaboration, or related subtopics. To capture this, we incorporate two interaction-control modules.
The question-detection module identifies whether an assistant response includes a query that requires a user reply; if triggered, the system generates the corresponding user response. The follow-up detection module determines when the user would naturally pose a clarifying or elaborative question; if triggered,  it produces an additional user query for the assistant.
Together, these mechanisms produce conversations that exhibit interactive, bidirectional behavior beyond simple turn-taking. After the conversation is generated, an automated procedure constructs a candidate set of probing questions, each tailored to the specific memory abilities in the benchmark. These candidates are then reviewed by a human evaluator, who selects valid questions and formulates the associated evaluation rubrics used for subsequent benchmarking. A case study and an example of the different generated components of a conversation is provided in Appendix~\ref{sec:beam_different_stages_examples}.

\subsubsection{Conversation Plan Generation}
\label{subsubsec:plan_generation}

A \textit{conversation plan} serves as the scaffold for each dialogue, providing a coherent storyline that unfolds chronologically. Each plan is generated using an LLM based on seed information, including: the conversation \emph{domain}; a \emph{title and theme}; \emph{subtopics} outlining specific topics; a set of \emph{narratives} defining evolving aspects (e.g., career progression, goals); a \emph{user profile} with attributes such as name, age, gender, location, profession, and personality traits sampled from the Myers–Briggs Type Indicator (MBTI); a \emph{relationship graph} linking the user to family, friends, and acquaintances, constrained for realism (e.g., age gaps); and an explicit \emph{timeline} specifying the span of the conversation. 

To generate candidate titles and themes, human annotators specify target domains, then GPT-4.1 \citep{openai2025gpt41} generates candidate titles, themes, and subtopics using Listing~\ref{seed_creation}. Human reviewers refine outputs for topical diversity. For each conversation, we generate 15-20 narratives using the open-source LLaMA-3.3 70B model \citep{meta_llama3.3_modelcard} with the prompt in Listing~\ref{narrative_generation} (Appendix~\ref{sec:prompts}). Given the conversation seed, this model produces narrative elements capturing the evolving storyline, forming the backbone of a coherent conversation.

Conversation plans consist of $N$ \textit{sub-plans}, each representing a distinct stage in the conversation. Each sub-plan contains $M$ \textit{bullet points}, defined by a \textit{narrative}, a descriptive statement of its role in the storyline, and a \textit{time anchor}. For conversations of 128K, 500K, and 1M tokens, a single plan is generated (line~\ref{line:singleplan} in Algorithm~\ref{alg:plan_generation}, Appendix~\ref{sec:algorithms}) by conditioning the LLM on the conversation seed, profile, relationship graph, timeline, and specified counts of sub-plans, bullet points, and narratives (prompt in Listing~\ref{general_plan_generation}, Appendix~\ref{sec:prompts}). The number of sub-plans varies with domain and target length to meet the token requirement; e.g., coding domains generally require fewer turns than broader domains. For 10M-token conversations, one plan cannot capture the scope, so we create ten interlocking plans forming a coherent longer narrative. The process begins with a global seed defining the overall topic and theme, but a single seed is insufficient; instead, we derive ten distinct seeds---one per plan---so the narrative can evolve across stages. We propose two strategies:
\begin{itemize}[leftmargin=*]
\item \textbf{Sequential Expansion:} The global seed defines the initial point in the conversation's chronology. Subsequent seeds represent successive events (e.g., a trip, job search, later milestones). Using the prompt in Listing~\ref{ten_million_sequential_seed_generation} (Appendix~\ref{sec:prompts}), each new seed is generated from the main seed, profile, and timeline. Plans are then produced sequentially (line~\ref{line:sequential_expansion} in Algorithm~\ref{alg:plan_generation}, Appendix~\ref{sec:algorithms}), with each plan conditioned on its predecessor to maintain continuity. Core relationships (e.g., parents) remain fixed, while new acquaintances are gradually introduced to reflect the evolving context.

\item \textbf{Hierarchical Decomposition:} 
The main seed is decomposed into ten sub-seeds, each representing a distinct topical and temporal segment. Together, these sub-seeds span the full storyline (e.g., an international trip: first three for preparation, next five for trip events, final two for reflections). Similar to sequential expansion, the user's core relationships remain constant, while new acquaintances are introduced to reflect the evolving context. These ten sub-seeds are generated using the prompt in Listing~\ref{ten_million_hierarchical_seed_generation} (Appendix~\ref{sec:prompts}), conditioned on the main seed, profile, and timeline.
\end{itemize}

Each conversation plan is assigned explicit topical and temporal boundaries—encoded in the seed—to avoid redundancy and ensure sub-themes appear in the right narrative stage. For coherence, the LLM conditions on summaries of prior plans and future seeds when producing a new plan, allowing anticipation of upcoming events (e.g., reserving tickets for travel dates). This procedure is implemented in line~\ref{line:hierarchical_decomposition} of Algorithm~\ref{alg:plan_generation} (Appendix~\ref{sec:algorithms}). Plans are generated using the prompt in Listing~\ref{ten_million_hierarchical_plan_generation} (Appendix~\ref{sec:prompts}), conditioned on the main seed, current sub-seed, number of sub-plans, narrative set, user profile, core and new relationships, preceding and subsequent sub-seeds, previous plan, a summary of earlier plans, current sub-seed index, and a binary flag for the first plan (triggering user introduction). Since initial plans may not sufficiently test three key memory abilities—\textit{contradiction resolution}, \textit{information update}, and \textit{instruction following}—we apply a two-stage augmentation: first generate the base plan, then use GPT-4.1 (Listing~\ref{add_special_bullets}) to augment each sub-plan with three targeted bullet points. Performing augmentation separately improves coverage and fidelity. The refinement follows the prompt in Listing~\ref{add_special_bullets} (Appendix~\ref{sec:prompts}), which takes plan as input and outputs the revised version. This stage corresponds to the first module in Figure~\ref{fig:data_generation_diagram}, which forms the first step of the overall data-generation pipeline. The detailed process for plan generation is reported in Appendix~\ref{sec:conversation_plan_generation_details}.

\subsubsection{User Utterance Generation}

Once conversation plans are constructed, user utterances are synthesized from the sub-plans. Each sub-plan contains $M$ bullet points, which are divided sequentially into $K$ contiguous batches of equal size. Batching narrows the LLM's focus, reducing repetition and low-quality outputs that can occur when conditioning on the entire sub-plan. For each batch, the LLM generates $I$ user questions (line~\ref{line:question_generation} in Algorithm~\ref{alg:question_generation} in Appendix~\ref{sec:algorithms}) using the prompt in Listing~\ref{question_generation_general} (Appendix~\ref{sec:prompts}), conditioned on the conversation seed, the current batch, preceding batches, and context from earlier sub-plans. Each generated user question constitutes a user turn in the dialogue, ensuring coherence and continuity across extended conversations. Values of $K$ and $I$ are manually specified based on domain and target conversation length to meet the token budget, with configurations reported in Table~\ref{tab:batching_config} (Appendix~\ref{sec:bench-design}). This provides fine-grained control over user interaction density, preventing under-generation or redundancy. To balance quality and cost, question generation uses the open-source LLaMA-3.3 70B model \citep{meta_llama3.3_modelcard}, which produces high-quality outputs efficiently as the backbone LLM. This user-utterance construction aligns with the second stage in Figure~\ref{fig:data_generation_diagram}. The details of this procedure for user utterance generation are provided in Appendix~\ref{sec:user_turn_generation_details}.

\vspace{-0.1cm}
\subsubsection{Assistant Utterance Generation}

Assistant-side responses are generated iteratively in a role-playing setup, where one LLM assumes the \textit{assistant role} and another the \textit{user role}. For each sub-plan, the assistant LLM is conditioned on the conversation seed (Section~\ref{subsubsec:plan_generation}), prior sub-plans, a summary of the last $M$ turns, and a compressed summary of earlier ones (using the prompt in Listing~\ref{assistant_llm_answer_generation} in Appendix~\ref{sec:bench-design}); for 10M-token conversations, additional summaries of prior plans are provided. The assistant first generates a response to the user's most recent question (line~\ref{line:assistant_response_generation} in Algorithm~\ref{alg:answer_generation} in Appendix~\ref{sec:algorithms}), which is analyzed by a \textit{question-detection module} (line~\ref{line:question_detection_module}  in Algorithm~\ref{alg:answer_generation} in Appendix~\ref{sec:algorithms}, using the prompt in Listing~\ref{check_include_question} Appendix~\ref{sec:bench-design}) to determine the presence of a counter-question. If detected, the response is passed to the user LLM, which generates a contextually consistent reply based on the current and prior sub-plans, relevant history, and conversation summaries (using the prompt in Listing~\ref{user_llm_answer_generation} in Appendix~\ref{sec:bench-design}, line~\ref{line:user_llm_response generation} in Algorithm~\ref{alg:answer_generation} in Appendix~\ref{sec:algorithms}). This loop continues until no further assistant questions are detected or the threshold $\delta_1=2$ is reached, balancing realism and avoiding infinite cycles. In addition, a \textit{follow-up detection module} (line~\ref{line:followup_detection_module} in Algorithm~\ref{alg:answer_generation} in Appendix~\ref{sec:algorithms}, using the prompt in Listing~\ref{check_need_followup} in Appendix~\ref{sec:bench-design}) evaluates whether a clarifying or elaborative user follow-up is warranted, based on factors such as subject complexity, ambiguity, or incomplete responses. When required, the module generates a follow-up query conditioned on the seed, current and prior sub-plans, the most recent $M$ turns, and earlier summaries (using the prompt in Listing~\ref{user_llm_ask_followup} in Appendix~\ref{sec:bench-design}), which is then passed back to the assistant LLM. The number of follow-up exchanges is limited by a threshold $\delta_2=2$, analogous to $\delta_1$. Together, these modules yield dialogues with bidirectional dynamics, contextual referencing, and realistic clarifications, approximating human–AI interactions. This assistant-side generation maps to the third module in Figure~\ref{fig:data_generation_diagram}. The details of this procedure are provided in Appendix~\ref{sec:assistant_turn_generation_details}. 

\vspace{-0.1cm}
\subsection{Probing Questions Generation}
\label{sec:probing_questions_generation}

After constructing conversations, we generate probing questions to evaluate memory abilities. The pipeline combines automated synthesis with human validation: an LLM first produces candidate probes, which annotators review to select valid ones. Probes are derived from both the conversation plan and chat to ensure each targets a specific ability, is grounded in dialogue turns, and includes explicit provenance. The process begins by passing the plan to GPT-4.1-mini \citep{openai2025gpt41mini}, which selects candidate bullet points conditioned on the ability under evaluation. For example, knowledge-update probes require bullet pairs encoding an initial fact and its later revision, while summarization and event-ordering probes span multiple bullets. Each bullet is linked to its corresponding user and assistant turns through indices introduced during user-assistant turn generation, enabling retrieval of the precise dialogue segments in which the content was created. Candidate bullet selection is performed using prompts~\ref{information_extraction_selection}--\ref{instruction_following_selection}, one per memory ability. For abstention, candidate selection is unnecessary; probes are created directly from the plan using the prompt shown in Listing~\ref{abstention_probe_generation} (Appendix~\ref{sec:prompts}).

Given the selected bullet points and aligned dialogue snippets, GPT-4.1-mini generates the probing question, a candidate answer, and source identifiers citing the specific messages containing the answer. For 10M-token dialogues, candidate selection and synthesis are performed with a sliding window across the ten interlocking plans, processing a limited number at a time to preserve topical locality and scalability. Probe generation uses prompts~\ref{information_extraction_probe_generation}--\ref{instruction_following_probe_generation} for each memory ability, mapping candidate bullet points and contexts into fully formed questions. Finally, a human evaluator reviews the generated candidates and selects those that are valid and consistent with the conversation. Samples of probing questions are provided in Appendix~\ref{sec:nugget_design}, items~\ref{item:abstention}--\ref{item:temporal_reasoning}.

\subsection{Evaluation}

We evaluate LLMs on the probing questions using nugget evaluation, a common approach for long-form text assessment \citep{pradeep2024initial, pradeep2025great}. Each probing question is manually validated: invalid or unsupported questions are discarded, and minor inconsistencies are corrected. From the validated set, two questions per memory ability are chosen for each conversation, yielding 20 probing questions per conversation. Rubric nuggets are then derived for each question. A nugget is an atomic, self-contained criterion that a system response must satisfy. Annotators decompose the ideal reference answer into minimal semantic units, ensuring each nugget is both atomic and self-contained. System responses are scored against these nuggets by an LLM judge (Listing~\ref{llm_as_a_judge}, Appendix~\ref{sec:prompts}), which assigns 0 (unsatisfied), 0.5 (partially satisfied), or 1 (fully satisfied). Scores are averaged across nuggets to produce ability-level metrics. This nugget-based procedure applies to nine memory abilities; the exception is event ordering, where quality depends on both recall and correct sequence. We evaluate event ordering using the Kendall tau-b coefficient \citep{kendall1945treatment}, which considers both order and presence. To apply this metric, an LLM equivalence detector (using the prompt in Listing~\ref{llm_equivalence_detector} in Appendix~\ref{sec:prompts}) aligns events in system responses with nuggets, outputting \texttt{yes} if two snippets denote the same event/topic and \texttt{no} otherwise. Kendall tau-b is then computed over the aligned sequences, capturing both recall and ordering fidelity. Examples of nugget construction for each memory ability are provided in Appendix~\ref{sec:nugget_design}.

\section{\ourmethod: Improving Memory Capabilities of LLMs}

Inspired by research in human cognitive science \citep{sridhar2023cognitive,binder2011neurobiology}, humans employ two primary mechanisms for remembering and using knowledge: \textit{episodic memory}, the ability to recall specific personal experiences along with their context, and \textit{working memory}, the capacity to retain and manipulate information about recent events over short periods. In addition, maintaining notes on a \textit{scratchpad} provides an external record that supports long-term recall and later retrieval. Since answering questions in long-context conversations similarly requires integrating past experiences and accumulated knowledge, we introduce a method that emulates these strategies by combining episodic recall, short-term working memory, and an external scratch-pad mechanism.

\begin{figure}[t]    
\includegraphics[width=\textwidth]{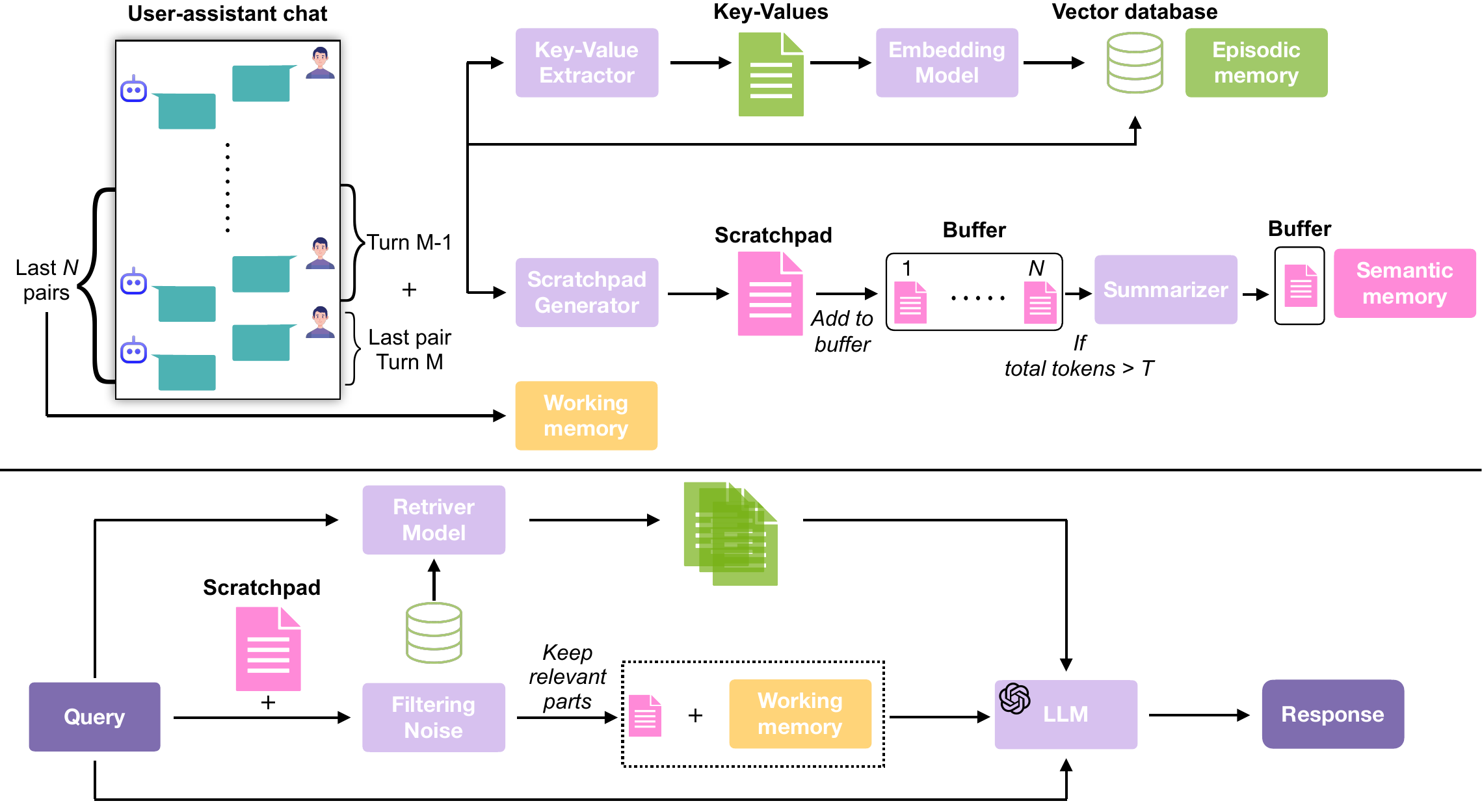}
    \caption{Overview of the \ourmethod framework.
    The system combines (i) \textbf{episodic retrieval}, (ii) a \textbf{scratchpad} and (iii) a \textbf{working memory buffer}. At inference, relevant items from the index and scratchpad, together with the full working memory, are integrated to generate the final response.}
    \label{fig:diagram}
\end{figure}

\paragraph{Overview:}
An overview of our method is shown in Figure~\ref{fig:diagram}.
Given a question $x$ about a conversation $\mathcal{T} = \{t_i\}_{i=1}^{|\mathcal{T}|}$, where $|\mathcal{T}|$ is the total number of turns, the framework first queries a retrieval model $R$ to obtain $k$ relevant segments from $\mathcal{T}$, simulating recall from episodic memory:
$E = R(x, k, \mathcal{T})$.
Next, the most recent $z$ dialogue pairs of the conversation are selected to form the working memory,
$W = \{ t_{|\mathcal{T}|-i} \}_{i=0}^{z}$.
In parallel, a pre-constructed scratchpad $S_{|\mathcal{T}|}$ contains up to $m$ salient notes.
A filtering function $f$ retains only the items pertinent to $x$, yielding
$S_{x} = f(S_{|\mathcal{T}|}, x)$.
Finally, the LLM $\pi$ generates the answer by conditioning on the question and these three memory components,
$y = \pi(x, E, W, S_{x})$ using the prompt shown in Listing~\ref{rag_answer_generation} in Appendix~\ref{sec:prompts}.
The remainder of this section details the construction and logic of each component in this pipeline.

\subsection{Retrieval from the Conversation}

\paragraph{Indexing the Conversation:}

After each user–assistant turn (Figure~\ref{fig:diagram}, top), we apply Qwen2.5-32B-AWQ \citep{qwen2.5} with the prompt in Listing~\ref{kv_extraction} (Appendix~\ref{sec:prompts}) to extract key–value pairs and a summary of the interaction. Keys represent entities and values capture attributes or descriptive details, providing fine-grained, event-level indices analogous to hippocampal memory traces \citep{teyler1986hippocampal}. These key–value pairs and summaries are embedded using the BAAI/bge-small-en-v1.5 embedding model \citep{bge_small_en_v1.5} and stored in a vector database as keys, while the original dialogue segments are kept as values to ensure faithful grounding.

\paragraph{Retrieval from the Index:} To retrieve information from the conversation as episodic memory, we embed the question $x$ using the same embedding model and compare it against the stored keys in the index, and the original dialogue segments corresponding to the top $k$ nearest neighbors are returned.

\subsection{Scratchpad Formation and Utilization}

\paragraph{Construction:}

In addition to episodic memory (Figure \ref{fig:diagram}, middle pathway), we build a higher-level representation that preserves information beyond individual dialogue events. It integrates semantic knowledge (facts and concepts), autobiographical details (life events), prospective memory (future intentions), and contextual metadata (time, place, acquisition context) \citep{binder2011neurobiology}. For each dialogue pair, we use Qwen2.5-32B-AWQ with the prompt in Listing \ref{scratchpad_creation} (Appendix \ref{sec:prompts}) to reason over the current and preceding turn and extract salient content. The resulting “scratchpad'' is iteratively merged with earlier versions; once content exceeds a 30K-token threshold—substantially shorter than the raw conversation—it is compressed into a 15K-token summary by GPT-4.1-nano using the prompt in Listing~\ref{summary_creation}. This process maintains efficiency and long-term coherence, analogous to the gradual abstraction of semantic memory in humans. Unlike the episodic index, the scratchpad is not stored in a retrieval database but is provided directly as contextual input during inference.

\paragraph{Filtering Scratchpad (function $f$):}

During inference, the scratchpad is selectively filtered with respect to the question. It is first divided into semantically coherent chunks using \emph{semantic chunking}. \footnote{\texttt{SemanticChunker} in LangChain is used, which segments text into variable-length passages based on semantic rather than fixed token windows.} Each chunk is evaluated by Qwen2.5-32B-AWQ with the prompt in Listing~\ref{noise_filtering} (Appendix~\ref{sec:prompts}), which assigns a binary relevance label (\texttt{yes}/\texttt{no}). Only the chunks judged relevant are retained, producing a condensed representation of scratchpad that is passed to the response generator.

\section{Experiments}

\subsection{Experimental Setup}

\paragraph{Baselines:}
We evaluate our approach against two types of baselines: long-context LLMs and a RAG method.
For long-context LLMs, the entire conversation history is provided, followed by the probing question. We include two proprietary LLMs (\textit{GPT-4.1-nano}, \textit{Gemini-2.0-flash}, both 1M context). and two open-source models (\textit{Qwen2.5-32B-AWQ}, \textit{Llama-4-Maverick-fp8}).
For long-context experiments, \textit{Qwen2.5-32B-AWQ} is evaluated with a 128K context length, while for the RAG baseline and our proposed method a 32K context length is used. At the 10M-token, since none of the four models support this length, they are evaluated on the largest recent dialogue segment fitting their window.\footnote{Among available models, only \textit{Llama-4-Scout} supports 10M-token context windows; however, due to its extreme computational requirements, we were unable to include it in our experiments.}
For RAG baselines, each user–assistant turn pair is treated as a document, embedded and stored in a vector database. At inference, the top five most similar documents are retrieved and passed to the LLM using the prompt in Listing~\ref{rag_answer_generation} (Appendix~\ref{sec:prompts}).

\begin{table}[!t]
    \centering
    \caption{Comparison of different LLMs and methods across conversation lengths and memory abilities using the created benchmark. Methods with the best performance per evaluation are bolded.}
    \adjustbox{max width=\textwidth}{
    \begin{tabular}{c|c|cc|c|cc|c|cc|c|cc|c}
        \hline
      \multirow{2}{*}{Length} & {Memory} & \multicolumn{3}{c}{Qwen 2.5} & \multicolumn{3}{|c}{Llama Maverick} & \multicolumn{3}{|c}{Gemini 2 Flash} & \multicolumn{3}{|c}{GPT-4.1-nano} \\
      \cline{3-14}
      & Ability & Vanilla & RAG & Ours & Vanilla & RAG & Ours & Vanilla & RAG & Ours & Vanilla & RAG & Ours \\
      \hline
      \multirow{11}{*}{100K} & Abstention & 0.300 & \textbf{0.650} & 0.475 & 0.200 & \textbf{0.800} & 0.600 & \textbf{0.800} & 0.800 & 0.675 & 0.475 & \textbf{0.800} & 0.575 \\
      & Contradiction Resolution & 0.031 & 0.025 & \textbf{0.037} & 0.025 & 0.031 & \textbf{0.031} & 0.006 & \textbf{0.050} & 0.018 & 0.012 & 0.018 & \textbf{0.031} \\
      & Event Ordering & 0.192 & 0.201 & \textbf{0.205} & \textbf{0.190} & 0.162 & 0.166 & 0.181 & \textbf{0.191} & 0.166 & \textbf{0.181} & 0.169 & 0.177 \\
      & Information Extraction & 0.425 & 0.338 & \textbf{0.479} & 0.510 & 0.392 & \textbf{0.518} & 0.333 & 0.341 & \textbf{0.464} & 0.273 & 0.362 & \textbf{0.538}	\\
      & Instruction Following & \textbf{0.400} & 0.375 & 0.362 & 0.412 & 0.375 & \textbf{0.412} & 0.275 & 0.287 & \textbf{0.362} & \textbf{0.425} & 0.350 & 0.400	\\
      & Knowledge Update & \textbf{0.437} & 0.275 & 0.362 & 0.300 & 0.350 & \textbf{0.450} & 0.125 & \textbf{0.325} & 0.300 & 0.275 & 0.375 & \textbf{0.375}	\\
      & Multi-Hop Reasoning & 0.222 & 0.203 & \textbf{0.281} & 0.152 & 0.225 & \textbf{0.353} & 0.200 &  0.148 & \textbf{0.225} & 0.178 & 0.263 & \textbf{0.365}	\\
      & Preference Following & 0.554 & 0.379 & \textbf{0.566} & 0.450 & 0.512 & \textbf{0.625} & 0.300 & 0.416 & \textbf{0.462} & 0.437 & 0.550 & \textbf{0.625}	\\
      & Summarization & 0.128 & 0.074 & \textbf{0.232} & 0.065 & 0.111 & \textbf{0.238} & 0.018 & 0.093 & \textbf{0.139} & 0.028 & 0.083 & \textbf{0.202} \\
      & Temporal Reasoning & 0.112 & \textbf{0.162} & 0.112 & 0.100 & \textbf{0.275} & 0.187 & \textbf{0.187} & 0.150 & 0.125 & 0.112 & 0.125 & \textbf{0.162}	\\
      \cline{2-14}
      & Average & 0.280 & 0.269 & \large\textbf{0.311} & 0.240 & 0.323 & \large\textbf{0.358} & 0.242 & 0.280 & \large\textbf{0.294} & 0.239 & 0.309 & \large\textbf{0.345}  \\
      \hline
      \multirow{11}{*}{500K} & Abstention & 0.314 & \textbf{0.728} & 0.571 & 0.185 & \textbf{0.785} & 0.628 & 0.714 & \textbf{0.800} & 0.685 & 0.557 & \textbf{0.828} & 0.600 \\
      & Contradiction Resolution & \textbf{0.053} & 0.017 & 0.017 & 0.035 & 0.028 & \textbf{0.042} & 0.010 & 0.021 & \textbf{0.021} & 0.017 & 0.025 & \textbf{0.035} \\
      & Event Ordering & 0.185 & 0.221 & \textbf{0.244} & \textbf{0.209} & 0.186 & 0.197 & \textbf{0.215} & 0.189 & 0.200 & 0.188 & 0.180 & \textbf{0.204}	\\
      & Information Extraction & 0.166 & 0.400 & \textbf{0.506} & \textbf{0.608} & 0.402 & 0.535 & 0.469 & 0.343 & \textbf{0.478} & 0.142 & 0.382 & \textbf{0.491}	\\
      & Instruction Following & 0.304 & \textbf{0.350} & 0.295 & 0.403 & \textbf{0.447} & 0.390 & 0.133 & \textbf{0.334} & 0.280 & 0.244 & 0.286 & \textbf{0.342} \\
      & Knowledge Update & 0.111 & 0.226 & \textbf{0.278} & 0.276 & \textbf{0.338} & 0.264 & 0.171 & 0.180 & \textbf{0.223} & 0.107 & \textbf{0.288} & 0.240	\\
      & Multi-Hop Reasoning & 0.125 & 0.187 & \textbf{0.214} & 0.219 & 0.313 & \textbf{0.350} & \textbf{0.198} & 0.135 & 0.157 & 0.070 & 0.233 & \textbf{0.266}	\\
      & Preference Following & 0.567 & 0.477 & \textbf{0.571} & 0.560 & 0.525 & \textbf{0.623} & 0.379 & 0.427 & \textbf{0.532} & 0.450 & 0.577 & \textbf{0.684}	\\
      & Summarization & 0.137 & 0.187 & \textbf{0.344} & 0.266 & 0.197 & \textbf{0.373} & 0.136 & 0.165 & \textbf{0.250} & 0.109 & 0.184 & \textbf{0.334} \\
      & Temporal Reasoning & 0.035 & 0.114 & \textbf{0.121} & 0.064 & 0.078 & \textbf{0.190} & \textbf{0.150} & 0.078 & 0.092 & 0.057 & \textbf{0.161} & 0.154	\\
      \cline{2-14}
      & Average & 0.200 & 0.291 & \large\textbf{0.316} & 0.283 & 0.330 & \large\textbf{0.359} & 0.257 & 0.267 & \large\textbf{0.292} & 0.194 & 0.314 & \large\textbf{0.335}  \\
      \hline
      \multirow{11}{*}{1M} & Abstention & 0.342 & \textbf{0.650} & 0.500 & 0.221 & \textbf{0.742} & 0.435 & 0.642 & \textbf{0.750} & 0.735 & 0.492 & \textbf{0.778} & 0.678 \\
      & Contradiction Resolution & \textbf{0.035} & 0.035 & 0.021 & \textbf{0.046} & 0.028 & 0.042 & 0.010 & \textbf{0.028} & 0.007 & \textbf{0.050} & 0.028 & 0.021 \\
      & Event Ordering & 0.183 & 0.195 & \textbf{0.200} & \textbf{0.214} & 0.179 & 0.193 & 0.190 & \textbf{0.198} & 0.185 & 0.191 & 0.179 & \textbf{0.211}	\\
      & Information Extraction & 0.138 & \textbf{0.407} & 0.366 & \textbf{0.489} & 0.431 & 0.474 & 0.374 & \textbf{0.380} & 0.341 & 0.153 & 0.399 & \textbf{0.410}	\\
      & Instruction Following	& 0.383 & 0.300 & \textbf{0.419} & \textbf{0.440} & 0.338 & 0.433 & 0.120 & 0.290 & \textbf{0.380} & 0.226 & 0.271 & \textbf{0.394} \\
      & Knowledge Update & 0.064 & \textbf{0.378} & 0.357 & 0.164 & 0.342 & \textbf{0.414} & 0.107 & \textbf{0.278} & 0.264 & 0.150 & 0.342 & \textbf{0.392}	\\
      & Multi-Hop Reasoning & 0.102 & 0.163 & \textbf{0.209} & 0.174 & 0.245 & \textbf{0.270} & 0.083 & 0.134 & \textbf{0.147} & 0.091 & \textbf{0.293} & 0.278	\\
      & Preference Following & 0.486 & 0.491 & \textbf{0.551} & 0.535 & 0.514 & \textbf{0.610} & 0.273 & 0.470 & \textbf{0.472} & 0.435 & 0.513 & \textbf{0.576}	\\
      & Summarization & 0.122 & 0.157 & \textbf{0.316} & 0.207 & 0.145 & \textbf{0.315} & 0.091 & 0.125 & \textbf{0.224} & 0.060 & 0.152 & \textbf{0.290} \\
      & Temporal Reasoning & 0.073 & 0.078 & \textbf{0.154} & 0.097 & 0.107 & \textbf{0.176} & \textbf{0.104} & 0.057 & 0.085 & 0.061 & 0.064 & \textbf{0.107}	\\
      \cline{2-14}
      & Average & 0.193 & 0.285 & \large\textbf{0.309} & 0.259 & 0.307 & \large\textbf{0.336} & 0.199 & 0.271 & \large\textbf{0.284} & 0.191 & 0.302 & \large\textbf{0.336} \\
      \hline
      \multirow{11}{*}{10M} & Abstention & 0.250 & \textbf{0.600} & 0.550 & 0.050 & \textbf{0.700} & 0.450 & \textbf{0.750} & 0.650 & 0.650 & 0.450 & \textbf{0.650} & 0.400 \\
      & Contradiction Resolution & \textbf{0.050} & 0.000 & 0.012 & \textbf{0.025} & 0.000 & 0.000 & 0.000 & \textbf{0.025} & 0.000 & 0.000 & 0.012 & \textbf{0.025} \\
      & Event Ordering & 0.180 & \textbf{0.221} & 0.197 & 0.190 & \textbf{0.220} & 0.176 & 0.220 & \textbf{0.266} & 0.193 & \textbf{0.215} & 0.201 & 0.173	\\
      & Information Extraction & 0.100 & 0.350 & \textbf{0.350} & 0.075 & \textbf{0.375} & 0.300 & 0.075 & \textbf{0.275} & 0.150 & 0.050 & 0.300 & \textbf{0.350}	\\
      & Instruction Following & 0.175 & 0.200 & \textbf{0.350} & 0.250 & 0.350 & \textbf{0.500} & 0.025 & 0.125 & \textbf{0.250} & 0.075 & 0.175 & \textbf{0.250}	\\
      & Knowledge Update & 0.100 & \textbf{0.300} & 0.275 & 0.100 & \textbf{0.375} & 0.325 & 0.050 & \textbf{0.325} & 0.200 & 0.050 & \textbf{0.325} & 0.300	\\
      & Multi-Hop Reasoning & 0.125 & 0.050 & \textbf{0.125} & 0.000 & 0.075 & \textbf{0.125} & 0.000 & 0.125 & \textbf{0.125} & 0.012 & 0.091 & \textbf{0.135}	\\
      & Preference Following & 0.241 & 0.291 & \textbf{0.308} & 0.291 & 0.316 & \textbf{0.483} & 0.075 & \textbf{0.300} & 0.150 & 0.175 & 0.366 & \textbf{0.425} 	\\
      & Summarization & 0.114 & 0.106 & \textbf{0.220} & 0.065 & 0.053 & \textbf{0.277} & 0.000 & 0.045 & \textbf{0.136} & 0.020 & 0.063 & \textbf{0.179} \\
      & Temporal Reasoning & 0.000 & 0.000 & 0.000 & 0.000 & 0.025 & \textbf{0.025} & 0.025 & 0.025 & \textbf{0.075} & \textbf{0.050} & 0.000 & 0.025	\\
      \cline{2-14}
      & Average & 0.133 & 0.211 & \large\textbf{0.238} & 0.104 & 0.249 & \large\textbf{0.266} & 0.122 & \large\textbf{0.216} & 0.192 & 0.109 & 0.218 & \large\textbf{0.226} \\
      \hline
    \end{tabular}}
    \label{tab:main_results}
    \vspace{-0.4cm}
\end{table}

\paragraph{Inference Setup:}
For inference, we use Nucleus \citep{Holtzman2020The} with temperature 0, except for conversation plan, user-turn, and assistant-turn generation, where temperature is 0.1 to encourage diversity. All open-source LLMs are served via VLLM for efficient inference. 
For Llama3.3-70B, we set the maximum output length to 6K tokens during user-turn generation, while for other LLMs we adopt their default maximum output length. For experiments involving both the RAG baseline and our proposed method, we employ FAISS as the vector database \citep{douze2024faiss}. For dense retrieval, we use the embedding model \textit{BAAI/bge-small-en-v1.5} \citep{bge_embedding}.

\subsection{Empirical Findings}
\paragraph{Main Results:}

Across all four conversation lengths (100K–10M tokens), our method consistently outperforms both long-context LLMs and RAG baselines (Table~\ref{tab:main_results}). At shorter contexts (100K), we observe strong gains, such as +49.1\% for Llama-4-Maverick and +44.3\% for GPT-4.1-nano over long-context baselines, showing that structured memory helps even when full history can be processed. The benefits grow with context length: at 1M tokens, improvements reach +75.9\% for GPT-4.1-nano and +60.1\% for Qwen2.5-32B. At 10M tokens—where no baseline natively supports the full context—our method achieves dramatic improvements, including +155.7\% for Llama-4-Maverick and +107.3\% for GPT-4.1-nano. The only exception is Gemini-2.0-flash at 10M, where our method surpasses the long-context baseline (+57.3\%) but slightly trails RAG, likely due to model-specific retrieval behavior. Overall, these findings underscore the scalability and robustness of our framework across diverse architectures and extreme context lengths.

When evaluated across the ten memory abilities, our method shows the largest relative gains in summarization (+160.6\%), multi-hop reasoning (+27.2\%), and preference following (+76.5\%). Strong improvements are also observed in information extraction (+56.7\%), instruction following (+39.5\%), and temporal reasoning (+56.3\%). These results highlight that our method is particularly effective for tasks requiring long-range recall and integration of dispersed information. 
In contrast, all methods---including ours---perform strongest in abstention and weakest in contradiction resolution, indicating that contradiction detection remains a challenging open problem.

\begin{figure*}[t] 
\centering
   \includegraphics[width=\textwidth]{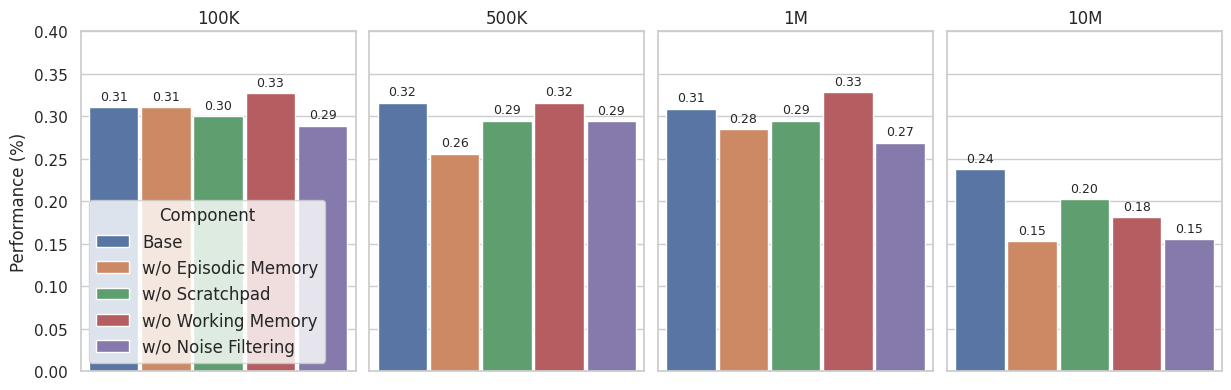}
    \caption{Ablation study illustrating the contribution of each component in \ourmethod (retrieval, scratchpad, working memory, and noise filtering) across different conversation lengths.} 
    \label{fig:ablation} 
    % \vspace{-0.8cm}
\end{figure*}

\paragraph{Ablation:}

We conduct an ablation to assess the role of each component—\emph{episodic memory}, \emph{scratchpad}, \emph{working memory}, and \emph{noise filtering}—across conversation lengths (Figure~\ref{fig:ablation}). At 100K, removing retrieval does not change performance and it remains steady, since the scratchpad alone suffices, while removing scratchpad or noise filtering reduces performance (–1.1\%, –2.2\%). Working memory also degrades results here (–1.6\%). At 500K, removing any component reduces performance except working memory, where removal enhances performance very slightly. At 1M, retrieval, scratchpad, and noise filtering remain beneficial, but removing working memory slightly improves performance, again reflecting its limited usefulness when few questions depend on the most recent turns. By 10M, all components are essential, with removals leading to large drops (–8.5\% for retrieval, –3.7\% for scratchpad, –5.7\% for working memory, –8.3\% for noise filtering). Overall, the ablations show that each module contributes increasingly as context length grows, and the full architecture consistently achieves the best performance. Detailed results across all memory abilities are provided in Table~\ref{tab:ablation_study}.

\paragraph{Effect of Retrieval Budget:}

\begin{wrapfigure}{r}{0.55\textwidth} 
    \vspace{-0.5cm}
    \includegraphics[width=0.55\textwidth]{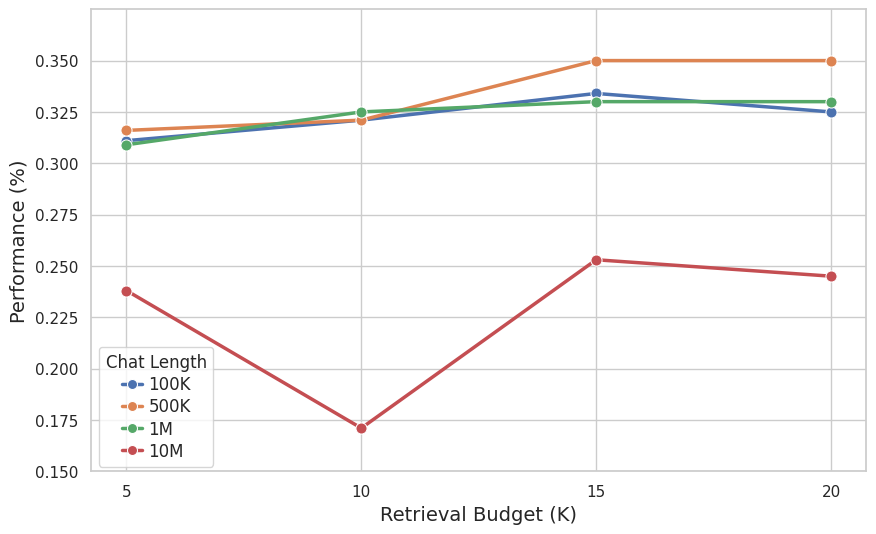}
    \vspace{-0.4cm}

    \caption{Effect of varying retrieval budget ($K$) on performance. The plot shows how the number of retrieved documents shapes the balance between recall and noise, highlighting different behaviors at short and long conversation lengths.} 
    \vspace{-0.4cm}
    \label{fig:number_of_documents_effect} 
\end{wrapfigure}

We examine the effect of retrieval budget (\emph{K}), testing 5, 10, 15, and 20 documents (Figure~\ref{fig:number_of_documents_effect}). Performance consistently improves when increasing \emph{K} from 5 to 15, with the best results at \emph{K}=15 (+7.39\%, +10.75\%, +6.79\%, and +6.3\% at 100K, 500K, 1M, and 10M). Increasing further to \emph{K}=20 slightly degrades performance, likely due to noisy context. Results at \emph{K}=10 are mixed---helpful at 100K, 500K and 1M but harmful at 10M---indicating additional documents sometimes add noisy information. Full results across memory abilities are shown in Table~\ref{tab:document_numbers_effect}. We also conducted complementary experiments analyzing the effect of retriever choice, where we observed that at 100K, 500K, and 1M token lengths, using a sparse retriever improves performance, whereas at 10M tokens, the dense retriever achieves better results. The full results and discussion are provided in Appendix~\ref{sec:retrieval_budget}.

\paragraph{Case Study}
A case study demonstrating the usefulness of the scratchpad is provided in Appendix~\ref{sec:scratchpad_case_study}.

Additional analyses on embedding choice, indexing setup, and a supplementary baseline are provided in Appendix~\ref{sec:embedding_model_effect}--\ref{sec:read_agent_baseline}.

\paragraph{Human Evaluation:}

We conducted a human evaluation to assess the quality of the generated conversations. Three dimensions were considered: \emph{Coherence and Flow}, \emph{Realism}, and \emph{Complexity and Depth}, each rated on a 5-point Likert scale (1 = lowest, 5 = highest). The average scores across all conversations were 4.53, 4.57, and 4.64, respectively, indicating consistently high quality. 
Details of the annotation protocol and inter-annotator agreement are provided in Appendix~\ref{sec:benchmakr_quality_evaluation}.
% The evaluation rubric and detailed scores are provided in Appendix~\ref{sec:benchmakr_quality_evaluation}. 
We further present a qualitative error analysis characterizing \ourmethod’s failure modes across memory abilities in Appendix~\ref{sec:qualitative_error_analysis}.

\section{Related Work}

The detailed related work is provided in Appendix~\ref{app:related-work}; here we present a concise summary.

Context windows of LLMs have expanded dramatically, from early limits of 512–2K tokens (GPT-2/3; \citep{radford2019language,brown2020language}) to 128K–1M (Claude-3, GPT-4-Turbo, Gemini 2.0; \citep{google2025gemini2flash,anthropic2025claude4,openai2025gpt41}) and even 10M (Llama 4; \citep{meta2025llama4}).
This growth is driven by advances in efficient attention (sparse, linear, memory-optimized kernels; \citep{beltagy2020longformer,wang2020linformer,dao2022flashattention}), improved positional encodings (relative, rotary with scaling, ALiBi; \citep{dai2019transformer,peng2023yarn}), long-context training strategies (continued-training, curriculum learning; \citep{xiong2023effective,ding2024longrope}), and inference optimizations such as paged attention, KV-cache compression, and distributed attention \citep{kwon2023efficient,zhang2023h2o,li2024snapkv,liu2023ring}.
Such capabilities are especially valuable for applications involving conversational histories, the main focus of our work.

Beyond expanding context windows, models incorporate additional mechanisms for persistent memory.
These include recurrence and compression (Transformer-XL, Compressive Transformer; \citep{dai2019transformer,rae2019compressive}),
state-space architectures (RWKV, Mamba, Hyena; \citep{peng2023rwkv,gu2023mamba,poli2023hyena}),
external memory modules (Memformer, RETRO, RMT; \citep{wu2020memformer,borgeaud2022improving,fan2024rmt}),
context summarization (AutoCompressor; \citep{chevalier2023adapting}),
and retrieval-augmented generation (REALM, RAG, HippoRAG; \citep{guu2020retrieval,lewis2020retrieval,jimenez2024hipporag}).
These approaches complement larger windows by enabling scalable and persistent long-term reasoning.

Existing benchmarks such as DialSim, MSC, LoCoMo, MemoryBank, DuLeMon, PerLTQA, LongMemEval, and MemBench \citep{kim2024dialsim,xu2021beyond,maharana2024evaluating,zhong2024memorybank,xu2022long,du2024perltqa,tan2025membench} evaluate recall, temporal reasoning, and multi-session reasoning, but typically span narrow domains, exhibit shallow dependencies, and concatenate separate user sessions to simulate long context, reducing realism.
Recent work such as MemoryCode \citep{rakotonirina2025tools} generates multi-session dialogues from template-driven instruction seeds to assess long-context reasoning, but focuses on a single domain.
Our benchmark instead scales to 10M tokens across diverse topics and introduces new tasks such as contradiction resolution, event ordering, and instruction following, generating coherent, single-user conversations that preserve narrative continuity for a more faithful assessment of long-term conversational memory.

\section{Conclusion}

This paper addresses the shortcomings of existing benchmarks for evaluating long-term memory in conversational systems. We introduce a scalable framework to generate \ourdataset, a new benchmark with long, coherent dialogues (up to 10M tokens) and diverse memory probes. To improve LLMs performance, we develop \ourmethod, a cognitive-inspired framework combining episodic, working, and scratchpad memories. Our experiments show that while standard LLMs' performance degrades over long contexts, \ourmethod provides substantial improvements, boosting memory performance by an average of 3.5\%-12.69\%. By offering a more robust evaluation and an effective memory enhancement technique, this work helps the development of more reliable long-context conversational systems.

\section*{Acknowledgments}
Ross Mitchell is the Alberta Health Services Chair in Artificial Intelligence in Health and is supported by CIFAR, the University Hospital Foundation, the Alberta Machine Intelligence Institute (Amii), and the Canada Foundation for Innovation. Mohamed Abdalla is supported by a CIFAR AI Chair. This research is supported by the Canadian Institutes of Health Research (FRF 196047). Carrie Ye is supported by the CRAF (CIORA)–Arthritis Society Canada Clinician Investigator Award (CI-24-0013). This research is supported in part by the Center for Intelligent Information Retrieval, in part by NSF grant \#2143434, in part by the Office of Naval Research contract \#N000142412612, and with support from Google.org. Any opinions, findings, and conclusions or recommendations expressed in this material are those of the authors and do not necessarily reflect those of the sponsors.

% % \subsubsection*{Author Contributions}

\newpage
\bibliography{iclr2026_conference}
\bibliographystyle{iclr2026_conference}

\newpage
\appendix

\section{Detailed Related Work}
\label{app:related-work}
\textbf{Long-Context Large Language Models.} 
The context window of LLMs has expanded from 512–2,048 tokens in early models (GPT-1/2/3, BERT, T5; \citep{radford2018improving,radford2019language,brown2020language,devlin2019bert,raffel2020exploring}) to 128K–1M tokens in recent systems (Claude-3, GPT-4-Turbo, Gemini 1.5 Pro, Gemini 2.0 Flash, Claude-4, GPT-4.1; \citep{anthropic2024claude3,achiam2023gpt,team2024gemini,google2025gemini2flash,anthropic2025claude4,openai2025gpt41}), with some reaching 10M tokens (Llama 4 Scout; \citep{meta2025llama4}). This growth has been enabled by innovations that address the quadratic cost of self-attention, including sparse mechanisms (Longformer, BigBird; \citep{beltagy2020longformer,zaheer2020big}), linear approximations (Linformer, Performer; \citep{wang2020linformer,choromanski2020rethinking}) and memory-efficient kernels (FlashAttention; \citep{dao2022flashattention}). Advances in positional encoding, such as relative encodings (Transformer-XL; \citep{dai2019transformer}), rotary embeddings (RoPE; \citep{su2024roformer}) with scaling methods (YaRN, NTK; \citep{peng2023yarn}), and linear biases (ALiBi; \citep{press2021train}), have extended usable context lengths. Training strategies like continued pre-training and curriculum learning (e.g., LLaMA-2-Long \citep{xiong2023effective}, LongRoPE \citep{ding2024longrope}) further expand capabilities, while inference optimizations such as PagedAttention \citep{kwon2023efficient}, KV-cache compression (H2O, SnapKV; \citep{zhang2023h2o,li2024snapkv}) and distributed approaches (Ring Attention; \citep{liu2023ring}) enable practical deployment at scale.

\textbf{Long-Term Memory Methods.} 
Researchers have developed approaches to enhance long-term memory beyond simply extending context windows. Architectural modifications include Transformer-XL \citep{dai2019transformer}, which introduced segment-level recurrence, and Compressive Transformer \citep{rae2019compressive}, which stored both recent states and compressed older information. State-space models such as RWKV \citep{peng2023rwkv}, Mamba \citep{gu2023mamba}, and Hyena \citep{poli2023hyena} replace attention with recurrent dynamics, allowing linear scaling and theoretically unbounded memory. Memory-augmented transformers such as Memformer \citep{wu2020memformer}, RETRO \citep{borgeaud2022improving} and RMT \citep{fan2024rmt} add external memory slots for explicit storage and recall. Context compression offers an orthogonal strategy by summarizing past information rather than storing it verbatim, as in AutoCompressor \citep{chevalier2023adapting}, which learns compact, information-preserving representations to reduce token usage. Retrieval-augmented generation (RAG) scales further by maintaining external knowledge stores: REALM \citep{guu2020retrieval} and RAG \citep{lewis2020retrieval} pioneered dense retrieval, RETRO \citep{borgeaud2022improving} integrated retrieval into transformers, and HippoRAG \citep{jimenez2024hipporag} incorporated structured knowledge graphs.

Building on these foundations, we propose a novel retrieval-augmented method that shows substantial improvements over baselines in long-memory evaluation.

\textbf{Long-Term Memory Benchmarks.} 
Several benchmarks have emerged to evaluate long-term memory capabilities in LLMs. DialSim \citep{kim2024dialsim} derives evaluation data from multiparty television scripts, producing dialogues extending to 350K tokens with naturalistic patterns but limited topical diversity. MSC \citep{xu2021beyond} introduces multisession human-assistant conversations testing memory across session boundaries, though with brief sessions and shallow dependencies. LoCoMo \citep{maharana2024evaluating} presents 50 conversations averaging 9K tokens in 35 sessions, while MemoryBank \citep{zhong2024memorybank} provides 300 sessions with 194 probing questions evaluating recall and temporal reasoning. DuLeMon \citep{xu2022long} focuses on dialogue-level memory and forgetting curves, PerLTQA \citep{du2024perltqa} targets memory classification and retrieval, and LongMemEval \citep{wu2024longmemeval} constructs multisession evaluations with 500 questions testing information extraction and temporal reasoning. More recently, MemBench \citep{tan2025membench} evaluates the memory of LLM-based agents by assessing their performance on information extraction, multi-hop reasoning, knowledge updating, preference following, and temporal reasoning. Recent work such as MemoryCode \citep{rakotonirina2025tools} generates multi-session dialogues from template-driven instruction seeds to assess long-context reasoning, but focuses on a single domain.

As summarized in Table~\ref{tab:benchmark_comparison}, the existing benchmarks are largely based on concatenated short sessions with limited coherence, narrow personal and casual domains, and few memory abilities. They also lack realistic bidirectional interactivity. In contrast, our benchmark spans diverse domains, scales up to 10M tokens, and introduces three additional dimensions—contradiction resolution, event ordering, and instruction following—yielding a more comprehensive framework for evaluating long-term memory in conversational systems.

\begin{table*}[t]
\caption{Comparison of our benchmark with existing long-term memory benchmarks. 
Memory abilities: IE = Information Extraction, MR = Multi-hop Reasoning, KU = Knowledge Update, TR = Temporal Reasoning, ABS = Abstention, CR = Contradiction Resolution, EO = Event Ordering, IF = Instruction Following, PF = Preference Following, SUM = Summarization.}
\label{tab:benchmark_comparison}
\vspace{0.5em}
\centering
\renewcommand{\arraystretch}{1.2}
\setlength{\tabcolsep}{4pt}
\resizebox{\textwidth}{!}{%
\begin{tabular}{l c c *{10}{c} c}
\toprule
\multirow{2}{*}{\textbf{Benchmark}} & 
\multirow{2}{*}{\textbf{Domain}} & 
\multirow{2}{*}{\textbf{Chat Length}} & 
\multicolumn{10}{c}{\textbf{Memory Abilities}} 
% & \multirow{2}{*}{\makecell[c]{\textbf{Follow-up} \\ \textbf{questions}}} 
\\
\cmidrule(lr){4-13}
 & & & IE & MR & KU & TR & ABS & CR & EO & IF & PF & SUM & \\
\midrule
MSC \citep{xu2021beyond} & Casual & $\sim$1K & \xmark & \xmark & \xmark & \xmark & \xmark & \xmark & \xmark & \xmark & \xmark & \xmark \\
DuLeMon \citep{xu2022long} & Casual & $\sim$1K & \xmark & \xmark & \xmark & \xmark & \xmark & \xmark & \xmark & \xmark & \xmark & \xmark \\
MemoryBank \citep{zhong2024memorybank} & Personal life & $\sim$5K & \cmark & \xmark & \xmark & \cmark & \xmark & \xmark & \xmark & \xmark & \xmark & \xmark \\
PerLTQA \citep{du2024perltqa} & Personal life & N/A & \cmark & \xmark & \xmark & \xmark & \cmark & \xmark & \xmark & \xmark & \xmark & \xmark \\
LoCoMo \citep{maharana2024evaluating} & Personal life & $\sim$10K & \cmark & \cmark & \xmark & \cmark & \cmark & \xmark & \xmark & \xmark & \xmark & \cmark \\
DialSim \citep{kim2024dialsim} & TV/Film scripts & $\sim$350K & \cmark & \cmark & \xmark & \cmark & \cmark & \xmark & \xmark & \xmark & \xmark & \xmark \\
LongMemEval \citep{wu2024longmemeval} & Personal life & 115K, 1M & \cmark & \cmark & \cmark & \cmark & \cmark & \xmark & \xmark & \xmark & \cmark & \xmark \\
MemBench \citep{tan2025membench} & Personal life & $\sim$100K & \cmark & \cmark & \cmark & \cmark & \xmark & \xmark & \xmark & \xmark & \cmark & \xmark \\
\midrule
\textbf{\ourdataset (This work)} & \makecell[l]{\textbf{Multi-domain:}\\ Coding, Math,\\ Health, Finance,\\ Personal life, ...} & \makecell[c]{\textbf{128K, 500K,}\\ \textbf{1M, 10M}} 
& \cmark & \cmark & \cmark & \cmark & \cmark & \cmark & \cmark & \cmark & \cmark & \cmark \\
\bottomrule
\end{tabular}}
\end{table*}

\section{Benchmark Design}
\label{sec:bench-design}

\subsection{Dataset Statistics}
\label{sec:dataset_statistics}

Table~\ref{tab:dataset_statistics} summarizes the statistics of the generated dataset, including averages of user messages, assistant messages, assistant and user follow-up questions, and dialogue turns across different chat sizes.

\begin{table}[H]
\centering
\renewcommand{\arraystretch}{1.2}
\setlength{\tabcolsep}{3pt}
\caption{Statistics of the dataset. Reported values are averages per chat in each chat size. 
\# User Messages and \# Assistant Messages denote the average number of utterances from the user and assistant, respectively. 
\# Answer Assistant Questions is the number of times the assistant posed a question that the user answered. 
\# Followup Questions is the number of follow-up questions asked by the user. 
\# Turns refers to the total number of dialogue turns.}
\label{tab:dataset_statistics}
\begin{tabularx}{\textwidth}{|c|c|c|c|c|c|}
\hline
\textbf{Chat Size} &
\textbf{\# User Messages} & 
\textbf{\# Assistant Messages} & 
\makecell{\textbf{\# Answer Assistant}\\\textbf{Questions}} & 
\makecell{\textbf{\# Followup}\\\textbf{Questions}} & 
\textbf{\# turns} \\
\hline
128K & 144 & 144 & 27 & 216 & 107 \\
500K & 544 & 544 & 79 & 51 & 416 \\
1M   & 1067 & 1067 & 105  & 120 & 842 \\
10M  & 10435  & 10435  & 1151  & 1528 & 7757 \\
\hline
\end{tabularx}
\end{table}

\subsection{Benchmark Quality Evaluation}
\label{sec:benchmakr_quality_evaluation}
To evaluate the quality of the generated conversations, we conducted a human assessment across all conversations. Two annotators rated each conversation on three dimensions using a 5-point Likert scale (1 = lowest, 5 = highest): \emph{Coherence and Flow}, \emph{Dialogue Realism}, and \emph{Complexity and Depth}.

\paragraph{Annotation Setup and Procedure.}

We used an internal system for the annotators. The annotations were completed by two annotators with a computer science background. Human involvement are as follows:

\begin{itemize}[noitemsep, topsep=2pt]
    \item \textbf{Probing Question Validation.}  
    The annotators were first provided with guidelines and instructions on how to perform the task, and a training session was held. They went through a calibration set together, completed multiple examples, and discussed their opinions on each example.

    \item \textbf{Rubric Design.}  
    The annotators were given guidelines on how to design rubrics along with examples and templates to make the task easier and more unified. A training and calibration session was then held, during which annotators reviewed many samples together, asked questions, and received clarification on ambiguous and edge cases.

    \item \textbf{Conversation Quality Evaluation.}  
    For this task, the annotators were again provided with guidelines on how to evaluate the generated conversations and assign scores. Because the conversations are very long and it is impractical to read them in full, annotators were instructed to read the initial 25 dialogue turns to understand what the conversation is about and to become familiar with the theme and flow of the chat. They were instructed to check the follow-up questions proposed by both the user and the assistant, the adherence of the conversation to the conversation plan and user utterances, and whether any hallucinations or unexpected behaviors occur. After that, they were asked to read several random turns between the initial 25 turns and the middle of the conversation to assess whether the conversation is progressing according to the conversation plan. They then had to read 25 dialogue turns in the middle of the conversation, followed by additional random turns between the middle and the end, and finally 25 dialogue turns near the end of the conversation. In this way, annotators read different parts of each conversation and ensured good coverage.
\end{itemize}

For every conversation, annotators rated the three quality dimensions described below using a 5-point Likert scale.

\begin{itemize}[noitemsep, topsep=2pt]
    \item \textbf{Coherence and Flow}:  
    Conversation continuity (each turn follows naturally from the previous one), 
    smooth transitions across topics and responses, 
    and thread consistency without abrupt or jarring shifts.  

    \item \textbf{Dialogue Realism}:  
    Naturalness of user queries (messages sound authentic), 
    realistic progression of topics over time, 
    human-like interactions (appropriate clarifications, follow-ups, etc.), 
    and believability of scenarios.  

    \item \textbf{Complexity and Depth}:  
    Handling of multi-layered, interconnected topics, 
    progressive increase in difficulty, 
    and demonstration of domain expertise when required.  
\end{itemize}

\paragraph{Inter-Annotator Agreement.}

Before performing the full evaluation, annotators independently assessed a set of 20 conversations. Agreement was measured using Cohen’s Kappa. The observed agreement levels were:

\begin{itemize}[noitemsep, topsep=2pt]
    \item \textbf{Coherence and Flow:} $\kappa = 0.7044$
    \item \textbf{Dialogue Realism:} $\kappa = 0.7391$
    \item \textbf{Complexity and Depth:} $\kappa = 0.7849$
\end{itemize}

The aggregated evaluation results are reported in Table~\ref{tab:human_eval_results}.  

\begin{table}[H]
\centering
\renewcommand{\arraystretch}{1.2}
\setlength{\tabcolsep}{6pt}
\caption{Conversation quality human evaluation results (1--5 scale). Higher is better.}
\label{tab:human_eval_results}
\begin{tabular}{|c|c|c|c|}
\hline
\textbf{Chat Size} & \textbf{Coherence and Flow} & \textbf{Dialogue Realism} & \textbf{Complexity and Depth} \\
\hline
128K & 4.4 & 4.55 & 4.35 \\
500K & 4.49 & 4.4 & 4.63 \\
1M   & 4.66 & 4.54 & 4.6 \\
10M  & 4.6 & 4.8 & 5 \\
\hline
\textbf{Average} & 4.53 & 4.57 & 4.64 \\
\hline
\end{tabular}
\end{table}

\subsection{Benchmark Creation Details}
\subsubsection{Domain Coverage of the Dataset}
To ensure broad coverage and realism, our dataset spans a diverse set of domains. The collection includes both technical and non-technical conversations, ranging from specialized domains such as coding, mathematics, financial investment and health to personal and social domains such as therapy, lifestyle, and trip planning. In total, we designed 100 multi-turn chats distributed across 19 domains, each represented by a set of distinct titles that capture the thematic scope of the dialogues. The full list of domains and their associated chat titles is provided in Table~\ref{tab:domains_chats}.

\begin{longtable}{p{3cm} p{10cm}}
\caption{Domains and associated chat titles in our dataset (100 total chats).}
\label{tab:domains_chats}\\
\toprule
\textbf{Domain} & \textbf{Chat Titles} \\
\midrule
\endfirsthead
\toprule
\textbf{Domain} & \textbf{Chat Titles} \\
\midrule
\endhead

% Your data goes here
Coding & Designing a Large-Scale Retrieval-Augmented Generation (RAG) System for Enterprise Search $\bullet$ Creating a Self-Driving Car Simulation Environment $\bullet$ Developing a Multi-Agent AI Research Platform $\bullet$ Building a Multi-Language AI Chatbot with Contextual Memory $\bullet$ Developing a Personalized News Aggregator with AI Summarization $\bullet$ Creating an Autonomous Stock Trading Bot $\bullet$ Implementing a Custom Image Captioning Model $\bullet$ Building a Multiplayer Online Game with Real-Time Physics $\bullet$ Building a Real-Time Chat Application with Node.js and Socket.io $\bullet$ Creating an AI-Powered Resume Analyzer with Python and NLP $\bullet$ Developing a Computer Vision App for Real-Time Object Detection $\bullet$ Creating a Restaurant Recommendation System $\bullet$ Automating Social Media Posts with Python $\bullet$ Building a Personal Budget Tracker Web App in Python and Flask $\bullet$ Creating a Command-Line To-Do List Manager in Go $\bullet$ Developing a Weather Forecast App in JavaScript with OpenWeather API $\bullet$ Training a Spam Email Classifier Using Python and Scikit-learn $\bullet$ Building a Portfolio Website with HTML, CSS, and Bootstrap \\
\midrule
Math & Partial Differential Equations (PDEs) in Depth $\bullet$ Functional Analysis and Infinite-Dimensional Spaces $\bullet$ Solving Ordinary Differential Equations (ODEs) $\bullet$ Deep Dive into Number Theory $\bullet$ Advanced Probability and Combinatorics $\bullet$ Exploring Non-Euclidean Geometry $\bullet$ Studying Multivariable Calculus $\bullet$ Diving into Analytic Geometry $\bullet$ Developing Skills in Mathematical Induction $\bullet$ Exploring Conic Sections in Depth $\bullet$ Understanding Sequences and Series $\bullet$ Mastering Basic Differential Calculus $\bullet$ Exploring the Geometry of Triangles $\bullet$ Understanding the Basics of Probability $\bullet$ Mastering Algebraic Equations for Everyday Problem Solving $\bullet$ Learning the Foundations of Trigonometry $\bullet$ Mastering Fractions, Decimals, and Percentages \\
\midrule
Writing Assistant \& Learning & Building a Portfolio-Ready Resume that Passes Any Applicant Tracking System $\bullet$ Mastering the Art of Persuasive Academic Essay Writing $\bullet$ Crafting a Standout Cover Letter for Competitive Job Markets $\bullet$ Designing a Multi-Purpose Personal Statement for Global Opportunities $\bullet$ Developing a Self-Editing System for Lifelong Writing Improvement \\
\midrule
Therapy \& Emotional Support & Recovering from Workplace Burnout and Chronic Stress $\bullet$ Healing After the Loss of a Loved One $\bullet$ Overcoming Childhood Trauma and Rebuilding Self-Trust $\bullet$ Coping with Post-Breakup Emotional Pain and Relationship Trauma \\
\midrule
Career \& Professional Development & Advancing from Mid-Level to Senior Leadership Roles $\bullet$ Building a Powerful Professional Network from Scratch $\bullet$ Landing Your Next Job: From Resume to Job Offer $\bullet$ Designing a 5-Year Career Growth Plan $\bullet$ Positioning Yourself for a Promotion \\
\midrule
Financial Investment & Building a Long-Term Stock Market Investment Strategy $\bullet$ Getting Started in Real Estate Investing $\bullet$ Navigating the World of Cryptocurrency $\bullet$ Creating a Balanced Investment Portfolio \\
\midrule
Health \& Wellness & Creating a Personalized Nutrition and Meal Planning System $\bullet$ Designing a Sustainable Fitness Routine $\bullet$ Improving Sleep Quality for Better Health $\bullet$ Understanding and Managing Chronic Illness $\bullet$ Recognizing Symptoms and Seeking Medical Help Early \\
\midrule
Relationship \& Family & Strengthening Communication in Romantic Relationships $\bullet$ Parenting Through Different Life Stages $\bullet$ Navigating In-Law and Extended Family Relationships $\bullet$ Rebuilding Relationships After Trust Has Been Broken \\
\midrule
Education \& Learning & Learning to Play a Musical Instrument from Scratch $\bullet$ Mastering a New Language for Real-World Communication $\bullet$ Becoming a Skilled Photographer $\bullet$ Exploring Performing Arts: Acting, Theater, and Dance \\
\midrule
Home \& Real Estate & Buying Your First Home with Confidence $\bullet$ Renting a Home or Apartment Without Stress $\bullet$ Selling Your Home for Maximum Value $\bullet$ DIY Home Improvement and Repairs $\bullet$ Making Your Home More Comfortable and Functional \\
\midrule
Lifestyle & Designing a Daily Routine That Boosts Productivity and Well-Being $\bullet$ Building Healthy and Sustainable Lifestyle Habits $\bullet$ Balancing Social Life and Personal Time \\
\midrule
Cooking & Mastering Quick and Healthy Weeknight Dinners $\bullet$ Baking Like a Pro at Home $\bullet$ Exploring Global Cuisines from Your Kitchen $\bullet$ Cooking for Special Diets and Allergies $\bullet$ Meal Prepping for the Week Ahead \\
\midrule
Business \& Entrepreneurship & Starting a Business from Scratch $\bullet$ Growing and Scaling Your Small Business $\bullet$ Building a Successful Startup \\
\midrule
Trip Planning & Preparing for a Week-Long Hiking and Camping Adventure in Patagonia $\bullet$ Organizing a Cross-Country USA Road Trip $\bullet$ Planning a Cultural Immersion Trip to Japan $\bullet$ Planning a Budget Backpacking Trip Across Southeast Asia $\bullet$ Arranging a Luxury Honeymoon in the Maldives \\
\midrule
Sport & Soccer – Playing, Watching, and Supporting the World's Most Popular Game $\bullet$ Basketball – From Street Courts to the NBA $\bullet$ Volleyball – Indoor, Beach, and Competitive Play $\bullet$ Hockey – Ice, Field, and Global Competitions $\bullet$ Tennis – From Local Courts to Grand Slams \\
\midrule
Event Planning & Planning a Surprise 30th Birthday Party for a Close Friend $\bullet$ Coordinating a Destination Beach Wedding for 100 Guests $\bullet$ Organizing a Weekend Community Food and Music Festival $\bullet$ Planning a Cozy Christmas Eve Dinner for Extended Family \\
\midrule
Asking Recommendation & Finding the Perfect Smartphone for Photography and Gaming $\bullet$ Choosing a Lightweight Laptop for Work, Travel, and Entertainment $\bullet$ Selecting a Must-Read Fiction Series for Winter Evenings $\bullet$ Finding the Best Streaming Movies for a Family Weekend $\bullet$ Choosing Comfortable and Stylish Sneakers for Daily Wear \\
\midrule
Legal \& Administrative & Filing for a Marriage-Based Green Card in the United States $\bullet$ Creating a Legally Valid Will and Estate Plan $\bullet$ Applying for a Patent to Protect a New Invention \\
\midrule
Philosophical \& Ethical Discussion & Deciding Whether to Use AI to Automate Hiring in My Company $\bullet$ Considering Whether to Believe in and Live by the Idea of Free Will \\
\bottomrule
\end{longtable}

\newpage
\subsubsection{Conversation Plan Generation}
\label{sec:conversation_plan_generation_details}
A \textit{conversation plan} serves as the central scaffold of each conversation, providing a coherent storyline that evolves chronologically. The process of constructing conversation plans is anchored by a \textit{seed} that specifies the \emph{domain} of the dialogue (e.g., sports, finance, programming, mathematics), a \emph{title} representing the high-level topic, and a \emph{theme} that provides a more detailed instantiation of the title. The seed also includes a set of \emph{subtopics}, which enumerate finer-grained subtopics and details to ensure topical diversity. However, a title, theme, and subtopics alone are insufficient to support detailed and information-rich conversations. To enrich the narrative, we introduce \emph{narratives set} that define the evolving aspects of a conversation (e.g., career progression, goals, relationships). Each narrative is paired with descriptive details that specify its scope and trajectory.

In addition to the seed and narrative set, each conversation incorporates a \emph{user profile}, a \emph{relationship graph}, and an explicit \emph{timeline}. The user profile includes attributes such as name, age, gender, location, profession, and personality traits. To avoid redundancy, personality traits are grounded in the Myers–Briggs Type Indicator (MBTI). Specifically, we randomly select six MBTI types, provide their descriptions, and instruct an LLM to synthesize a composite trait profile, enabling the creation of 8,008 unique user profiles. Relationship graphs are then constructed, linking the main user to family members (parents, partner, children), friends, and acquaintances, subject to constraints (e.g., plausible age gaps) to preserve realism. The timeline specifies the temporal span of the conversation, defining the range between its beginning and end.  

In order to generate titles and themes of the chats, target domains are first specified by human. Given these domains, GPT-4.1 \citep{openai2025gpt41} is prompted using the prompt shown in Listing~\ref{seed_creation} in Appendix~\ref{sec:prompts}, to produce candidate titles, themes, and subtopics. These candidates are refined by human to ensure topical diversity by removing the similar chat titles and selecting diverse chat titles. Finally, for each conversation, we generate 15–20 narratives using open-source LLaMA-3.3 70B \citep{meta_llama3.3_modelcard} with the prompt shown in Listing~\ref{narrative_generation} to save cost. In this prompt, given the conversation seed as input, the LLM produces narratives that capture evolving aspects of the storyline, providing the backbone for constructing coherent conversation plans.

Conversation plans are structured as a sequence of $N$ \textit{sub-plans}, where each sub-plan corresponds to a distinct stage of the conversation. Each sub-plan contains a fixed number of $M$ \textit{bullet-points}, and each bullet-point is defined by a \textit{narrative} and a descriptive statement specifying how that narrative unfolds in the storyline. To maintain temporal coherence, each sub-plan also includes a \textit{time anchor} specifying a concrete date or period.

For conversations of sizes 128K, 500K, and 1M tokens, a single conversation plan is generated, as shown in line~\ref{line:singleplan} of Algorithm~\ref{alg:plan_generation} in Appendix~\ref{sec:algorithms}. The plan is produced by conditioning the LLM on the conversation seed, user profile, relationship graph, timeline, the number of sub-plans, the number of bullet points within each sub-plan and narrative set, using the prompt shown in Listing~\ref{general_plan_generation} in Appendix~\ref{sec:prompts}. The number of sub-plans is not fixed but varies with both the domain and the target conversation length, in order to adhere to the length budget. For instance, domains such as coding typically require fewer dialogue turns to reach the same token budget compared to more general domains.

For 10M-token conversations, a single plan cannot adequately capture the scope and continuity required at this scale. To address this, we construct ten distinct yet interlocking conversation plans that together produce a coherent long-term narrative. While the process begins with a main seed that defines the global topic and theme of the conversation, a single seed is insufficient for producing ten plans. Instead, we generate ten distinct conversation seeds---one for each plan---so that the narrative can unfold across multiple stages.
The procedure for deriving these seeds—and the plans that follow—differs depending on the strategy. We propose two strategies for constructing them:
\begin{itemize}[leftmargin=*]
\item \textbf{Sequential Expansion:} 
The conversation seed is used as the first seed in the sequence. The remaining seeds are generated to represent successive stages of the user's life, extending the storyline chronologically. For instance, if the main seed concerns an international trip, the first plan covers the trip itself, the second covers the period after returning (e.g., job search), and subsequent seeds correspond to later milestones. We generate these seeds using the prompt shown in Listing~\ref{ten_million_sequential_seed_generation}, which conditions on the main seed, user profile, and timeline to produce a sequence of temporally aligned seeds. Each conversation plan is then generated sequentially, with every plan conditioned on its predecessor to maintain continuity, as specified in line~\ref{line:sequential_expansion} of Algorithm~\ref{alg:plan_generation} in Appendix~\ref{sec:algorithms}. The plans are generated using the prompt shown in Listing~\ref{ten_million_sequential_plan_generation}, yielding a temporally ordered series of interconnected narrative arcs. To maintain realism, the user's core relationships (e.g., parents, children, partner) remain fixed across plans, while new acquaintances are gradually introduced.

\item \textbf{Hierarchical Decomposition:} 
Instead of extending the seed chronologically, the main seed is decomposed into ten sub-seeds, each corresponding to a distinct topical or temporal slice of the overall storyline. Together, these seeds span the full narrative. For example, if the main seed concerns an international trip, the first three seeds may cover preparation steps (e.g., reservations, document gathering), the next five capture events during the trip, and the final two represent post-trip activities (e.g., reflections, recounting experiences). Like in Sequential Expansion, the user's core relationships (e.g., parents, children, partner) remain fixed across plans, while new acquaintances are gradually introduced. We generate these ten sub-seeds using the prompt shown in Listing~\ref{ten_million_hierarchical_seed_generation}, which takes the main seed, user profile, and timeline, and outputs ten derived seeds.
\end{itemize}

Each plan is assigned explicit topical and temporal boundaries to prevent redundancy or thematic overlap, ensuring that sub-themes unfold in the correct stage of the narrative. These boundaries are encoded in the conversation seed itself. For coherence, summaries of all prior plans are provided to the LLM when generating a new plan, allowing contextual references to past events. Moreover, when generating each plan, future seeds are also supplied, encoding their own topical and temporal boundaries. This design allows earlier plans to anticipate upcoming events with consistent references (e.g., booking tickets for the correct travel dates before the trip actually occurs). This strategy is implemented in line~\ref{line:hierarchical_decomposition} of Algorithm~\ref{alg:plan_generation} in Appendix~\ref{sec:algorithms}. Conversation plans are generated using the prompt shown in Listing~\ref{ten_million_hierarchical_plan_generation}, which takes as input the main seed, the current sub-seed, the number of sub-plans, the narrative set, the user profile, core and newly introduced relationships, the preceding and subsequent sub-seeds, the previous plan, the summary of all previous plans, the index of the current sub-seed, and a binary indicator specifying whether the plan is the first in the sequence (in which case the introduction of the user is included). The output is a fully specified conversation plan.

After the conversation plan is constructed, it is expanded into user-turn questions and subsequently assistant responses, yielding complete dialogues that can be used to evaluate memory abilities. However, in its initial form, the plan may not include sufficient information to evaluate three critical memory abilities: \textit{contradiction resolution}, \textit{knowledge update}, and \textit{instruction following}. To address this, after the initial plan generation, we pass the plan to GPT-4.1 to generate high-quality plans and augment each sub-plan with additional bullet points specifically designed to enable evaluation of these abilities. Importantly, this augmentation is performed in a second stage rather than during the initial plan generation, since incorporating such information directly in a single-pass generation leads to lower quality and less reliable coverage of these abilities. The augmentation is implemented using the prompt shown in Listing~\ref{add_special_bullets}, which takes an existing conversation plan as input and outputs a revised version where each sub-plan includes three additional bullet points targeting these abilities.

\subsubsection{User Utterance Generation}
\label{sec:user_turn_generation_details}
Once conversation plans are constructed, user turns are synthesized directly from them. Each sub-plan within a conversation plan consists of $M$ bullet-points, which are partitioned into $K$ contiguous batches of equal size. Partitioning is performed sequentially, such that each batch corresponds to a consecutive segment of the sub-plan. Partitioning is necessary because conditioning the LLM on an entire sub-plan at once tends to yield repetitive or low-quality questions; batching mitigates this by narrowing the focus of generation. For each batch, the LLM produces $I$ user questions (line~\ref{line:question_generation} of Algorithm~\ref{alg:question_generation} in Appendix~\ref{sec:algorithms}) using the prompt presented in Listing~\ref{question_generation_general}. The model is conditioned on the conversation seed, the current batch specification, preceding batches within the same sub-plan, and contextual information from earlier sub-plans. This setup ensures that generated questions remain grounded in prior context, yielding conversations that are coherent and continuous over extended spans.

The values of $K$ and $I$ vary depending on the domain and the target conversation length, in order to adhere to the overall length budget. We specify the values for $K$ and $I$ manually. The specific configurations of $K$ and $I$ across domains and conversation sizes are reported in Table~\ref{tab:batching_config}. This provides fine-grained control over the density of user interactions and helps prevent both under-generation and excessive redundancy. Additonally, to better capture domain-specific conversational patterns, we incorporate domain-specific features during question generation:
\begin{itemize}[leftmargin=*]
\item \textbf{Programming:} To reflect realistic developer–assistant interactions, we incorporate questions that involve sharing code snippets. These include (i) buggy code requiring debugging assistance, (ii) correct code seeking optimization, and (iii) natural language descriptions of desired functionality for which code is requested. We use the prompt shown in Listing~\ref{question_generation_coding} to generate questions specific to the programming domain.

\item \textbf{Mathematics.} To capture authentic problem-solving dynamics, we incorporate questions that involve sharing mathematical work, requesting corrections, asking for the next logical step in a solution, or introducing problems to be solved. We use the prompt shown in Listing~\ref{question_generation_math} to generate questions specific to the mathematics domain.
\end{itemize}

To reduce computational cost while maintaining generation quality, question generation is performed using the open-source LLaMA-3.3 70B model \citep{meta_llama3.3_modelcard}, which produces high-quality questions.

\subsubsection{Assistant Utterance Generation}
\label{sec:assistant_turn_generation_details}
After generating user-side questions, assistant-side responses are generated in an iterative, role-playing framework where one LLM assumes the \textit{assistant role} and another assumes the \textit{user role}. For each sub-plan, the assistant LLM is conditioned on the seed as explained in Section~\ref{subsubsec:plan_generation}, prior sub-plans of the conversation plan, a summary of the most recent $M$ dialogue turns, and a compressed summary of older turns (generated using the prompt shown in Listing~\ref{assistant_llm_answer_generation}). For 10M-token conversations, additional summaries of prior plans are also provided.

The response generation process unfolds as an iterative interaction between the assistant and user roles. First, the assistant LLM produces an answer to the user's most recent question (line~\ref{line:assistant_response_generation}). This output is then analyzed by a \textit{question-detection module}, which determines whether the assistant's response contains a counter-question directed at the user (line~\ref{line:question_detection_module}), using the prompt shown in Listing~\ref{check_include_question} that takes the assistant response as input and outputs \texttt{yes} if a question is present and \texttt{no} otherwise. If such a counter-question is detected, the response---together with the current and previous sub-plans, relevant past context, and conversation summaries---is passed to the user LLM, which generates a realistic reply that reflects the storyline and contextual details using the prompt shown in Listing~\ref{user_llm_answer_generation} (line~\ref{line:user_llm_response generation}). This new user reply is subsequently passed back to the assistant LLM, continuing the conversation. This loop repeats until no further assistant questions are detected or the predefined threshold $\delta_1$ (which is set to two) is reached, preventing infinite cycles. For $\delta_1$ we tested values 2, 3 and 5 which we selected 2 as it produces more realistic dialogues.

Beyond direct question–answer exchanges, a \textit{follow-up detection module} (line~\ref{line:followup_detection_module}) evaluates whether, in a realistic setting, the user would naturally ask a clarifying or elaborative follow-up. The need for a follow-up is determined using the prompt shown in Listing~\ref{check_need_followup}, which takes as input the seed, dialogue history, and the assistant's most recent response, and outputs \texttt{yes} or \texttt{no}. This decision is guided by factors such as subject complexity, ambiguity in the assistant's answer, or incompleteness of the response. When a follow-up is required, the module conditions on the seed, the current and prior sub-plans, the most recent $M$ turns, and summaries of earlier turns to generate the follow-up query using the prompt shown in Listing~\ref{user_llm_ask_followup}. The generated query is then passed back to the assistant LLM for resolution. As with the assistant-question loop, a strict threshold $\delta_2$ (which is set to two like $\delta_1$) limits the number of follow-up exchanges, preventing unbounded cycles.

Through the interaction of these two threshold-controlled modules, the system produces conversations that exhibit naturalistic bidirectional dynamics, rich contextual references, and realistic clarification behaviors characteristic of human–AI dialogues.

\subsubsection{Algorithms}
\label{sec:algorithms}
\begin{algorithm}[h]
\caption{Conversation plan generation.}
\label{alg:plan_generation}
\begin{algorithmic}[1]
\Require domain $c$, length budget $L$, title $\theta$, theme $\tau$, subtopics $\Sigma$, user profile $u$, user relationships $\rho$, timeline $\Gamma$, number of conversation sub-plans $N$, number of bullet-points in each conversation sub-plan $M$, generator $G$
\Ensure Conversation plan set $p$

\State $S \gets (c, \theta, \tau, \Sigma)$ \Comment{Initialize seed}

\If{$L \in \{128K, 500K, 1M\}$}
    \State $\Lambda \gets G(S)$ \Comment{Generate narratives using  Listing~\ref{narrative_generation}}
    \State $P \gets G(S, u, \rho, \Gamma, N, M, \Lambda)$ \label{line:singleplan} \Comment{Generate a single conversation plan with Listing~\ref{general_plan_generation}}
\ElsIf{$L = 10M$}
    \State $P \gets \{\}$ \Comment{Initialize set of plans}

    \If{$\sigma =$ Sequential Expansion}
        \State $S' \gets G_{\text{seeds}}(S, \Gamma)$ \Comment{Generate sequential sub-seeds with Listing~\ref{ten_million_sequential_seed_generation}}
        \For{each $s'_i \in S'$}
            \State $\Lambda_i \gets G(s'_i)$ \Comment{Generate narratives for sub-seed}
            \State $b \gets \mathbf{1}[i=0]$ \Comment{Binary indicator: 1 if first plan, else 0}
            \State $P_i \gets G(s'_i, \Gamma_i, N, \Lambda_i, u, \rho, P_{i-1}, i, b)$ \label{line:sequential_expansion} \Comment{Generate plan with Listing~\ref{ten_million_sequential_plan_generation}}
            \State $P \gets P \cup \{P_i\}$
        \EndFor
    \ElsIf{$\sigma =$ Hierarchical Decomposition}
        \State $S' \gets G_{\text{decompose}}(S, \Gamma)$ \Comment{Decompose seed with Listing~\ref{ten_million_hierarchical_seed_generation}}
        \For{each $s'_i \in S'$}
            \State $\Lambda_i \gets G(s'_i)$ \Comment{Generate narratives for sub-seed}
            \State $b \gets \mathbf{1}[i=0]$
            \State $P_i \gets G(S, S', s'_i, \Gamma_i, N, \Lambda_i, u, \rho, P_{i-1}, \overline{P_{0,\ldots,i-1}}, i, b)$ \label{line:hierarchical_decomposition} \Comment{Generate plan with Listing~\ref{ten_million_hierarchical_plan_generation}}
            \State $P \gets P \cup \{P_i\}$
        \EndFor
    \EndIf
\EndIf

\State \Return $P$

\end{algorithmic}
\end{algorithm}

\begin{algorithm}[t]
\caption{User questions generation.}
\label{alg:question_generation}
\begin{algorithmic}[1]
\Require seed $S$, conversation plan $p$, number of questions per iteration $I$, generator $G$
\Ensure Question set $Q$

\State $p \gets \{p_1, \ldots, p_N\}$ \Comment{Conversation plan with $N$ sub-plans}
\State $Q \gets \{\}$ \Comment{Initialize empty question set}

\For{each $p_i \in P$}
    \State $p_i = \{p_{i1}, \ldots, p_{iK}\}$
    \For{each $p_{ij} \in p_i$}
        \State $Q_{ij} \gets G(S, p_{ij}, \{p_{i1}, \ldots, p_{i(j-1)}\}, \{p_1, \ldots, p_{i-1}\}, $I$)$ \label{line:question_generation} \Comment{Generate $I$ questions using Listing~\ref{question_generation_general}}
        \State $Q \gets Q \cup \{Q_{ij}\}$ \Comment{Append generated questions to the question set}
    \EndFor
\EndFor

\State \Return $Q$

\end{algorithmic}
\end{algorithm}

\begin{algorithm}[H]
\caption{Answer generation.}
\label{alg:answer_generation}
\begin{algorithmic}[1]
\Require question set $Q = \{Q_1, \ldots, Q_N\}$, seed $S$, conversation plan set $P$, thresholds $\delta_1, \delta_2$, assistant-question detector $\phi$, follow-up detector $\psi$, generator $G$
\Ensure conversation list $T$

\State $\mathcal{T} \gets \{\}$ \Comment{Initialize empty conversation list}

\For{each $Q_i \in Q$} 
    \State $Q_i = \{q_1, \ldots, q_J\}$ \Comment{Questions in sub-plan $i$}
    \For{each $q_j \in Q_i$}
        \State $t \gets \{\}$ \Comment{Initialize turn sequence}
        
        \State $H_t^{(M)} \gets$ recent-$M$ turn window at turn $t$
        \State $\overline{H}_t \gets$ summary of turns prior to $H_t^{(M)}$
        \State $\overline{P}^{(<p)} \gets$ summaries of conversation plans preceding $p$
        
        \State $a_{ij} \gets G_{\text{assistant}}(S, p_{1:(i-1)}, H_t^{(M)}, \overline{H}_t, \overline{P}^{(<p)})$ \label{line:assistant_response_generation} \Comment{Generate assistant response with Listing~\ref{assistant_llm_answer_generation}}
        \State $t \gets t \cup \{a_{ij}\}$ \Comment{Add assistant's response to current dialogues turn}

        \State $isQ \gets \phi(a_{ij}, H_t^{(M)}, \overline{H}_t)$ \label{line:question_detection_module} \Comment{Checks if assistant response contains question from user with Listing~\ref{check_include_question}}
        \State $count \gets 0$
        \While{$isQ$ \textbf{and} $count < \delta_1$}
            \State $u_{ij} \gets G_{\text{user}}(S, p_i, p_{1:(i-1)}, \overline{P}^{(<p)}, H_t^{(M)}, \overline{H}_t, a_{ij})$ \label{line:user_llm_response generation} \Comment{Generate user's response to assistant question with Listing~\ref{user_llm_answer_generation}}
            \State $t \gets t \cup \{u_{ij}\}$ \Comment{Add user's response to current dialogues turn}

            \State $a_{ij} \gets G_{\text{assistant}}(S, p_{1:(i-1)}, H_t^{(M)}, \overline{H}_t, \overline{P}^{(<p)})$ \Comment{Generate assistant's response}
            \State $t \gets t \cup \{a_{ij}\}$ \Comment{Add assistant's response to current dialogues turn}

            \State $count \gets count + 1$
            \State $isQ \gets \phi(a_{ij}, H_t^{(M)}, \overline{H}_t)$
        \EndWhile

        \State $needFU \gets \psi(a_{ij}, H_t^{(M)}, \overline{H}_t, S)$ \label{line:followup_detection_module} \Comment{Checks if user need to ask followup question with Listing~\ref{check_need_followup}}
        \State $fu\_count \gets 0$
        \While{$needFU$ \textbf{and} $fu\_count < \delta_2$}
            \State $u_{ij} \gets G_{\text{user}}(S, p_i, p_{1:(i-1)}, \overline{P}^{(<p)}, H_t^{(M)}, \overline{H}_t, a_{ij})$ \Comment{Generate user's followup question with Listing~\ref{user_llm_ask_followup}}
            \State $t \gets t \cup \{u_{ij}\}$

            \State $a_{ij} \gets G_{\text{assistant}}(S, p_{1:(i-1)}, H_t^{(M)}, \overline{H}_t, \overline{P}^{(<p)})$ \Comment{Generate assistant's response to user's followup question}
            \State $t \gets t \cup \{a_{ij}\}$

            \State $fu\_count \gets fu\_count + 1$
            \State $needFU \gets \psi(a_{ij}, H_t^{(M)}, \overline{H}_t, S)$
        \EndWhile

        \State $ \mathcal{T} \gets \mathcal{T} \cup \{t\}$
    \EndFor
\EndFor

\State \Return $T$

\end{algorithmic}
\end{algorithm}

\subsection{User Utterance Generation Hyperparameters}
\begin{table}[H]
\renewcommand{\arraystretch}{1.2}
\setlength{\tabcolsep}{4pt}
\centering
\caption{Batching configuration by chat size and domain category for user-turn question generation.
NUM\_SUBPLANS denotes the number of conversation sub-plans, 
$K$ the number of batches per sub-plan, and $I$ the number of questions generated per batch. \\}
\label{tab:batching_config}
\begin{tabular}{|c|c|c|c|c|}
\hline
\textbf{Chat Size} & \textbf{Category} & \textbf{NUM\_SUBPLANS} & \textbf{K} & \textbf{I} \\
\hline
\multirow{3}{*}{128K} & General & 5 & 10 & 2 \\
 & Coding  & 3 & 23 & 1 \\
 & Math    & 3 & 25 & 1 \\
\hline
\multirow{3}{*}{500K} & General & 10 & 10 & 4 \\
 & Coding  & 10 & 10 & 3 \\
 & Math    & 10 & 10 & 4 \\
\hline
\multirow{3}{*}{1M} & General & 10 & 10 & 9 \\
 & Coding  & 10 & 10 & 6 \\
 & Math    & 10 & 10 & 6 \\
\hline
\multirow{3}{*}{10M} & General & 10 & 10 & 9 \\
 & Coding  & 10 & 10 & 6 \\
 & Math    & 10 & 10 & 6 \\
\hline
\end{tabular}
\end{table}

\subsection{Created Probing Questions Distribution}
We measure which parts of the dialogue contain the information required to answer the probing questions. To this end, each conversation is divided into ten equal segments, and we record the segment(s) where the supporting evidence for each probing question resides. The detailed methodology for aligning probing questions with dialogue segments is described in Section~\ref{sec:probing_questions_generation}. The resulting distributions across conversation lengths are reported in Table~\ref{tab:probing_distribution}.

\begin{table}[ht]
\centering
\renewcommand{\arraystretch}{1.2}
\setlength{\tabcolsep}{10pt}
\caption{Percentage distribution of created probing questions across ten equal chat segments (deciles) for different chat sizes. Each row corresponds to a segment of the dialogue, moving from the beginning (Segment~1) to the end (Segment~10).}
\label{tab:probing_distribution}
\begin{tabular}{|c|c|c|c|c|}
\hline
\textbf{Chat Segment (Decile)} & \textbf{100K} & \textbf{500K} & \textbf{1M} & \textbf{10M} \\
\hline
1  & 0.00\% & 0.65\% & 0.19\% & 0.00\% \\
2  & 11.05\% & 23.70\% & 21.60\% & 10.24\% \\
3  & 14.83\% & 15.91\% & 20.11\% & 16.27\% \\
4  & 12.79\% & 14.45\% & 15.83\% & 15.06\% \\
5  & 13.08\% & 7.95\%  & 9.50\%  & 14.46\% \\
6  & 13.37\% & 9.09\%  & 8.01\%  & 9.64\% \\
7  & 11.92\% & 6.33\%  & 5.96\%  & 10.24\% \\
8  & 8.14\%  & 5.52\%  & 5.21\%  & 13.25\% \\
9  & 9.59\%  & 4.55\%  & 4.47\%  & 8.43\% \\
10 & 5.23\%  & 11.85\% & 9.12\%  & 2.41\% \\
\hline
\end{tabular}
\end{table}

\newpage
\subsection{Memory Abilities Examples}
\label{appendix:ability_examples}
To illustrate how our benchmark evaluates different aspects of long-term conversational memory, we provide representative probing questions and their ideal answers for each of the ten memory abilities. These examples demonstrate how each ability is operationalized in practice.

\begin{enumerate}[leftmargin=1.5em]

\item \textbf{Abstention (withholding answers when information is missing)}
\begin{tcolorbox}[colback=gray!5, colframe=black!40, boxrule=0.3pt, sharp corners, breakable]
\textbf{Probing Question:} What specific advice did Manuel give about property management companies during the March 5 Investors Meetup? \\
\textbf{Ideal Answer:} Based on the provided chat, there is no information related to the specific advice Manuel gave about property management companies.
\end{tcolorbox}

\item \textbf{Contradiction Resolution (detecting and reconciling inconsistencies)}
\begin{tcolorbox}[colback=gray!5, colframe=black!40, boxrule=0.3pt, sharp corners, breakable]
\textbf{Probing Question:} Have I ever attended any real estate webinars or investor meetups? \\
\textbf{Ideal Answer:} I notice you've mentioned contradictory information about this. You said you have never attended any real estate webinars or investor meetups, but you also mentioned attending a webinar about Turkey's rising demand for multi-family rentals. Which statement is correct?
\end{tcolorbox}

\item \textbf{Event Ordering (reasoning about chronological sequence)}
\begin{tcolorbox}[colback=gray!5, colframe=black!40, boxrule=0.3pt, sharp corners, breakable]
\textbf{Probing Question:} How did my focus on different aspects of property investment and management develop throughout our conversations in order? Mention ONLY and ONLY ten items. \\
\textbf{Ideal Answer:} Your focus developed in this sequence: 1) Agent interaction and preparation, 2) Financial evaluation, 3) Financing and mortgage, 4) Contractor management, 5) Zoning and permits, 6) Renovation planning, 7) Viewing preparation, 8) Negotiation and cash flow, 9) Renovation completion, 10) Commercial property considerations.
\end{tcolorbox}

\item \textbf{Information Extraction (recalling factual details)}
\begin{tcolorbox}[colback=gray!5, colframe=black!40, boxrule=0.3pt, sharp corners, breakable]
\textbf{Probing Question:} What amount did I say I had set aside as my initial capital for investing? \\
\textbf{Ideal Answer:} You said you had set aside \$50,000 as your initial capital.
\end{tcolorbox}

\item \textbf{Instruction Following (adhering to explicit user constraints)}
\begin{tcolorbox}[colback=gray!5, colframe=black!40, boxrule=0.3pt, sharp corners, breakable]
\textbf{Probing Question:} How should I allocate my funds for the upcoming quarter? \\
\textbf{Ideal Answer:} The response should include an itemized list of costs, category-by-category breakdown, and detailed cost analysis, reflecting the instruction to always provide detailed financial breakdowns when budgeting.
\end{tcolorbox}

\item \textbf{Information Update (revising prior facts when updated)}
\begin{tcolorbox}[colback=gray!5, colframe=black!40, boxrule=0.3pt, sharp corners, breakable]
\textbf{Probing Question:} How much initial capital have I allocated for real estate investing? \\
\textbf{Ideal Answer:} \$60,000
\end{tcolorbox}

\item \textbf{Multi-hop Reasoning (integrating evidence across turns)}
\begin{tcolorbox}[colback=gray!5, colframe=black!40, boxrule=0.3pt, sharp corners, breakable]
\textbf{Probing Question:} How many different banks did I consider for my mortgage options across my conversations? \\
\textbf{Ideal Answer:} Two banks: Halkbank and Ziraat Bank.
\end{tcolorbox}

\item \textbf{Preference Following (adapting to evolving user preferences)}
\begin{tcolorbox}[colback=gray!5, colframe=black!40, boxrule=0.3pt, sharp corners, breakable]
\textbf{Probing Question:} I'm looking at several properties and want to understand how to approach comparing them effectively. What steps would you suggest I take? \\
\textbf{Ideal Answer:} A structured, step-by-step process that emphasizes data collection and analysis, breaking comparisons into measurable factors, reflecting the user’s preference for methodical, data-driven decisions.
\end{tcolorbox}

\item \textbf{Summarization (abstraction and compression of content)}
\begin{tcolorbox}[colback=gray!5, colframe=black!40, boxrule=0.3pt, sharp corners, breakable]
\textbf{Probing Question:} Can you give me a comprehensive summary of my journey and decision-making process around investing in rental properties, including how my budget, property choices, management considerations, and financing plans have developed over time? \\
\textbf{Ideal Answer:} Your journey began with an initial capital of \$50,000, followed by exploration of market conditions, renovation planning, property type tradeoffs, financing through Halkbank and Ziraat Bank, and a structured plan for purchase and management.
\end{tcolorbox}

\item \textbf{Temporal Reasoning (reasoning about durations and timelines)}
\begin{tcolorbox}[colback=gray!5, colframe=black!40, boxrule=0.3pt, sharp corners, breakable]
\textbf{Probing Question:} How many days are there between my first property viewing with Mehmet Yilmaz and the last one I scheduled? \\
\textbf{Ideal Answer:} There are 2 days between the first property viewing on March 25 and the last one on March 27.
\end{tcolorbox}

\end{enumerate}

\newpage
\section{Detailed Experiments}
\label{sec:detailed_experiments}
\subsection{Ablation Study}
In this section, we present the complete results of our ablation experiments. All ablations are conducted using Qwen2.5-32B-AWQ as the base model. We evaluate the contribution of individual components in our proposed module as shown in table~\ref{tab:ablation_study}.

\begin{table}[h]
    \centering
    \caption{Ablation study showing the impact of removing key memory components (retrieval, scratchpad, working memory, and noise filtering) on performance across various conversation lengths (100K–10M).}
    \adjustbox{max width=\textwidth}{
    \begin{tabular}{c|c|c|c|c|c|c}
        \hline
        Length & Memory Ability & Base & w/o Retrieval from Index & w/o Scratchpad & w/o Working Memory & w/o Noise Filtering \\
        \hline
        \multirow{11}{*}{100K} & Abstention & 0.475 & 0.725 & 0.600 & 0.575 & 0.700 \\
        & Contradiction Resolution & 0.037 & 0.043 & 0.012 & 0.043 & 0.018 \\
        & Event Ordering & 0.205 & 0.190 & 0.194 & 0.220 & 0.200 \\
        & Information Extraction & 0.479 & 0.329 & 0.510 & 0.451 & 0.485 \\
        & Instruction Following & 0.362 & 0.375 & 0.287 & 0.387 & 0.312 \\
        & Knowledge Update & 0.362 & 0.237 & 0.350 & 0.362 & 0.312 \\
        & Multi-Hop Reasoning & 0.281 & 0.201 & 0.248 & 0.303 & 0.181 \\
        & Preference Following & 0.566 & 0.675 & 0.533 & 0.579 & 0.491 \\
        & Summarization & 0.232 & 0.232 & 0.143 & 0.223 & 0.103 \\
        & Temporal Reasoning & 0.112 & 0.075 & 0.125 & 0.125 & 0.087 \\
        \cline{2-7}
        & Average & 0.311 & 0.311 & 0.300 & \textbf{0.327} & 0.289 \\
        \hline
        \multirow{11}{*}{500K} & Abstention & 0.571 & 0.571 & 0.585 & 0.657 & 0.585 \\
        & Contradiction Resolution & 0.017 & 0.007 & 0.014 & 0.017 & 0.014 \\
        & Event Ordering & 0.244 & 0.222 & 0.266 & 0.262 & 0.229 \\
        & Information Extraction & 0.506 & 0.254 & 0.466 & 0.485 & 0.464 \\
        & Instruction Following & 0.295 & 0.307 & 0.316 & 0.334 & 0.286 \\
        & Knowledge Update & 0.278 & 0.192 & 0.285 & 0.235 & 0.314 \\
        & Multi-Hop Reasoning & 0.214 & 0.104 & 0.227 & 0.192 & 0.247 \\
        & Preference Following & 0.571 & 0.553 & 0.450 & 0.547 & 0.465 \\
        & Summarization & 0.344 & 0.312 & 0.225 & 0.353 & 0.203 \\
        & Temporal Reasoning & 0.121 & 0.042 & 0.116 & 0.114 & 0.130 \\
        \cline{2-7}
        & Average & 0.316 & 0.256 & 0.295 & \textbf{0.320} & 0.294 \\
        \hline
        \multirow{11}{*}{1M} & Abstention & 0.500 & 0.664 & 0.600 & 0.557 & 0.507 \\
        & Contradiction Resolution & 0.021 & 0.021 & 0.035 & 0.042 & 0.032 \\
        & Event Ordering & 0.200 & 0.215 & 0.221 & 0.227 & 0.199 \\
        & Information Extraction & 0.366 & 0.246 & 0.391 & 0.397 & 0.366 \\
        & Instruction Following & 0.419 & 0.427 & 0.335 & 0.384 & 0.351 \\
        & Knowledge Update & 0.357 & 0.185 & 0.321 & 0.400 & 0.285 \\
        & Multi-Hop Reasoning & 0.209 & 0.129 & 0.227 & 0.221 & 0.169 \\
        & Preference Following & 0.551 & 0.602 & 0.536 & 0.597 & 0.540 \\
        & Summarization & 0.316 & 0.310 & 0.169 & 0.330 & 0.128 \\
        & Temporal Reasoning & 0.154 & 0.050 & 0.111 & 0.121 & 0.111 \\
        \cline{2-7}
        & Average & 0.309 & 0.285 & 0.295 & \textbf{0.328} & 0.269 \\
        \hline
        \multirow{11}{*}{10M} & Abstention & 0.550 & 0.800 & 0.650 & 0.650 & 0.600 \\
        & Contradiction Resolution & 0.012 & 0.000 & 0.012 & 0.000 & 0.000 \\
        & Event Ordering & 0.197 & 0.199 & 0.199 & 0.209 & 0.181 \\
        & Information Extraction & 0.350 & 0.000 & 0.200 & 0.150 & 0.200 \\
        & Instruction Following & 0.350 & 0.175 & 0.175 & 0.175 & 0.050 \\
        & Knowledge Update & 0.275 & 0.050 & 0.300 & 0.150 & 0.225 \\
        & Multi-Hop Reasoning & 0.125 & 0.000 & 0.125 & 0.125 & 0.075 \\
        & Preference Following & 0.308 & 0.191 & 0.241 & 0.200 & 0.175 \\
        & Summarization & 0.220 & 0.119 & 0.068 & 0.008 & 0.050 \\
        & Temporal Reasoning & 0.000 & 0.000 & 0.050 & 0.075 & 0.000 \\
        \cline{2-7}
        & Average & \textbf{0.238} & 0.153 & 0.202 & 0.181 & 0.155 \\
        \hline
    \end{tabular}}
    \label{tab:ablation_study}
\end{table}

\subsection{Retrieval Budget}
\label{sec:retrieval_budget}
We investigate the impact of the retrieval budget through two sets of experiments: (i) varying the retrieval depth by setting the number of retrieved documents $K \in \{5, 10, 15, 20\}$, and (ii) comparing a dense retriever against a sparse retriever (SPLADE).

The full results examining the effect of different retrieval depths (number of retrieved documents) are presented in Table~\ref{tab:document_numbers_effect}.

\begin{table}[H]
    \centering
    \caption{Effect of retrieval depth on performance across conversation lengths (100K–10M) and memory abilities. Results are shown for different numbers of retrieved documents ($K \in \{5, 10, 15, 20\}$).}
    \adjustbox{max width=\textwidth}{
    \begin{tabular}{c|c|c|c|c|c}
        \hline
        Length & Memory Ability & K=5 & K=10 & K=15 & K=20 \\
        \hline
        \multirow{11}{*}{100K} & Abstention & 0.475 & 0.500 & 0.625 & 0.625 \\
        & Contradiction Resolution & 0.037 & 0.025 & 0.025 & 0.031 \\
        & Event Ordering & 0.205 & 0.191 & 0.218 & 0.210 \\
        & Information Extraction & 0.479 & 0.450 & 0.412 & 0.391 \\
        & Instruction Following & 0.362 & 0.362 & 0.475 & 0.462 \\
        & Knowledge Update & 0.362 & 0.375 & 0.350 & 0.300 \\
        & Multi-Hop Reasoning & 0.281 & 0.322 & 0.321 & 0.309 \\
        & Preference Following & 0.566 & 0.591 & 0.562 & 0.575 \\
        & Summarization & 0.232 & 0.231 & 0.218 & 0.213 \\
        & Temporal Reasoning & 0.112 & 0.162 & 0.137 & 0.137 \\
        \cline{2-6}
        & Average & 0.311 & 0.321 & \textbf{0.334} & 0.325 \\
        \hline
        \multirow{11}{*}{500K} & Abstention & 0.571 & 0.514 & 0.614 & 0.642 \\
        & Contradiction Resolution & 0.017 & 0.021 & 0.071 & 0.071 \\
        & Event Ordering & 0.244 & 0.229 & 0.238 & 0.247 \\
        & Information Extraction & 0.506 & 0.531 & 0.503 & 0.507 \\
        & Instruction Following & 0.295 & 0.341 & 0.390 & 0.373 \\
        & Knowledge Update & 0.278 & 0.307 & 0.326 & 0.326 \\
        & Multi-Hop Reasoning & 0.214 & 0.188 & 0.234 & 0.213 \\
        & Preference Following & 0.571 & 0.597 & 0.628 & 0.607 \\
        & Summarization & 0.344 & 0.354 & 0.375 & 0.376 \\
        & Temporal Reasoning & 0.121 & 0.128 & 0.121 & 0.135 \\
        \cline{2-6}
        & Average & 0.316 & 0.321 & \textbf{0.350} & 0.350 \\
        \hline
        \multirow{11}{*}{1M} & Abstention & 0.500 & 0.521 & 0.600 & 0.585 \\
        & Contradiction Resolution & 0.021 & 0.021 & 0.057 & 0.053 \\
        & Event Ordering & 0.200 & 0.224 & 0.240 & 0.242 \\
        & Information Extraction & 0.366 & 0.398 & 0.377 & 0.391 \\
        & Instruction Following & 0.419 & 0.476 & 0.439 & 0.446 \\
        & Knowledge Update & 0.357 & 0.350 & 0.400 & 0.407 \\
        & Multi-Hop Reasoning & 0.209 & 0.189 & 0.209 & 0.190 \\
        & Preference Following & 0.551 & 0.596 & 0.535 & 0.514 \\
        & Summarization & 0.316 & 0.317 & 0.325 & 0.351 \\
        & Temporal Reasoning & 0.154 & 0.154 & 0.119 & 0.199 \\
        \cline{2-6}
        & Average & 0.309 & 0.325 & \textbf{0.330} & 0.330 \\
        \hline
        \multirow{11}{*}{10M} & Abstention & 0.550 & 0.600 & 0.650 & 0.600 \\
        & Contradiction Resolution & 0.012 & 0.012 & 0.025 & 0.025 \\
        & Event Ordering & 0.197 & 0.210 & 0.213 & 0.236 \\
        & Information Extraction & 0.350 & 0.150 & 0.300 & 0.300 \\
        & Instruction Following & 0.350 & 0.150 & 0.450 & 0.400 \\
        & Knowledge Update & 0.275 & 0.200 & 0.300 & 0.300 \\
        & Multi-Hop Reasoning & 0.125 & 0.100 & 0.125 & 0.150 \\
        & Preference Following & 0.308 & 0.175 & 0.275 & 0.275 \\
        & Summarization & 0.220 & 0.089 & 0.196 & 0.164 \\
        & Temporal Reasoning & 0.000 & 0.025 & 0.000 & 0.000 \\
        \cline{2-6}
        & Average & 0.238 & 0.171 & \textbf{0.253} & 0.245 \\
        \hline
    \end{tabular}}
    \label{tab:document_numbers_effect}
\end{table}

In a complementary experiment, we analyzed the impact of retriever choice. Our base architecture employs a dense retriever, which we compare against the sparse SPLADE-V2 retriever \citep{formal2022spladev2}. As shown in Figure~\ref{fig:retriever_effect} in Appendix~\ref{sec:retrieval_budget}, SPLADE yields performance gains of 1.7\% at 100K tokens, 0.7\% at 500K, and 0.8\% at 1M, but results in a slight performance drop of 0.7\% at 10M. On average, the sparse retriever provides a modest improvement across conversation lengths. Complete results comparing the dense retriever with SPLADE are presented in Table~\ref{tab:retriever_effect}.

\begin{figure*}[h] 
   \includegraphics[width=\textwidth]{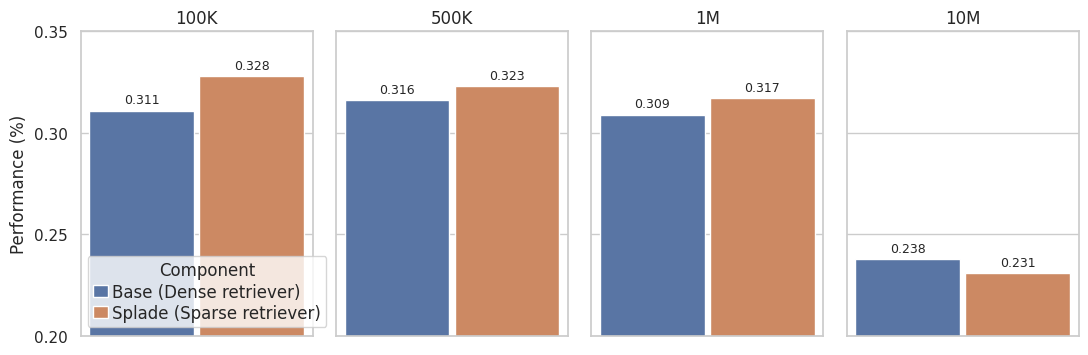}
    \caption{Performance comparison between dense retrieval and sparse retrieval (SPLADE) in \ourmethod.} 
    \label{fig:retriever_effect} 
\end{figure*}

\begin{table}[H]
    \centering
    \caption{Comparison of dense and sparse retrieval strategies across conversation lengths (100K–10M) and ten memory abilities. The table reports performance when using the default dense retriever versus a sparse retriever (SPLADE).}
    \adjustbox{max width=\textwidth}{
    \begin{tabular}{c|c|c|c}
        \hline
        Length & Memory Ability & Base (Dense retriever) & Sparse retriever (SPLADE) \\
        \hline
        \multirow{11}{*}{100K} & Abstention & 0.475 & 0.525 \\
        & Contradiction Resolution & 0.037 & 0.43 \\
        & Event Ordering & 0.205 & 0.181 \\
        & Information Extraction & 0.479 & 0.596 \\
        & Instruction Following & 0.362 & 0.400 \\
        & Knowledge Update & 0.362 & 0.350 \\
        & Multi-Hop Reasoning & 0.281 & 0.267 \\
        & Preference Following & 0.566 & 0.562 \\
        & Summarization & 0.232 & 0.230 \\
        & Temporal Reasoning & 0.112 & 0.125 \\
        \cline{2-4}
        & Average & 0.311 & \textbf{0.328} \\
        \hline
        \multirow{11}{*}{500K} & Abstention & 0.571 & 0.557 \\
        & Contradiction Resolution & 0.017 & 0.025 \\
        & Event Ordering & 0.244 & 0.226 \\
        & Information Extraction & 0.506 & 0.559 \\
        & Instruction Following & 0.295 & 0.345 \\
        & Knowledge Update & 0.278 & 0.307 \\
        & Multi-Hop Reasoning & 0.214 & 0.212 \\
        & Preference Following & 0.571 & 0.565 \\
        & Summarization & 0.344 & 0.330 \\
        & Temporal Reasoning & 0.121 & 0.107 \\
        \cline{2-4}
        & Average & 0.316 & 0.\textbf{323} \\
        \hline
        \multirow{11}{*}{1M} & Abstention & 0.500 & 0.564 \\
        & Contradiction Resolution & 0.021 & 0.028 \\
        & Event Ordering & 0.200 & 0.196 \\
        & Information Extraction & 0.366 & 0.392 \\
        & Instruction Following & 0.419 & 0.401 \\
        & Knowledge Update & 0.357 & 0.371 \\
        & Multi-Hop Reasoning & 0.209 & 0.193 \\
        & Preference Following & 0.551 & 0.595 \\
        & Summarization & 0.316 & 0.300 \\
        & Temporal Reasoning & 0.154 & 0.133 \\
        \cline{2-4}
        & Average & 0.309 & \textbf{0.317} \\
        \hline
        \multirow{11}{*}{10M} & Abstention & 0.550 & 0.700 \\
        & Contradiction Resolution & 0.012 & 0.000 \\
        & Event Ordering & 0.197 & 0.202 \\
        & Information Extraction & 0.350 & 0.350 \\
        & Instruction Following & 0.350 & 0.250 \\
        & Knowledge Update & 0.275 & 0.375 \\
        & Multi-Hop Reasoning & 0.125 & 0.125 \\
        & Preference Following & 0.308 & 0.200 \\
        & Summarization & 0.220 & 0.090 \\
        & Temporal Reasoning & 0.000 & 0.025 \\
        \cline{2-4}
        & Average & \textbf{0.238} & 0.231 \\
        \hline
    \end{tabular}}
    \label{tab:retriever_effect}
\end{table}

\subsection{Effect of Embedding Model Choice}
\label{sec:embedding_model_effect}
We also examined how the choice of embedding model affects the performance of both the RAG baseline and the episodic memory component of \ourmethod. In the primary experiments, we used the \textit{BAAI/bge-small-en-v1.5} embedding model. To assess robustness, we re-ran all experiments using the larger \textit{BAAI/bge-large-en-v1.5} model while keeping the LLM reader fixed to \textit{GPT-4.1-nano}. As shown in Table~\ref{tab:embedding_model_effect}, \ourmethod consistently outperforms the RAG baseline under both embedding configurations. Moreover, \ourmethod exhibits larger gains when switching to the higher-capacity embedding model, achieving an additional 2.08\% improvement at the 1M-token setting and 16.37\% at the 10M-token setting. These results indicate that \ourmethod is robust to changes in embedding quality and can effectively leverage stronger embedding models to enhance long-term memory abilities.

\begin{table}[H]
    \centering
    \caption{Effect of the embedding model on performance across conversation lengths (100K–10M) and ten memory abilities. Results are shown for the RAG baseline and \ourmethod using two different embedding models.}
    \adjustbox{max width=\textwidth}{
    \begin{tabular}{c|c|c|c|c|c}
        \hline
        Length & Memory Ability & RAG (bge-small) & Ours (bge-small) & RAG (bge-large) & Ours (bge-large) \\
        \hline
        \multirow{11}{*}{100K} & Abstention & 0.800 & 0.575 & 0.825 & 0.600 \\
        & Contradiction Resolution & 0.018 & 0.031 & 0.012 & 0.031 \\
        & Event Ordering & 0.169 & 0.177 & 0.185 & 0.171 \\
        & Information Extraction & 0.362 & 0.538 & 0.404 & 0.562 \\
        & Instruction Following & 0.350 & 0.400 & 0.337 & 0.462 \\
        & Knowledge Update & 0.375 & 0.375 & 0.325 & 0.375 \\
        & Multi-Hop Reasoning & 0.263 & 0.365 & 0.224 & 0.341 \\
        & Preference Following & 0.550 & 0.625 & 0.537 & 0.562 \\
        & Summarization & 0.083 & 0.202 & 0.089 & 0.148 \\
        & Temporal Reasoning & 0.125 & 0.162 & 0.112 & 0.162 \\
        \cline{2-6}
        & Average & 0.309 & \textbf{0.345} & 0.305 & 0.341 \\
        \hline
        \multirow{11}{*}{500K} & Abstention & 0.828 & 0.600 & 0.814 & 0.571 \\
        & Contradiction Resolution & 0.025 & 0.035 & 0.028 & 0.032 \\
        & Event Ordering & 0.180 & 0.204 & 0.178 & 0.202 \\
        & Information Extraction & 0.382 & 0.491 & 0.345 & 0.454 \\
        & Instruction Following & 0.286 & 0.342 & 0.303 & 0.363 \\
        & Knowledge Update & 0.288 & 0.240 & 0.380 & 0.321 \\
        & Multi-Hop Reasoning & 0.233 & 0.266 & 0.272 & 0.282 \\
        & Preference Following & 0.577 & 0.684 & 0.571 & 0.650 \\
        & Summarization & 0.184 & 0.334 & 0.153 & 0.316 \\
        & Temporal Reasoning & 0.161 & 0.154 & 0.126 & 0.126 \\
        \cline{2-6}
        & Average & 0.314 & \textbf{0.335} & 0.317 & 0.331 \\
        \hline
        \multirow{11}{*}{1M} & Abstention & 0.778 & 0.678 & 0.771 & 0.657 \\
        & Contradiction Resolution & 0.028 & 0.021 & 0.021 & 0.025 \\
        & Event Ordering & 0.179 & 0.211 & 0.194 & 0.211 \\
        & Information Extraction & 0.399 & 0.410 & 0.360 & 0.439 \\
        & Instruction Following & 0.271 & 0.394 & 0.269 & 0.421 \\
        & Knowledge Update & 0.342 & 0.392 & 0.371 & 0.378 \\
        & Multi-Hop Reasoning & 0.293 & 0.278 & 0.204 & 0.254 \\
        & Preference Following & 0.513 & 0.576 & 0.497 & 0.598 \\
        & Summarization & 0.152 & 0.290 & 0.116 & 0.296 \\
        & Temporal Reasoning & 0.064 & 0.107 & 0.119 & 0.150 \\
        \cline{2-6}
        & Average & 0.302 & 0.336 & 0.292 & \textbf{0.343} \\
        \hline
        \multirow{11}{*}{10M} & Abstention & 0.650 & 0.400 & 0.800 & 0.550 \\
        & Contradiction Resolution & 0.012 & 0.025 & 0.025 & 0.037 \\
        & Event Ordering & 0.201 & 0.173 & 0.203 & 0.171 \\
        & Information Extraction & 0.300 & 0.350 & 0.300 & 0.450 \\
        & Instruction Following & 0.175 & 0.250 & 0.175 & 0.275 \\
        & Knowledge Update & 0.325 & 0.300 & 0.325 & 0.300 \\
        & Multi-Hop Reasoning & 0.091 & 0.135 & 0.066 & 0.075 \\
        & Preference Following & 0.366 & 0.425 & 0.316 & 0.525 \\
        & Summarization & 0.063 & 0.179 & 0.100 & 0.224 \\
        & Temporal Reasoning & 0.000 & 0.025 & 0.000 & 0.025 \\
        \cline{2-6}
        & Average & 0.218 & 0.226 & 0.231 & \textbf{0.263} \\
        \hline
    \end{tabular}}
    \label{tab:embedding_model_effect}
\end{table}

\subsection{Effect of Indexing Setup}
\label{sec:indexing_setup_effect}
We also investigated the effect of the vector database indexing setup on the performance of \ourmethod. In the primary experiments, we used \textit{IndexFlatIP}, and in the experiments below, we examined the effect of switching the indexing setup to \textit{IndexHNSWFlat}. For this experiment, the reader LLM was GPT-4.1-nano. The results are shown in Table~\ref{tab:indexing_setup_effect}.

\begin{table}[H]
    \centering
    \caption{Effect of vector database indexing setup on performance across conversation lengths (100K–10M) and ten memory abilities. Results are reported for \ourmethod under two different indexing configurations.}
    \adjustbox{max width=\textwidth}{
    \begin{tabular}{c|c|c|c}
        \hline
        Length & Memory Ability & Ours (IndexFlatIP) & Ours (IndexHNSWFlat) \\
        \hline
        \multirow{11}{*}{100K} 
        & Abstention & 0.575 & 0.600 \\
        & Contradiction Resolution & 0.031 & 0.031 \\
        & Event Ordering & 0.177 & 0.173 \\
        & Information Extraction & 0.538 & 0.565 \\
        & Instruction Following & 0.400 & 0.375 \\
        & Knowledge Update & 0.375 & 0.400 \\
        & Multi-Hop Reasoning & 0.365 & 0.285 \\
        & Preference Following & 0.625 & 0.662 \\
        & Summarization & 0.202 & 0.217 \\
        & Temporal Reasoning & 0.162 & 0.162 \\
        \cline{2-4}
        & Average & 0.345 & \textbf{0.347} \\
        \hline
        \multirow{11}{*}{500K} 
        & Abstention & 0.600 & 0.528 \\
        & Contradiction Resolution & 0.035 & 0.032 \\
        & Event Ordering & 0.204 & 0.207 \\
        & Information Extraction & 0.491 & 0.503 \\
        & Instruction Following & 0.342 & 0.332 \\
        & Knowledge Update & 0.240 & 0.226 \\
        & Multi-Hop Reasoning & 0.266 & 0.269 \\
        & Preference Following & 0.684 & 0.666 \\
        & Summarization & 0.334 & 0.317 \\
        & Temporal Reasoning & 0.154 & 0.176 \\
        \cline{2-4}
        & Average & \textbf{0.335} & 0.325 \\
        \hline
        \multirow{11}{*}{1M} 
        & Abstention & 0.678 & 0.578 \\
        & Contradiction Resolution & 0.021 & 0.025 \\
        & Event Ordering & 0.211 & 0.211 \\
        & Information Extraction & 0.410 & 0.420 \\
        & Instruction Following & 0.394 & 0.386 \\
        & Knowledge Update & 0.392 & 0.385 \\
        & Multi-Hop Reasoning & 0.278 & 0.278 \\
        & Preference Following & 0.576 & 0.567 \\
        & Summarization & 0.290 & 0.257 \\
        & Temporal Reasoning & 0.107 & 0.128 \\
        \cline{2-4}
        & Average & \textbf{0.336} & 0.324 \\
        \hline
        \multirow{11}{*}{10M} 
        & Abstention & 0.400 & 0.600 \\
        & Contradiction Resolution & 0.025 & 0.025 \\
        & Event Ordering & 0.173 & 0.168 \\
        & Information Extraction & 0.350 & 0.350 \\
        & Instruction Following & 0.250 & 0.300 \\
        & Knowledge Update & 0.300 & 0.225 \\
        & Multi-Hop Reasoning & 0.135 & 0.075 \\
        & Preference Following & 0.425 & 0.433 \\
        & Summarization & 0.179 & 0.194 \\
        & Temporal Reasoning & 0.025 & 0.000 \\
        \cline{2-4}
        & Average & 0.226 & \textbf{0.237} \\
        \hline
    \end{tabular}}
    \label{tab:indexing_setup_effect}
\end{table}

\subsection{Supplementary Baseline Evaluation}
\label{sec:read_agent_baseline}
Alongside long-context LLMs and RAG, we also evaluated ReadAgent \citep{lee2024human}, another method designed to enhance long-term memory in LLMs, on \ourdataset and compared it with \ourmethod. The results demonstrate that \ourmethod consistently outperforms ReadAgent across all four conversation lengths (100K, 500K, 1M, and 10M). The full results are shown in Table~\ref{tab:read_agent_baseline}.
\begin{table}[H]
    \centering
    \caption{Comparing \ourmethod with ReadAgent across conversation lengths (100K–10M) and ten memory abilities.}
    \adjustbox{max width=\textwidth}{
    \begin{tabular}{c|c|c|c}
        \hline
        Length & Memory Ability & ReadAgent & Ours (\ourmethod) \\
        \hline
        \multirow{11}{*}{100K} 
        & Abstention & 0.850 & 0.475 \\
        & Contradiction Resolution & 0.000 & 0.037 \\
        & Event Ordering & 0.200 & 0.205 \\
        & Information Extraction & 0.066 & 0.479 \\
        & Instruction Following & 0.237 & 0.362 \\
        & Knowledge Update & 0.150 & 0.362 \\
        & Multi-Hop Reasoning & 0.095 & 0.281 \\
        & Preference Following & 0.425 & 0.566 \\
        & Summarization & 0.045 & 0.232 \\
        & Temporal Reasoning & 0.000 & 0.112 \\
        \cline{2-4}
        & Average & 0.206 & \large\textbf{0.311} \\
        \hline
        \multirow{11}{*}{500K} 
        & Abstention & 0.928 & 0.571 \\
        & Contradiction Resolution & 0.007 & 0.017 \\
        & Event Ordering & 0.237 & 0.244 \\
        & Information Extraction & 0.047 & 0.506 \\
        & Instruction Following & 0.166 & 0.295 \\
        & Knowledge Update & 0.014 & 0.278 \\
        & Multi-Hop Reasoning & 0.022 & 0.214 \\
        & Preference Following & 0.386 & 0.571 \\
        & Summarization & 0.069 & 0.344 \\
        & Temporal Reasoning & 0.028 & 0.121 \\
        \cline{2-4}
        & Average & 0.191 & \large\textbf{0.316} \\
        \hline
        \multirow{11}{*}{1M} 
        & Abstention & 0.792 & 0.500 \\
        & Contradiction Resolution & 0.003 & 0.021 \\
        & Event Ordering & 0.211 & 0.200 \\
        & Information Extraction & 0.106 & 0.366 \\
        & Instruction Following & 0.166 & 0.419 \\
        & Knowledge Update & 0.014 & 0.357 \\
        & Multi-Hop Reasoning & 0.105 & 0.209 \\
        & Preference Following & 0.391 & 0.551 \\
        & Summarization & 0.041 & 0.316 \\
        & Temporal Reasoning & 0.033 & 0.154 \\
        \cline{2-4}
        & Average & 0.186 & \large\textbf{0.309} \\
        \hline
        \multirow{11}{*}{10M} 
        & Abstention & 0.750 & 0.550 \\
        & Contradiction Resolution & 0.000 & 0.012 \\
        & Event Ordering & 0.205 & 0.197 \\
        & Information Extraction & 0.000 & 0.350 \\
        & Instruction Following & 0.300 & 0.350 \\
        & Knowledge Update & 0.000 & 0.275 \\
        & Multi-Hop Reasoning & 0.000 & 0.125 \\
        & Preference Following & 0.166 & 0.308 \\
        & Summarization & 0.061 & 0.220 \\
        & Temporal Reasoning & 0.000 & 0.000 \\
        \cline{2-4}
        & Average & 0.148 & \large\textbf{0.238} \\
        \hline
    \end{tabular}}
    \label{tab:read_agent_baseline}
\end{table}

\newpage
\section{Nugget Design}
\label{sec:nugget_design}
In this section, we provide illustrative examples for each memory ability, demonstrating how nuggets are derived from the corresponding probing questions.

\begin{enumerate}[leftmargin=1.5em]
\item \label{item:abstention} \textbf{Abstention}
\begin{tcolorbox}[colback=gray!5, colframe=black!40, boxrule=0.3pt, sharp corners, breakable]
\textbf{Objective:} The correct behavior is to acknowledge that the requested information is not present in the provided conversation.  \\
\textbf{Rubric pattern:} Each atomic unit should be in this format:  
\emph{States that, based on the provided chat, there is no information about \texttt{<target topic>}} \\
\textbf{Example JSON:}  
\begin{lstlisting}[language=json, basicstyle=\ttfamily\scriptsize, breaklines=true]
{
    "question": "What specific advice did Manuel give about property management companies during the March 5 Investors Meetup?",
    "ideal_response": "Based on the provided chat, there is no information related to the specific advice Manuel gave about property management companies.",
    "source_chat_ids": {},
    "rubric": [
        "Based on the provided chat, there is no information related to the specific advice Manuel gave about property management companies."
    ]
}
\end{lstlisting}
\end{tcolorbox}

\item \label{item:contradiction_resolution} \textbf{Contradiction Resolution}  
\begin{tcolorbox}[colback=gray!5, colframe=black!40, boxrule=0.3pt, sharp corners, breakable]
\textbf{Objective:} Correct behavior is that the LLM should detect the contradiction and state both contradictory information while requesting clarification. \\ 
\textbf{Rubric pattern:}  
\begin{itemize}[noitemsep, topsep=0pt]
  \item States there is contradictory information.  
  \item Mentions claim \texttt{<A>}  
  \item Mentions claim \texttt{<B>}  
  \item Requests clarification about which statement is correct
\end{itemize} 
\textbf{Example JSON:}  
\begin{lstlisting}[language=json, basicstyle=\ttfamily\scriptsize, breaklines=true]
{
    "question": "Have I ever attended any real estate webinars or investor meetups?",
    "ideal_answer": "I notice you've mentioned contradictory information about this. You said you have never attended any real estate webinars or investor meetups, but you also mentioned attending a webinar about Turkey's rising demand for multi-family rentals. Which statement is correct?",
    "source_chat_ids": {
        "first_statement": [
        ],
        "second_statement": [
        ]
    },
    "rubric": [
        "LLM response should state: there is contradictory information",
        "LLM response should mention: You said you have never attended any real estate webinars or investor meetups",
        "LLM response should mention: you also mentioned attending a webinar about Turkey\u2019s rising demand for multi-family rentals",
        "LLM response should mention: which statement is correct?"
    ]
}
\end{lstlisting}
\end{tcolorbox}

\item \label{item:event_ordering} \textbf{Event Ordering}  
\begin{tcolorbox}[colback=gray!5, colframe=black!40, boxrule=0.3pt, sharp corners, breakable]
\textbf{Objective:} Correct behavior is the model lists a sequence of events/topics in the correct chronological order. \\
\textbf{Rubric pattern:}  
\begin{itemize}[noitemsep, topsep=0pt]
  \item LLM response should mention: \texttt{<event 1>}  
  \item \ldots  
  \item LLM response should mention: \texttt{<event N>}
\end{itemize}
\textbf{Example JSON:}  
\begin{lstlisting}[language=json, basicstyle=\ttfamily\scriptsize, breaklines=true]
{
    "question": "How did my focus on different aspects of property investment and management develop throughout our conversations in order? Mention ONLY and ONLY ten items.",
    "answer": "Your focus on property investment and management developed in this sequence: 1) Initial engagement with the local agent and preparation for property viewings, 2) Evaluation of property financials including ROI and rental income potential, 3) Exploration of financing options and mortgage concerns, 4) Handling contractor performance and repair negotiations, 5) Understanding zoning regulations and permit requirements for property conversions, 6) Planning and prioritizing renovations and investment risks for multi-family properties, 7) Detailed preparation for property viewings involving both agent and contractor, 8) Negotiation strategies and cash flow implications related to repair costs, 9) Final renovation project completion steps and portfolio diversification strategies, 10) Consideration of commercial property types and location factors for long-term investment.",
    "ordering_tested": [
        "1st: Agent interaction and viewing preparation",
        "2nd: Property financial evaluation",
        "3rd: Financing and mortgage concerns",
        "4th: Contractor management",
        "5th: Zoning and permits",
        "6th: Renovation planning and investment risks",
        "7th: Viewing preparation with agent and contractor",
        "8th: Repair cost negotiation and cash flow",
        "9th: Renovation completion and portfolio diversification",
        "10th: Commercial property and location considerations"
    ],
    "source_chat_ids": [],
    "rubric": [
        "LLM response should mention: Agent interaction and viewing preparation",
        "LLM response should mention: Property financial evaluation",
        "LLM response should mention: Financing and mortgage concerns",
        "LLM response should mention: Contractor management",
        "LLM response should mention: Zoning and permits",
        "LLM response should mention: Renovation planning and investment risks",
        "LLM response should mention: Viewing preparation with agent and contractor",
        "LLM response should mention: Repair cost negotiation and cash flow",
        "LLM response should mention: Renovation completion and portfolio diversification",
        "LLM response should mention: Commercial property and location considerations",
        "Presents the events in the correct chronological order"
    ]
}
\end{lstlisting}
\end{tcolorbox}

\item \label{item:information_extraction} \textbf{Information Extraction}  
\begin{tcolorbox}[colback=gray!5, colframe=black!40, boxrule=0.3pt, sharp corners, breakable]
\textbf{Objective:} LLM should answer the questioned facts correctly. \\ 
\textbf{Rubric pattern:}  
\begin{itemize}[noitemsep, topsep=0pt]
  \item Instantiate one criterion per fact directly from the ideal answer, using the stem ``LLM response should state/mention:''
\end{itemize} 
\textbf{Example JSON:}  
\begin{lstlisting}[language=json, basicstyle=\ttfamily\scriptsize, breaklines=true]
{
    "question": "What amount did I say I had set aside as my initial capital for investing?",
    "ideal_answer": "You said you had set aside $50,000 as your initial capital.",
    "source_chat_ids": [],
    "rubric": [
        "LLM response should state: $50,000"
    ]
}
\end{lstlisting}
\end{tcolorbox}

\item \label{item:instruction_following} \textbf{Instruction Following}  
\begin{tcolorbox}[colback=gray!5, colframe=black!40, boxrule=0.3pt, sharp corners, breakable]
\textbf{Objective:} LLM should adhere to format and/or content priorities stated in the conversation. \\ 
\textbf{Rubric pattern:}  
\begin{itemize}[noitemsep, topsep=0pt]
  \item Use \texttt{instruction\_being\_tested} (the explicit instruction) and decompose \texttt{expected\_compliance} into atomic criteria
\end{itemize} 
\textbf{Example JSON:}  
\begin{lstlisting}[language=json, basicstyle=\ttfamily\scriptsize, breaklines=true]
{
    "question": "How should I allocate my funds for the upcoming quarter?",
    "instruction_being_tested": "Always provide detailed financial breakdowns when I ask about budgeting decisions.",
    "expected_compliance": "Response should include itemized costs, specific amounts for different categories, and detailed breakdown rather than just a total estimate",
    "source_chat_ids": [],
    "rubric": [
        "LLM response should contain: itemized list of costs",
        "LLM response should contain: category-by-category breakdown",
        "LLM response should contain: detailed cost analysis"
    ]
}
\end{lstlisting}
\end{tcolorbox}

\item \label{item:knowledge_update} \textbf{Knowledge Update}  
\begin{tcolorbox}[colback=gray!5, colframe=black!40, boxrule=0.3pt, sharp corners, breakable]
\textbf{Objective:} LLM must reflect updated values when prior values have changed over time. \\ 
\textbf{Rubric pattern:}  
\begin{itemize}[noitemsep, topsep=0pt]
  \item Derive criteria from the ideal answer, using the stem ``LLM response should state/mention:'' for the updated value(s).
\end{itemize} 
\textbf{Example JSON:}  
\begin{lstlisting}[language=json, basicstyle=\ttfamily\scriptsize, breaklines=true]
{
    "question": "How much initial capital have I allocated for real estate investing?",
    "answer": "$60,000",
    "source_chat_ids": {
        "original_info": [

        ],
        "updated_info": [
        ]
    },
    "rubric": [
        "LLM response should state: $60,000"
    ]
}
\end{lstlisting}
\end{tcolorbox}

\item \label{item:Multi-hop_reasoning} \textbf{Multi-hop Reasoning}  
\begin{tcolorbox}[colback=gray!5, colframe=black!40, boxrule=0.3pt, sharp corners, breakable]
\textbf{Objective:} LLM must aggregate or compare information spanning multiple sessions. \\ 
\textbf{Rubric pattern:}  
\begin{itemize}[noitemsep, topsep=0pt]
  \item Instantiate criteria from the ideal answer for each required intermediate or aggregated fact.
\end{itemize} 
\textbf{Example JSON:}  
\begin{lstlisting}[language=json, basicstyle=\ttfamily\scriptsize, breaklines=true]
{
    "question": "How many different banks did I consider for my mortgage options across my conversations?",
    "answer": "Two banks: Halkbank and Ziraat Bank.",
    "source_chat_ids": [],
    "rubric": [
        "LLM response should state: Two banks",
        "LLM response should state: Halkbank",
        "LLM response should state: Ziraat Bank"
    ]
}
\end{lstlisting}
\end{tcolorbox}

\item \label{item:preference_following} \textbf{Preference Following}  
\begin{tcolorbox}[colback=gray!5, colframe=black!40, boxrule=0.3pt, sharp corners, breakable]
\textbf{Objective:} LLM must generate content consistent with user-specified preferences. \\ 
\textbf{Rubric pattern:}  
\begin{itemize}[noitemsep, topsep=0pt]
  \item Use \texttt{preference\_being\_tested} (the user's stated preference) and decompose \texttt{expected\_compliance} into atomic criteria.
\end{itemize} 
\textbf{Example JSON:}  
\begin{lstlisting}[language=json, basicstyle=\ttfamily\scriptsize, breaklines=true]
{
    "question": "I'm looking at several properties and want to understand how to approach comparing them effectively. What steps would you suggest I take?",
    "preference_being_tested": "I prefer methodical, data-driven decisions over impulsive purchases, reflecting my analytical nature.",
    "expected_compliance": "Response should outline a structured, step-by-step approach that involves gathering and analyzing relevant data before making a choice, rather than suggesting quick or impulsive actions.",
    "source_chat_ids": [],
    "rubric": [
        "LLM response should contain: provides a clear, logical process",
        "LLM response should contain: emphasizes data collection and analysis",
        "LLM response should contain: breaks down comparison into measurable factors"
    ]
}
\end{lstlisting}
\end{tcolorbox}

\item \label{item:summarization} \textbf{Summarization}  
\begin{tcolorbox}[colback=gray!5, colframe=black!40, boxrule=0.3pt, sharp corners, breakable]
\textbf{Objective:} LLM must provide a comprehensive summary covering required content elements. \\ 
\textbf{Rubric pattern:}  
\begin{itemize}[noitemsep, topsep=0pt]
  \item Decompose \texttt{ideal\_summary} into atomic content units; use the stem ``LLM response should contain:''.
\end{itemize} 
\textbf{Example JSON:}  
\begin{lstlisting}[language=json, basicstyle=\ttfamily\scriptsize, breaklines=true]
{
    "question": "Can you give me a comprehensive summary of my journey and decision-making process around investing in rental properties, including how my budget, property choices, management considerations, and financing plans have developed over time?",
    "ideal_summary": "Your journey toward investing in rental properties began with an initial capital of $50,000, which you questioned as potentially insufficient for purchasing a property within 12 months. Early discussions highlighted the need to research local market conditions, down payment requirements, and additional costs like closing fees and renovations, revealing that typical investments might exceed your initial capital. You explored identifying good fixer-upper properties by learning to recognize signs such as structural issues and outdated features, emphasizing the importance of cost-benefit analysis for renovations. As your plans progressed, you weighed the pros and cons of investing close to your location versus elsewhere, balancing ease of management against market diversity and growth potential. You also considered the choice between single-family homes and multi-family units, analyzing factors like rental yield, management complexity, and investment scale, with examples showing similar yields but differing capital needs. Financing options were carefully compared, particularly between Halkbank and Ziraat Bank mortgages, focusing on interest rates, fees, and service quality to optimize costs. Throughout, you developed a step-by-step plan for purchasing your first rental property, including market research, budgeting, inspections, financing, and tenant management, with timelines to reduce anxiety and ensure readiness. This comprehensive process reflects a thoughtful evolution from initial capital concerns to detailed investment strategies, property evaluation, financing decisions, and management planning, all aimed at making informed, balanced real estate investment choices.",
    "source_chat_ids": [],
    "rubric": [
        "LLM response should contain: investing in rental properties began with an initial capital of $50,000",
        "LLM response should contain: Early discussions highlighted the need to research local market conditions, down payment requirements, and additional costs like closing fees",
        "LLM response should contain: You explored identifying good fixer-upper properties by learning to recognize signs such as structural issues and outdated features",
        "LLM response should contain: you weighed the pros and cons of investing close to your location versus elsewhere, balancing ease of management against market diversity and growth potential",
        "LLM response should contain: You also considered the choice between single-family homes and multi-family units, analyzing factors like rental yield, management complexity, and investment scale",
        "LLM response should contain: Financing options were carefully compared, particularly between Halkbank and Ziraat Bank mortgages, focusing on interest rates, fees, and service quality to optimize costs",
        "LLM response should contain: you developed a step-by-step plan for purchasing your first rental property, including market research, budgeting, inspections, financing, and tenant management"
    ]
}
\end{lstlisting}
\end{tcolorbox}

\item \label{item:temporal_reasoning} \textbf{Temporal Reasoning}  
\begin{tcolorbox}[colback=gray!5, colframe=black!40, boxrule=0.3pt, sharp corners, breakable]
\textbf{Objective:} LLM must compute or restate durations and timeline relations correctly. \\ 
\textbf{Rubric pattern:}  
\begin{itemize}[noitemsep, topsep=0pt]
  \item Derive criteria from the ideal answer, using the stem ``LLM response should state:''.
\end{itemize} 
\textbf{Example JSON:}  
\begin{lstlisting}[language=json, basicstyle=\ttfamily\scriptsize, breaklines=true]
{
    "question": "How many days are there between my first property viewing with Mehmet Yilmaz and the last one I scheduled?",
    "answer": "There are 2 days between the first property viewing on March 25 and the last one on March 27.",
    "calculation_required": "March 27 - March 25 = 2 days",
    "source_chat_ids": {
        "first_event": [],
        "second_event": []
    },
    "rubric": [
        "LLM response should state: 2 days",
        "LLM response should state: from March 25, 2024 till March 27, 2024"
    ]
}
\end{lstlisting}
\end{tcolorbox}

\end{enumerate}

\newpage
\section{Examples from Different Components of \ourdataset}
\label{sec:beam_different_stages_examples}
In this section, we provide illustrative examples of generating a chat in the \textit{coding} domain. 
Specifically, we include a representative \textit{chat seed} with its domain, title, theme, and subtopics, 
followed by the corresponding \textit{narratives}, where only a truncated set is shown for brevity. 
We then present the \textit{user profile} and the user's social \textit{relationships}. 
Next, we provide excerpts from the \textit{conversation plans}, showing only a subset of bullet points from each sub-plan while preserving their full descriptions to maintain clarity. 
Finally, we provide samples of the \textit{generated chat}, highlighting exchanges where the user shares or requests code, and including follow-up turns to demonstrate the naturalistic back-and-forth flow. 
Together, these examples illustrate how different components of \ourdataset interact to form coherent, long-context dialogues.

%===============================
% Chat Seed
%===============================
\begin{tcolorbox}[colback=gray!5, colframe=black!40, boxrule=0.3pt, sharp corners, breakable, title={Chat Seed}]
\textbf{Domain:} Coding \\
\textbf{Title:} Automating Social Media Posts with Python \\
\textbf{Theme:} Scheduling and posting content across multiple platforms \\
\textbf{Subtopics:}
\begin{itemize}[leftmargin=*, itemsep=1pt]
  \item Twitter API integration \quad\(\cdot\)\quad Facebook Graph API usage \quad\(\cdot\)\quad Instagram automation tools
  \item Scheduling with cron jobs / APScheduler
  \item Image and caption management; hashtag generation
  \item Error handling for failed posts; tracking engagement metrics
\end{itemize}
\end{tcolorbox}

%===============================
% Narratives (curated, de-duplicated)
%===============================
\begin{tcolorbox}[colback=gray!5, colframe=black!40, boxrule=0.3pt, sharp corners, breakable, title={Narratives (Truncated)}]
\textbf{Technical Problem-Solving:} Debugging Twitter OAuth/403/429; fixing hashtag validation; profiling scheduler bottlenecks. \\
\textbf{Learning \& Knowledge:} API docs comprehension (Twitter v2, Facebook Graph v12--15); best practices for Instagram automation; mastering cron/APScheduler. \\
\textbf{Progress \& Development:} Setting up Twitter/Facebook integrations; building Instagram tools; designing scheduling algorithms. \\
\textbf{Implementation:} Feature implementation and refactoring for efficiency; async migration; retry and backoff strategies. \\
\textbf{Framework \& Technology:} Python libraries (Tweepy, facebook-sdk, requests); APScheduler/cron; Redis; asyncio. \\
\textbf{Testing \& QA:} Unit/integration/E2E tests (pytest, Selenium); TDD for schedulers and hashtag rules. \\
\textbf{DevOps \& Deployment:} CI/CD (GitHub Actions), containerization (Docker), EC2 deployment, blue–green releases. \\
\textbf{Data:} PostgreSQL schemas, indices, ETL for engagement metrics, Redis caching. \\
\textbf{Integration \& APIs:} Webhooks, message queues (RabbitMQ), API Gateway, SNS/Lambda. \\
\textbf{Performance:} Caching, load balancing (HAProxy), CPU/memory targets, throughput goals. \\
\textbf{Security/Compliance:} OAuth, token rotation, TLS, GDPR. \\
\textbf{PM \& Workflow:} Sprints, reviews, documentation standards.
\end{tcolorbox}

%===============================
% User Profile
%===============================
\begin{tcolorbox}[colback=gray!5, colframe=black!40, boxrule=0.3pt, sharp corners, breakable, title={User Profile}]
\textbf{Name:} John Brooks \quad \textbf{Age:} 52 \quad \textbf{Gender:} Male \\
\textbf{Location:} Port Charles, Luxembourg \quad \textbf{Profession:} Secretary/Administrator

\medskip
\textbf{Personality:}     He is a pillar of his community, always ready to lend a helping hand and offer guidance when needed. With a strong sense of tradition and order, he values honesty and dedication, often taking on a mentorship role to help others. His diligent and efficient approach to planning and organization makes him a reliable asset to those around him. He has a warm and welcoming demeanor, always willing to open his heart and home to friends, loved ones, and neighbors. Despite his strong convictions, he believes in the power of hospitality and good manners, often going out of his way to make others feel supported and cared for. With a dry sense of humor and a quick wit, he can be entertaining to be around, but he's not afraid to speak his mind and challenge the status quo when necessary. His practical and responsible nature makes him a respected member of his community, and his ability to stay grounded and logical in stressful situations is a valuable asset to those around him.
\end{tcolorbox}

%===============================
% Relationships
%===============================
\begin{tcolorbox}[colback=gray!5, colframe=black!40, boxrule=0.3pt, sharp corners, breakable, title={Relationships}]
\textbf{Parents:} Elizabeth (74), Robert (76) \\
\textbf{Partner:} Shannon (48) \\
\textbf{Close Friends:} Taylor (51), Teresa (62), Thomas (44), Charles (56), Patricia (46) \\
\textbf{Acquaintances/Colleagues:} Wesley (26), Jason (59), Claudia (15), Janice (13), Dana (55)
\end{tcolorbox}

%===============================
% Conversation Plan (selected high-signal instances)
%===============================
\begin{tcolorbox}[colback=gray!5, colframe=black!40, boxrule=0.3pt, sharp corners, breakable, title={Conversation Plan (Only a few representative bullets from each sub-plan)}]

\textbf{Subplan 1 — March 1, 2024}
\begin{itemize}[leftmargin=*, itemsep=2pt]
  \item \textbf{Project Initialization:} I'm setting up a Python 3.10 environment with Tweepy v4.10.1 and Facebook SDK v3.1.0 for API integrations.
  \item \textbf{Security \& Compliance Labels: Authentication for Twitter API Integration:} Implemented OAuth 1.0a with environment variables \texttt{TWITTER\_API\_KEY} and \texttt{TWITTER\_API\_SECRET} securely stored.
  \item \textbf{Database \& Data Management Labels: Database Design for Social Media Posting:} Designed PostgreSQL 14 schema with tables for posts, platforms, and scheduling metadata.
  \item \textbf{User Instruction:} Always include exact API version numbers when I ask about integration details.
  \item \textbf{Logical Contradiction:} I have never registered a Twitter Developer account or created any Twitter app.
\end{itemize}

\medskip
\textbf{Subplan 2 — March 20, 2024}
\begin{itemize}[leftmargin=*, itemsep=2pt]
  \item \textbf{Technical Problem-Solving Labels: Debugging Twitter API Integration:} Fixed “403 Forbidden” error caused by missing media upload step before tweet creation.
  \item \textbf{Implementation \& Development Labels: Code Refactoring for Performance:} Refactored \texttt{twitter\_post.py} to async functions using \texttt{asyncio}, improved throughput by 30\%.
  \item \textbf{Security \& Compliance Labels: Authorization for Facebook Graph API:} Implemented OAuth 2.0 flow with refresh tokens stored encrypted using Fernet symmetric encryption.
  \item \textbf{Information Update:} The Instagram automation prototype sprint deadline was adjusted to April 5, 2024, to allow additional testing of media upload features.
\end{itemize}

\medskip
\textbf{Subplan 3 — April 5, 2024}
\begin{itemize}[leftmargin=*, itemsep=2pt]
  \item \textbf{Implementation \& Development Labels: Implementing Error Handling:} Added retry logic with exponential backoff for Instagram API 429 Too Many Requests errors.
  \item \textbf{Performance \& Optimization Labels: Caching Strategies for Image and Caption Management:} Implemented Redis caching for resized images, reducing image processing time from 800ms to 200ms.
  \item \textbf{Debugging \& Troubleshooting Labels: Incident Response for Social Media Automation:} Responded to March 30, 2024, outage caused by expired Instagram tokens, implemented alerting via Slack webhook.
\end{itemize}

\medskip
\textbf{Subplan 4 — April 20, 2024}
\begin{itemize}[leftmargin=*, itemsep=2pt]
  \item \textbf{Implementation \& Development Labels: Algorithm Optimization for Scheduling:} Rewrote scheduling algorithm to use async priority queues, reducing average job dispatch latency from 500ms to 150ms.
  \item \textbf{Framework \& Technology Labels: Integrating Twitter API with Python:} Upgraded Tweepy from v4.10.1 to v4.12.1 to leverage new media upload endpoints.
  \item \textbf{Security \& Compliance Labels: Authentication for Twitter API Integration:} Rotated Twitter API keys on April 15, 2024, updated environment variables \texttt{TWITTER\_API\_KEY} and \texttt{TWITTER\_API\_SECRET}.
\end{itemize}

\medskip
\textbf{Subplan 5 — May 5, 2024}
\begin{itemize}[leftmargin=*, itemsep=2pt]
  \item \textbf{Progress \& Development Labels: Building Hashtag Generation Tools:} Developed hashtag generator supporting dynamic keyword extraction using spaCy v3.5.0 NLP library.
  \item \textbf{Database \& Data Management Labels: Data Warehousing for Engagement Metrics:} Designed PostgreSQL 14 schema for engagement\_metrics with partitioning by month for scalability.
  \item \textbf{Debugging \& Troubleshooting Labels: Log Analysis for Facebook Graph API:} Detected “OAuthException: Error validating access token” on May 1, 2024, resolved by token refresh automation.
\end{itemize}

\medskip
\textbf{Subplan 6 — May 20, 2024}
\begin{itemize}[leftmargin=*, itemsep=2pt]
  \item \textbf{Implementation \& Development Labels: Implementing Error Handling:} Added centralized error handler middleware in posting API, logging errors with Sentry v1.12.0.
  \item \textbf{Debugging \& Troubleshooting Labels: Error Diagnosis for Twitter API Integration:} Fixed intermittent “ConnectionResetError” during media upload by adding retry with jitter.
  \item \textbf{DevOps \& Deployment Labels: Containerization for Instagram Automation:} Updated Dockerfile to use multi-stage builds, reduced image size from 120MB to 85MB.
\end{itemize}

\medskip
\textbf{Subplan 7 — June 5, 2024}
\begin{itemize}[leftmargin=*, itemsep=2pt]
  \item \textbf{DevOps \& Deployment Labels: Deploying Social Media Automation Tools:} Deployed v1.0.0 release on AWS EC2 t3.medium with 99.9\% uptime SLA.
  \item \textbf{Integration \& API Labels: Event-Driven Architecture for Social Media Automation:} Implemented AWS SNS topics for post status updates, integrated with Lambda v3.2.1 functions.
  \item \textbf{User Experience \& Interface Labels: Mobile App Design for Social Media Automation:} Released beta version of React Native app on Android with basic scheduling and metrics display.
\end{itemize}

\medskip
\textbf{Subplan 8 — June 20, 2024}
\begin{itemize}[leftmargin=*, itemsep=2pt]
  \item \textbf{Progress \& Development Labels: Developing Instagram Automation Tools:} Implemented batch media uploads for Instagram, supporting up to 10 images per carousel post.
  \item \textbf{User Experience \& Interface Labels: Responsive Design for Scheduling:} Enhanced React 18.2 dashboard for scheduling with drag-and-drop post reordering, tested on Chrome and Safari.
  \item \textbf{Security \& Compliance Labels: Encryption for Instagram Automation:} Migrated token storage to AWS KMS encrypted S3 bucket, access controlled via IAM roles.
\end{itemize}

\medskip
\textbf{Subplan 9 — July 5, 2024}
\begin{itemize}[leftmargin=*, itemsep=2pt]
  \item \textbf{Security \& Compliance Labels: Authentication for Twitter API Integration:} Implemented OAuth 2.0 PKCE flow for enhanced security, tested with Postman v10.15.0.
  \item \textbf{Performance \& Optimization Labels: Code Optimization for Twitter API Integration:} Reduced memory usage by 20\% by optimizing data serialization and connection pooling.
  \item \textbf{Debugging \& Troubleshooting Labels: Error Diagnosis for Twitter API Integration:} Fixed “TimeoutError” during high load by increasing HTTP client timeout from 10s to 30s.
\end{itemize}

\medskip
\textbf{Subplan 10 — July 20, 2024}
\begin{itemize}[leftmargin=*, itemsep=2pt]
  \item \textbf{DevOps \& Deployment Labels: Deploying Social Media Automation Tools:} Completed production rollout on July 18, 2024, with zero downtime and 99.95\% uptime in first 48 hours.
  \item \textbf{Integration \& API Labels: Event-Driven Architecture for Social Media Automation:} Fully implemented AWS SNS event bus with Lambda v3.2.1 for asynchronous post status updates.
  \item \textbf{Testing \& Quality Assurance Labels: End-to-End Testing for Instagram Automation:} Verified Instagram automation stability with 1000+ posts over 7 days, no failures reported.
\end{itemize}

\end{tcolorbox}

%===============================
% Generated Chat: Exemplars (Revised with mid-ellipsis reveals)
%===============================
\begin{tcolorbox}[colback=gray!5, colframe=black!40, boxrule=0.3pt,
                  sharp corners, breakable, title={Generated Dialogues (Exemplars, Truncated)}]

Turn 1 (Mar 1, 2024) \\
\textbf{User:}
I'm trying to set up a Python 3.10 environment for automating social media posts, and I've chosen Tweepy v4.10.1 and Facebook SDK v3.1.0 for API integrations, but I'm not sure how to structure my project for better outreach. Can you help me plan this out in terms of scheduling and deadlines?  

\medskip
\textbf{Assistant:} \\
Certainly! Setting up an automated posting system with Tweepy and Facebook SDK is a great start. \\ 
Project Structure: \\
1. Project Setup (Mar 1–7) \textcolor{red}{[Truncated ...]} \\
2. Content Creation (Mar 8–14) \textcolor{red}{[Truncated ...]} \\
3. Automation Development (Mar 15–21) \textcolor{red}{[Truncated ...]} \\
4. Deployment and Monitoring (Mar 22–28) \textcolor{red}{[Truncated ...]} \\

\medskip
\textbf{User (Follow-up)} \\
That sounds like a great plan! One thing though—could we add a step for testing posts on a smaller scale before full deployment? Just to make sure everything looks good and functions as expected.

\medskip
\textbf{Assistant} \\
Absolutely, adding a step for testing the posts on a smaller scale is a great idea to ensure everything looks good and functions as expected before full deployment. Here's how you can incorporate that into the existing plan: \\
\#\#\# Updated Project Structure \\
\#\#\#\# 1. **Project Setup \textcolor{red}{[Trunicated ...]} \\
\#\#\#\# 4. **Testing Phase (March 22 - March 24, 2024)**\\ - **Days 22-23:** Conduct a small-scale test.\\ - Post a few test messages to a limited audience or a test account.\\ - Verify that the posts look good and function as expected.\\ - **Day 24:** Review the test results.\\ - Address any issues found during the test phase.\\ - Make final adjustments to the automation script and content.\\\\ \textcolor{red}{[Trunicated ...]}

\noindent\rule{\linewidth}{0.4pt}

\medskip
\textbf{User (Mid-turn) \texttt{->} Code Sharing} \\
I'm trying to implement a feature to collect engagement metrics for Instagram posts, but I'm not sure how to integrate it with the existing PostgreSQL schema. I've designed a table for instagram\_posts with foreign key to posts table, but I'm not sure how to handle the data migration. Can you help me with that? Here's my code: \\
\begin{lstlisting}[basicstyle=\ttfamily\scriptsize, breaklines=true]
import psycopg2
conn = psycopg2.connect(dbname=\"engagement_metrics\",    user=\"username\",
password=\"password\",
host=\"localhost\",
port=\"5432\")
cur = conn.cursor()
cur.execute(\"\"\"
CREATE TABLE instagram_posts (id SERIAL PRIMARY KEY,        post_id INTEGER REFERENCES posts(id),        engagement_metrics JSONB);
\"\"\")\n\nconn.commit()
cur.close()
conn.close()
\end{lstlisting}

\medskip
\textbf{Assistant \texttt{->} Code Review} \\
Your approach is good! Consider adding \texttt{NOT NULL}, timestamps, and indexes for performance \dots  
\begin{lstlisting}[basicstyle=\ttfamily\scriptsize, breaklines=true]
CREATE TABLE IF NOT EXISTS instagram_posts (
  id SERIAL PRIMARY KEY,
  post_id INTEGER NOT NULL REFERENCES posts(id),
  engagement_metrics JSONB,
  created_at TIMESTAMP NOT NULL DEFAULT CURRENT_TIMESTAMP
);
CREATE INDEX IF NOT EXISTS idx_instagram_posts_post_id 
  ON instagram_posts(post_id);
\end{lstlisting}
\textcolor{red}{[Truncated ...]} plus safe migration with INSERT SELECT to backfill existing posts. \\

\noindent\rule{\linewidth}{0.4pt}

\medskip
\textbf{User \texttt{->} Deployment} \\
I'm trying to deploy my updated posting services to an AWS EC2 t3.medium instance with 2 vCPUs and 4GB RAM, but I'm running into some issues with the deployment process.
Can you help me figure out how to properly deploy my application to this instance, considering the DevOps and deployment aspects?
I've been using Docker to containerize my application, and I'm not sure what changes I need to make to ensure a smooth deployment.
Here's a snippet of my current Dockerfile:
\begin{lstlisting}[basicstyle=\ttfamily\scriptsize, breaklines=true]
FROM python:3.10-slim
WORKDIR /app
COPY requirements.txt .
RUN pip install -r requirements.txt
COPY . .
CMD ["python", "app.py"]
\end{lstlisting}

\medskip
\textbf{Assistant \texttt{->} Deployment Guidance} \\
Install Docker on EC2, push image to registry, pull with pinned tag, and run with  
\texttt{--restart=always} \textcolor{red}{[Truncated ...]} add healthchecks and a minimal \texttt{systemd} unit \textcolor{red}{[Truncated ...]}  

\medskip
\textbf{User (Follow-up)} \\
hmm, what's the best way to automate the pulling and running of the Docker container?  

\medskip
\textbf{Assistant \texttt{->} CI/CD Automation} \\
Use GitHub Actions with SSH deploy step \textcolor{red}{[Truncated ...]} 
\begin{lstlisting}[basicstyle=\ttfamily\scriptsize, breaklines=true]
name: Deploy
on: { push: { branches: [main] } }
jobs:
  deploy:
    runs-on: ubuntu-latest
    steps:
      - uses: actions/checkout@v4
      - run: |
          ssh -i id_rsa ec2-user@EC2_DNS \
          "docker stop socauto || true && docker rm socauto || true &&
           docker pull username/your-app:sha-${GITHUB_SHA} &&
           docker run -d --restart=always -p 80:80 username/your-app:sha-${GITHUB_SHA}"
\end{lstlisting}

\medskip
\textbf{User (Follow-up)} \\
hmm, what's the best way to handle environment variables for different stages like dev and prod?  

\medskip
\textbf{Assistant \texttt{->} Env Vars} \\
Use separate \texttt{.env.dev} / \texttt{.env.prod} files locally.  
In production, prefer AWS SSM Parameter St46ore or Secrets Manager \textcolor{red}{[Truncated ...]}  inject via \texttt{--env-file} or \texttt{systemd} environment exports.
\end{tcolorbox}

\section{Case Study}
\label{sec:scratchpad_case_study}
\noindent The scratchpad serves as a persistent, iteratively compressed semantic layer that (i) consolidates scattered entities and facts (improving \emph{Information Extraction}), (ii) preserves user-level meta-instructions alongside task context (improving \emph{Instruction Following}), (iii) records state changes and overrides (improving \emph{Information Update}), and (iv) normalizes timeline cues into explicit anchors (improving \emph{Temporal Reasoning}). Out of the ten memory abilities in our benchmark, we present case studies from these four; however, the scratchpad provides similar benefits for the remaining six abilities as well. Below we contrast ``\ourmethod with all components'' versus ``\ourmethod w/o scratchpad'' on representative probes, and show the exact scratchpad cue that enabled the correct answer.

\begin{tcolorbox}[colback=gray!5, colframe=black!40, boxrule=0.3pt, sharp corners, breakable, title={Information Extraction}]
\small
\textbf{(1) Question:} Which versions of the tools did I say I was using to start my project?\\
\textbf{Correct Answer:} Python 3.10, spaCy v3.5, Flask 2.2, PyMuPDF 1.22.0\\
\textbf{\ourmethod with all components:} Python 3.10, spaCy v3.5, Flask 2.2, PyMuPDF 1.22.0\\
\textbf{\ourmethod wo scratchpad:} Based solely on the provided context, there is no direct answer to this question \\
\textbf{Scratchpad cue (excerpt):} \\
**KEY ENTITIES \& RELATIONSHIPS:** \\
...... [trunecated] \\
- Technologies: Python 3.10, spaCy v3.5, Flask 2.2/2.2.3, PyMuPDF 1.22.0, scikit-learn v1.2.2, Redis, RabbitMQ, Docker  \\
...... [trunecated] \\

\vspace{6pt}
\noindent\rule{\linewidth}{0.4pt}
\vspace{6pt}

\textbf{(2) Question:} Can you remind me of the city and country I said I live in, along with my age?\\
\textbf{Correct Answer:} New Gary, Saint Helena; 26\\
\textbf{\ourmethod with all components:} New Gary, Saint Helena; 26\\
\textbf{\ourmethod wo scratchpad:} No information provided about your city, country, or age in the given context\\
\textbf{Scratchpad cue (excerpt):} \\
**KEY ENTITIES \& RELATIONSHIPS:** \\
- **Primary individuals:** User and partner Wendy (both turning 26 on March 3, 2024) \\
- **Friends and collaborators:** Joe (writer, 35), Stacy ...... [trunecated] \\
- **Locations:** New Gary, Saint Helena; Island Art Gallery; Saint Helena Book Fair; Saint Helena Community Center; GreenLeaf Nursery; Island Finds thrift store; Saint Helena Charity; The Lighthouse Bistro
\end{tcolorbox}

\begin{tcolorbox}[colback=gray!5, colframe=black!40, boxrule=0.3pt, sharp corners, breakable, title={Instruction Following}]
\small
\textbf{Question:} Can you tell me about the mentoring activities I've been involved in recently?\\
\textbf{User instruction:} Always provide the number of team members involved when I ask about mentoring activities.\\
\textbf{Correct Answer:} Mentions number of team members\\
\textbf{\ourmethod with all components:} Mentoring 3 junior editors\\
\textbf{\ourmethod wo scratchpad:} No recent mentoring activities noted in the provided context\\
\textbf{Scratchpad cue (excerpt):} \\
...... [trunecated] \\
The user requested to always include the number of team members when discussing mentoring activities. \\
...... [trunecated] \\
*Mentoring \& Crisis Management:* \\
- Mentoring 3 junior editors; balancing mentoring with crisis management. \\
...... [trunecated]
\end{tcolorbox}

\begin{tcolorbox}[colback=gray!5, colframe=black!40, boxrule=0.3pt, sharp corners, breakable, title={Knowledge Update}]
\small
\textbf{(1) Question:} What is the typical memory usage per session for batch resume processing?\\
\textbf{Correct Answer:} Around 140MB per session\\
\textbf{\ourmethod with all components:} \textasciitilde140MB\\
\textbf{\ourmethod wo scratchpad:} The typical memory usage per session for batch resume processing can vary widely depending on the complexity of the resumes and the system's available memory, but optimizing with batch sizes and efficient data handling can keep it under 100MB per resume.\\
\textbf{Scratchpad cue (excerpt):} \\
...... [trunecated] \\
- Memory optimization: \\
- Reduce peak memory usage from 250MB to ~140MB per batch. \\
...... [trunecated] \\

\noindent\rule{\linewidth}{0.4pt}

\textbf{(2) Question:} What is the deadline for completing the Instagram automation prototype sprint?\\
\textbf{Correct Answer:} April 5, 2024\\
\textbf{\ourmethod with all components:} April 5, 2024\\
\textbf{\ourmethod wo scratchpad:} April 1, 2024\\
\textbf{Scratchpad cue (excerpt):} \\
**PROCESSES \& WORKFLOWS:** \\
1. **Project Planning \& Sprint Management:** \\
...... [trunecated] \\
- Instagram automation sprint extended to April 5, 2024.
\end{tcolorbox}

\begin{tcolorbox}[colback=gray!5, colframe=black!40, boxrule=0.3pt, sharp corners, breakable, title={Temporal Reasoning}]
\small
\textbf{(1) Question:} How many days passed between when I met my mom at home and when I completed my first successful batch of Miche's sourdough?\\
\textbf{Correct Answer:} Jan 6 -- Jan 1 = 5 days\\
\textbf{\ourmethod with all components:} 5 days \\ \textbf{\ourmethod wo scratchpad:} 37 days\\
\textbf{Scratchpad cue (excerpt):} \\
**IMPORTANT DATES:** \\
- **January 1, 2024:** Met mother Brittney at home. \\
- **January 5-6, 2024:** Completed first batch of Michele's sourdough. \\
...... [trunecated]

\noindent\rule{\linewidth}{0.4pt}

\textbf{(2) Question:} How many days do I have between finalizing my survey plans at Caf\'e Soleil and attending the Raymondburgh Startup Meetup to prepare effectively?\\
\textbf{Correct Answer:} Mar 28 -- Mar 10 = 18 days\\
\textbf{\ourmethod with all components:} 18 days \\ \textbf{\ourmethod wo scratchpad:} 28 days\\
\textbf{Scratchpad cue (excerpt):} \\
**IMPORTANT DATES:** \\
- **March 10, 2024**: Paper-based customer survey at Cafe Soleil. \\
- **March 28, 2024**: Raymondburgh Startup Meetup. \\
...... [trunecated]

\end{tcolorbox}

\vspace{4pt}
\noindent\textbf{Takeaways.}
Across abilities, removing the scratchpad consistently causes failures that the full model avoids. In \emph{Information Extraction}, the scratchpad aggregates dispersed entity/version mentions so the model can recover exact tool versions and bios (city/age). For \emph{Instruction Following}, it retains user meta-preferences (e.g., ``always include team count''), ensuring style/format compliance even many turns later. For \emph{Knowledge Update}, it encodes overrides (e.g., extended deadline; reduced memory), preventing stale answers. For \emph{Temporal Reasoning}, it surfaces normalized date anchors, enabling simple, correct day-difference calculations. These examples show that the scratchpad provides a high-utility semantic scaffold that complements working (recency) and episodic (retrieval) memory, yielding robust long-context behavior.

\newpage
\section{Qualitative Error Analysis}
\label{sec:qualitative_error_analysis}
We conduct a qualitative analysis of failure cases across the ten memory abilities in our benchmark to better characterize the limitations of \ourmethod and identify systematic patterns. For each ability, we manually inspected probing questions that \ourmethod answered incorrectly and analyzed the underlying reasons. Below, we summarize the dominant error modes observed for each ability.

\paragraph{Abstention} In this ability, the LLM should abstain from answering because the answer to the probing question is not present in the conversation. Therefore, the context that \ourmethod provides to the LLM does not contain the required information. One failure mode occurs when the context contains nothing relevant to the question, yet the LLM hallucinates and generates an answer. This is because these LLMs are usually trained to always provide an answer, regardless of actually having this knowledge \citep{kalai2025language}. Another hallucination pattern occurs when the LLM produces an answer entirely unrelated to the question, which stems from the long-context nature of the task and the inability of the LLM to understand the context correctly. The main failure mode, however, arises when the context contains information about entities, dates, or concepts that are similar to, but not the same as, the information requested by the question. In these cases, the LLM uses these similar details and generates an answer instead of abstaining. This pattern is the primary failure mode for abstention.

\paragraph{Contradiction Resolution} 
For this ability, the LLM should identify the contradiction, state both sides, and request clarification. One common failure mode occurs when the context contains only one side of the contradiction, leading the LLM to answer based solely on that information. Since the model does not have access to the other side of the story, it cannot detect the conflict. Another common failure occurs when both sides of the contradiction are present in the context, but the LLM still overweighs one side of the contradiction due to position and frequency bias for that side in the context. 

\paragraph{Event Ordering} In this ability, the LLM should recognize and reconstruct the sequence of evolving information in the conversation. A common failure occurs when the context contains items from the sequence but the LLM does not include them in the response. Another failure mode occurs when the model includes the items but presents them in the wrong order. This happens because the retrieval model retrieves based on similarity, which does not necessarily preserve temporal order, leaving the LLM without clues about the correct sequence of events. Also, in many cases, the retriever does not retrieve all the events related to the question.

\paragraph{Information Extraction} In this ability, one failure mode arises when the context does not contain the answer to the question; thus, the LLM cannot extract the answer from the retrieved context. Another occurs when the answer is present, but the LLM produces an incorrect answer because it becomes confused by details in the context that are similar to the answer. A third failure happens when the answer is present but the LLM provides an incomplete answer.

\paragraph{Instruction Following} For this ability, the LLM should adhere to user-specified instructions. Failures occur either when the user instruction is present in the context, but the LLM does not follow it, or when the instruction is missing from the context, and the LLM cannot answer the question without it.

\paragraph{Knowledge Update} Here, the LLM should answer the question using the updated version of the facts. One failure mode occurs when the context contains only the old value and does not include the updated value, causing the LLM to respond using outdated information, because the retrieval model retrieves based on similarity, which does not necessarily preserve temporal order. A more common failure mode occurs when the context contains both the old and updated values, but the LLM still bases its answer on the old value, because the retrieved documents are not necessarily presented in the correct temporal order, again due to the retriever.

\paragraph{Multi-Hop Reasoning} For this ability, the LLM often fails when the context contains the necessary pieces of information but the model does not use them to answer the question. Another failure occurs when the context is missing some components required for the multi-hop reasoning chain, making it impossible for the LLM to answer correctly.

\paragraph{Preference Following} In this ability, the LLM should incorporate user-stated preferences into its answer. Failures occur when the context does not contain the user's preference and the model therefore answers without considering it, or when the preference is present but the LLM does not use it when generating the response.

\paragraph{Summarization} For this ability, one failure mode occurs when the context contains some components of the correct answer but the LLM fails to include them in the summary. Another failure arises when the context is missing some parts of the answer, which leads the LLM to omit those details as well.

\paragraph{Temporal Reasoning} In this ability, the LLM should reason about explicit and implicit temporal relations. A common failure mode occurs when the context contains the required dates, but the LLM becomes confused by another date (or dates) in the context and answers incorrectly. Another failure occurs when the context contains both dates but the LLM performs the arithmetic incorrectly, producing an answer with a numerical error. A third failure mode occurs when the context does not contain one of the necessary dates, causing the LLM to incorrectly substitute another date when answering.

\newpage
\section{Prompts}
\label{sec:prompts}
Here we provide the prompts used in different stages of our framework.

%%%% Probing Questions %%%% 
% Candidate selection % 
\lstinputlisting[backgroundcolor=\color{lightgray},breaklines=true,basicstyle=\tiny\ttfamily, label={information_extraction_selection}, caption={Candidate selection information extraction prompt}, captionpos=b]{Prompts/probing_questions/information\_extraction.tex}

\lstinputlisting[backgroundcolor=\color{lightgray},breaklines=true,basicstyle=\tiny\ttfamily, label={multi_hop_reasoning_selection}, caption={Candidate selection multi-hop reasoning prompt}, captionpos=b]{Prompts/probing_questions/multi\_hop\_reasoning.tex}

\lstinputlisting[backgroundcolor=\color{lightgray},breaklines=true,basicstyle=\tiny\ttfamily, label={knowledge_update_selection}, caption={Candidate selection knowledge update prompt}, captionpos=b]{Prompts/probing_questions/knowledge\_update.tex}

\lstinputlisting[backgroundcolor=\color{lightgray},breaklines=true,basicstyle=\tiny\ttfamily, label={temporal_reasoning_selection}, caption={Candidate selection temporal reasoning prompt}, captionpos=b]{Prompts/probing_questions/temporal\_reasoning.tex}

\lstinputlisting[backgroundcolor=\color{lightgray},breaklines=true,basicstyle=\tiny\ttfamily, label={preference_following_selection}, caption={Candidate selection preference following prompt}, captionpos=b]{Prompts/probing_questions/preference\_following.tex}

\lstinputlisting[backgroundcolor=\color{lightgray},breaklines=true,basicstyle=\tiny\ttfamily, label={event_ordering_selection}, caption={Candidate selection event ordering prompt}, captionpos=b]{Prompts/probing_questions/event\_ordering.tex}

\lstinputlisting[backgroundcolor=\color{lightgray},breaklines=true,basicstyle=\tiny\ttfamily, label={contradiction_resolution_selection}, caption={Candidate selection contradiction resolution prompt}, captionpos=b]{Prompts/probing_questions/contradiction\_resolution.tex}

\begin{lstlisting}[backgroundcolor=\color{lightgray},breaklines=true,basicstyle=\tiny\ttfamily, label={summarization_selection}, caption={Candidate selection summarization prompt}, captionpos=b]
This is a plan that contains detailed bullet points about a topic. This plan is used to generate realistic chat conversations between a user and an AI assistant, which are then used to evaluate the long-term memory capabilities of LLMs. 
Your task is to analyze this plan and select GROUPS of bullet points that would be most effective for testing summarization abilities when incorporated into chat conversations.
Summarization tests whether the LLM can synthesize and condense information from across multiple conversation sessions into coherent, comprehensive summaries.
Analyze this plan and identify GROUPS of 8-12 related bullet points that represent topics suitable for summarization testing. Groups can vary in size depending on the richness and complexity of the topic.

## INPUT DATA
- **PLAN**: <plan>

## CRITICAL REQUIREMENT: EARLY BATCH PRIORITIZATION
**SELECTION PRIORITY ORDER:**
1. **Groups starting in Batches 1-3 (HIGHEST PRIORITY)**: Select 60-70% of your groups with foundational content from early batches
2. **Groups spanning early to middle batches (MEDIUM PRIORITY)**: Select 20-30% of groups that bridge early-to-middle timeline
3. **Groups from later batches only (LOW PRIORITY)**: Select only 10-20% from purely later batches

## CRITICAL REQUIREMENT: CONTENT-BASED ANALYSIS
**ANALYZE BULLET CONTENT, NOT CATEGORY NAMES:**
- Read the actual bullet point text to identify mentions of entities (people, places, items, topics, amounts, processes)
- The same entity might appear in different category types - include mentions across all categories
- The same process might appear in different category types - include mentions across all categories
- The same project might appear in different category types - include mentions across all categories

**SEARCH METHODOLOGY:**
1. **Identify key entities** in bullet content: names, places, amounts, equipment, topics, processes
2. **Search ALL batches** for any mention of these entities in ANY category
3. **Group by content similarity**, not category similarity  
4. **Include 8-12 mentions** regardless of how they're categorized

Focus on complete topic clusters that enable summarization of:
- **Entity or Relationship Histories**: interactions, developments, or changes related to a specific person, organization, group, or other identifiable entity across the entire plan (complete relationship or entity arc)
- **End-to-End Processes**: steps, stages, or phases of a specific process, workflow, or methodology from initiation to conclusion across the entire plan (no missing steps)
- **Resource or Asset Lifecycles**: mentions of acquisition, allocation, usage, modification, and outcomes for a specific resource, asset, or material across the entire plan
- **Decision and Strategy Journeys**: details of decision-making, planning, and strategy development from problem identification through implementation across the entire plan
- **Problem/Challenge Resolution Narratives**: instances of identifying, analyzing, addressing, and resolving a particular issue or challenge across the entire plan
- **Timeline-Driven Developments**: events and updates showing chronological evolution of a specific project, initiative, or topic across the entire plan
- **Knowledge or Skill Development Sequences**: progress updates, milestones, and learning activities related to acquiring or improving a specific skill or knowledge area across the entire plan
- **Discussion and Agreement Processes**: discussions, debates, negotiations, and agreements related to a specific matter across the entire plan

Prioritize bullet point groups that:
- Contain rich, interconnected information suitable for synthesis
- Enable questions like "Can you summarize my interactions with X?" or "Summarize the [X] process"
- Include both factual/quantitative details and qualitative/narrative elements for well-rounded summaries
- Have varying complexity levels (simple single-topic vs. complex multi-faceted stories)
- Allow testing of information condensation across multiple conversation sessions
- Include both factual details and narrative elements for comprehensive summarization
- Create opportunities to test synthesis of scattered information into coherent narratives

Return your analysis in this exact JSON format where each object contains 8-12 related bullet points:
[{"capability": "summarization", "batch_numbers": "1, 1, 2, 3, 4, 5", "bullet_numbers": "5, 22, 18, 19, 23, 31",
        "bullet_points": "• **Character & Relationship:Close Friend:** Jeremiah, 37, met at festival, recommended consultant. | • **Conflict & Resolution:Relationship Boundaries:** Jeremiah requested exclusive studio access; agreed to 3 hours only. | • **Character & Relationship:Close Friend:** Jeremiah helped install MLV, bringing snacks from bakery. | • **Character & Relationship:Close Friend:** Jeremiah invited me to music club to test prototype. | • **Character & Relationship:Close Friend:** Jeremiah brought dinner during late work session. | • **Goals & Progress:Milestone Celebration:** Hosted listening party with Jeremiah, Jon, and Tonya."}
]

Important formatting notes:
- The "batch_numbers" and "bullet_numbers" correspond to each other positionally
- "1, 1, 2, 3, 4, 5" and "5, 22, 18, 19, 23, 31" means: Batch 1 Bullet 5, Batch 1 Bullet 22, Batch 2 Bullet 18, Batch 3 Bullet 19, Batch 4 Bullet 23, Batch 5 Bullet 31
- Each bullet point separated by " | "
- Use comma-separated values for batch_number and bullet_number
- Vary group sizes based on topic complexity and richness
- Focus on groups that enable comprehensive summarization questions
- Include both simple single-topic and complex multi-topic groups
- **NO LIMIT on number of bullet points** - include as many as needed for complete coverage

CRITICAL NOTES: 
- **ANALYZE BULLET CONTENT, NOT CATEGORY NAMES**: Search for mentions of entities in the actual text
- **IGNORE CATEGORY LABELS**: The same entity mentioned in different category types should all be grouped together
- **COMPREHENSIVE ENTITY SEARCH**: For each entity/topic/process, scan ALL batches and ALL categories for any mention
- Include 8-12 mentions regardless of bullet point category if they reference the same entity/topic in the content

Select 7-9 groups of bullet points with COMPLETE mention coverage that would enable the most sophisticated and comprehensive summarization questions across different complexity levels.

NOTE: Only output the list without any explanation before or after the list.
\end{lstlisting}

\lstinputlisting[backgroundcolor=\color{lightgray},breaklines=true,basicstyle=\tiny\ttfamily, label={instruction_following_selection}, caption={Candidate selection instruction following prompt}, captionpos=b]{Prompts/probing_questions/instruction\_following.tex}

% Question generation % 
\lstinputlisting[backgroundcolor=\color{lightgray},breaklines=true,basicstyle=\tiny\ttfamily, label={information_extraction_probe_generation}, caption={Information extraction probing question generation prompt}, captionpos=b]{Prompts/probing_questions/information\_extraction\_hard\_generation.tex}

\lstinputlisting[backgroundcolor=\color{lightgray},breaklines=true,basicstyle=\tiny\ttfamily, label={multi_hop_reasoning_probe_generation}, caption={Multi-hop reasoning probing question generation prompt}, captionpos=b]{Prompts/probing_questions/multi\_hop\_reasoning\_hard\_generation.tex}

\lstinputlisting[backgroundcolor=\color{lightgray},breaklines=true,basicstyle=\tiny\ttfamily, label={knowledge_update_probe_generation}, caption={Knowledge update probing question generation prompt}, captionpos=b]{Prompts/probing_questions/knowledge\_update\_generation.tex}

\lstinputlisting[backgroundcolor=\color{lightgray},breaklines=true,basicstyle=\tiny\ttfamily, label={temporal_reasoning_probe_generation}, caption={Temporal reasoning probing question generation prompt}, captionpos=b]{Prompts/probing_questions/temporal\_reasoning\_hard\_generation.tex}

\lstinputlisting[backgroundcolor=\color{lightgray},breaklines=true,basicstyle=\tiny\ttfamily, label={abstention_probe_generation}, caption={Abstention probing question generation prompt}, captionpos=b]{Prompts/probing_questions/abstention\_generation.tex}

\lstinputlisting[backgroundcolor=\color{lightgray},breaklines=true,basicstyle=\tiny\ttfamily, label={preference_following_probe_generation}, caption={Preference following probing question generation prompt}, captionpos=b]{Prompts/probing_questions/preference\_following\_generation.tex}

\lstinputlisting[backgroundcolor=\color{lightgray},breaklines=true,basicstyle=\tiny\ttfamily, label={event_ordering_probe_generation}, caption={Event ordering probing question generation prompt}, captionpos=b]{Prompts/probing_questions/event\_ordering\_generation.tex}

\lstinputlisting[backgroundcolor=\color{lightgray},breaklines=true,basicstyle=\tiny\ttfamily, label={contradiction_resolution_probe_generation}, caption={Contradiction resolution probing question generation prompt}, captionpos=b]{Prompts/probing_questions/contradiction\_resolution\_generation.tex}

\lstinputlisting[backgroundcolor=\color{lightgray},breaklines=true,basicstyle=\tiny\ttfamily, label={summarization_probe_generation}, caption={Summarization probing question generation prompt}, captionpos=b]{Prompts/probing_questions/summarization\_generation.tex}

\lstinputlisting[backgroundcolor=\color{lightgray},breaklines=true,basicstyle=\tiny\ttfamily, label={instruction_following_probe_generation}, caption={Instruction following probing question generation prompt}, captionpos=b]{Prompts/probing_questions/instruction\_following.tex}

% LLM as a judege % 
\lstinputlisting[backgroundcolor=\color{lightgray},breaklines=true,basicstyle=\tiny\ttfamily, label={llm_as_a_judge}, caption={Rubric scoring for nugget satisfaction prompt}, captionpos=b]{Prompts/llm\_as\_a\_judge\_prompt.tex}

\lstinputlisting[backgroundcolor=\color{lightgray},breaklines=true,basicstyle=\tiny\ttfamily, label={llm_equivalence_detector}, caption={Fact equivalence detection prompt}, captionpos=b]{Prompts/llm\_equivalence\_detector.tex}

%%%% Seed Generation %%%% 
\lstinputlisting[backgroundcolor=\color{lightgray},breaklines=true,basicstyle=\tiny\ttfamily, label={seed_creation}, caption={Chat titles generation prompt}, captionpos=b]{Prompts/seed\_creation.tex}

%%%% Label Generation %%%% 
\lstinputlisting[backgroundcolor=\color{lightgray},breaklines=true,basicstyle=\tiny\ttfamily, label={narrative_generation}, caption={Narrative generation prompt}, captionpos=b]{Prompts/narrative\_generation.tex}

%%%% Plan Generation %%%% 
\lstinputlisting[backgroundcolor=\color{lightgray},breaklines=true,basicstyle=\tiny\ttfamily, label={general_plan_generation}, caption={General domain conversation plan generation prompt}, captionpos=b]{Prompts/conversation\_plan\_general\_generation.tex}

\lstinputlisting[backgroundcolor=\color{lightgray},breaklines=true,basicstyle=\tiny\ttfamily, label={coding_plan_generation}, caption={Coding domain conversation plan generation prompt}, captionpos=b]{Prompts/conversation\_plan\_coding\_generation.tex}

\lstinputlisting[backgroundcolor=\color{lightgray},breaklines=true,basicstyle=\tiny\ttfamily, label={math_plan_generation}, caption={Math domain conversation plan generation prompt}, captionpos=b]{Prompts/conversation\_plan\_math\_generation.tex}

% Add special bullet % 
\lstinputlisting[backgroundcolor=\color{lightgray},breaklines=true,basicstyle=\tiny\ttfamily, label={add_special_bullets}, caption={Adding special bulletpoints to conversation plan prompt}, captionpos=b]{Prompts/add\_special\_bullets\_to\_conversation\_plan.tex}

% ten milion % 
% seed generation % 
\lstinputlisting[backgroundcolor=\color{lightgray},breaklines=true,basicstyle=\tiny\ttfamily, label={ten_million_sequential_seed_generation}, caption={Ten million sequential seed generation prompt}, captionpos=b]{Prompts/ten\_million\_generate\_seed\_following.tex}

\lstinputlisting[backgroundcolor=\color{lightgray},breaklines=true,basicstyle=\tiny\ttfamily, label={ten_million_hierarchical_seed_generation}, caption={Ten million hierarchical seed generation prompt}, captionpos=b]{Prompts/ten\_million\_generate\_seed\_similar.tex}

% Plan Generation % 
\lstinputlisting[backgroundcolor=\color{lightgray},breaklines=true,basicstyle=\tiny\ttfamily, label={ten_million_sequential_plan_generation}, caption={Ten million sequential conversation plan generation prompt}, captionpos=b]{Prompts/ten\_million\_sequential\_plan\_generation.tex}

\lstinputlisting[backgroundcolor=\color{lightgray},breaklines=true,basicstyle=\tiny\ttfamily, label={ten_million_hierarchical_plan_generation}, caption={Ten million hierarchical conversation plan generation prompt}, captionpos=b]{Prompts/ten\_million\_hierarchical\_plan\_generation.tex}

%%%% Question Generation %%%% 
\lstinputlisting[backgroundcolor=\color{lightgray},breaklines=true,basicstyle=\tiny\ttfamily, label={question_generation_general}, caption={Question generation general domain prompt}, captionpos=b]{Prompts/question\_generation\_general.tex}

\lstinputlisting[backgroundcolor=\color{lightgray},breaklines=true,basicstyle=\tiny\ttfamily, label={question_generation_coding}, caption={Question generation coding domain prompt}, captionpos=b]{Prompts/question\_generation\_coding.tex}

\lstinputlisting[backgroundcolor=\color{lightgray},breaklines=true,basicstyle=\tiny\ttfamily, label={question_generation_math}, caption={Question generation math domain prompt}, captionpos=b]{Prompts/question\_generation\_math.tex}

%%%% Answer Generation %%%% 
\begin{lstlisting}[backgroundcolor=\color{lightgray},breaklines=true,basicstyle=\tiny\ttfamily, label={check_include_question}, caption={Check assistant's response include question prompt}, captionpos=b]
You will receive an AI assistant's reply. Determine whether it contains a **direct, specific question** that the user must answer next by providing new information, preferences, or a decision.

- **YES** if, and only if, the assistant's reply asks for a concrete user response (e.g. "What's your budget for this trip?", "Which option would you prefer?").
- **NO** for generic or rhetorical prompts (e.g. "Any questions?", "Would you like to dive deeper?", "Consider your budget") that do not demand an immediate, specific answer.  

AI ASSISTANT RESPONSE:
<assistant_response>

CRITICAL NOTE: Respond only in English. Do not include any Chinese.
Output exactly **YES** or **NO**, nothing else.
\end{lstlisting}

\begin{lstlisting}[backgroundcolor=\color{lightgray},breaklines=true,basicstyle=\tiny\ttfamily, label={check_need_followup}, caption={Check need for followup prompt}, captionpos=b]
You are simulating a typical user in conversation. Decide if you would ask a follow-up question after the AI's response.

**CONVERSATION CONTEXT:**
- DOMAIN: <domain>
- TITLE: <title>
- THEME: <theme>
- SUBTOPICS: <subtopics>
- Recent History: <formatted_history>
- AI's Last Response: <assistant_response>

**ASK FOLLOW-UP ("yes") WHEN:**
1. **Missing Info**: The response lacks details you genuinely need to proceed
   - Specific steps for a process you're trying to follow
   - Key parameters (dates, amounts, requirements) for a decision you're making
   - Clarification on which option applies to your specific situation

2. **Genuine Confusion**: Something is unclear or contradictory
   - Technical terms used without explanation that block understanding
   - Conflicting information that affects your next action
   - Ambiguous instructions where the wrong interpretation has consequences

3. **Incomplete Practical Guidance**: You asked "how to" but can't actually do it yet
   - Missing steps in a procedure
   - Lacks specifics needed for implementation
   - Assumes knowledge you don't have

**NO FOLLOW-UP ("no") WHEN:**
1. **Good Enough to Proceed**: You have what you need

**OUTPUT:** Only "yes" or "no"
\end{lstlisting}

\begin{lstlisting}[backgroundcolor=\color{lightgray},breaklines=true,basicstyle=\tiny\ttfamily, label={assistant_llm_answer_generation}, caption={Assistant LLM answer generation prompt}, captionpos=b]
You must respond only in English. Never switch to Chinese or any other language mid-sentence. All responses should be entirely in English.
Respond to the user's message by either:
  • Fully answering their question
  . Provide a comprehensive answer
  • Answering plus asking at most ONE follow-up question if you need more detail  

Always honor these rules:
1. Do NOT ask questions the user already answered.  
2. Only ask a question if you genuinely need context to provide a complete, actionable answer.  
3. Keep your main answer clear and comprehensive before you ask.  
4. Use the following system inputs:

DOMAIN: <domain>
TITLE: <title>
THEME: <theme>
SUBTOPICS: <subtopics>
PREVIOUS PLANS SUMMARY: <previous_plans_summary>
PREVIOUS BATCHES OF THIS PLAN: <previous_batches> 
CURRENT HISTORY: <current_batch_messages>  

CRITICAL NOTE: Respond only in English. Do not include any Chinese.

**Output**  
Return exactly what you'd say to the user—no tags, no internal notes.
\end{lstlisting}

\lstinputlisting[backgroundcolor=\color{lightgray},breaklines=true,basicstyle=\tiny\ttfamily, label={user_llm_answer_generation}, caption={User LLM answer generation prompt}, captionpos=b]{Prompts/user\_llm\_answer\_generation.tex}

\lstinputlisting[backgroundcolor=\color{lightgray},breaklines=true,basicstyle=\tiny\ttfamily, label={user_llm_ask_followup}, caption={User LLM ask followup question prompt}, captionpos=b]{Prompts/user\_llm\_ask\_followup.tex}

%%%% Our method %%%% 
\lstinputlisting[backgroundcolor=\color{lightgray},breaklines=true,basicstyle=\tiny\ttfamily, label={kv_extraction}, caption={Key-value extraction prompt}, captionpos=b]{Prompts/baseline/kv\_creation.tex}

\lstinputlisting[backgroundcolor=\color{lightgray},breaklines=true,basicstyle=\tiny\ttfamily, label={scratchpad_creation}, caption={Scratchpad creation prompt}, captionpos=b]{Prompts/baseline/scratchpad\_creation.tex}

\lstinputlisting[backgroundcolor=\color{lightgray},breaklines=true,basicstyle=\tiny\ttfamily, label={summary_creation}, caption={Scratchpad summarization prompt}, captionpos=b]{Prompts/baseline/summary\_creation.tex}

\lstinputlisting[backgroundcolor=\color{lightgray},breaklines=true,basicstyle=\tiny\ttfamily, label={noise_filtering}, caption={Scratchpad noise filtering prompt}, captionpos=b]{Prompts/baseline/noise\_filtering.tex}

\lstinputlisting[backgroundcolor=\color{lightgray},breaklines=true,basicstyle=\tiny\ttfamily, label={rag_answer_generation}, caption={Answer generation with RAG prompt}, captionpos=b]{Prompts/baseline/rag\_answer\_generation.tex}

% \section{LLM Usage}
% We used ChatGPT\footnote{https://chatgpt.com} as a writing assistant. Specifically, we first drafted the paper and then employed ChatGPT to refine the text/assist with rephrasing and grammar. The suggestions were manually reviewed and edited before inclusion in the final version.

\end{document}

%% file: math_commands.tex
\usepackage{amsmath,amsfonts,bm}

\def\eqref#1{equation~\ref{#1}}
\def\1{\bm{1}}

\DeclareMathAlphabet{\mathsfit}{\encodingdefault}{\sfdefault}{m}{sl}
\SetMathAlphabet{\mathsfit}{bold}{\encodingdefault}{\sfdefault}{bx}{n}